\documentclass[preprint,journal]{vgtc}       

\ifpdf
  \pdfoutput=1\relax                   
  \usepackage{graphicx}                
  \DeclareGraphicsExtensions{.pdf,.png,.jpg,.jpeg} 
\else
  \usepackage{graphicx}                
  \DeclareGraphicsExtensions{.eps}     
\fi%

\graphicspath{{figures/}{pictures/}{images/}{./}} 

\usepackage{microtype}                 
\PassOptionsToPackage{warn}{textcomp}  
\usepackage{textcomp}                  
\usepackage{mathptmx}                  
\usepackage{times}                     
\usepackage{cite}                      
\usepackage{tabu}                      
\usepackage{booktabs}                  

\usepackage{amsbsy}
\usepackage{amsfonts}
\usepackage{amsmath}
\usepackage{bm}
\DeclareMathAlphabet{\mathcal}{OMS}{cmsy}{m}{n}
\usepackage{paralist}
\usepackage[utf8]{inputenc}
\usepackage{pgfplots}
\DeclareUnicodeCharacter{2212}{−}
\usepgfplotslibrary{groupplots,dateplot}
\usetikzlibrary{patterns,shapes.arrows}
\pgfplotsset{compat=newest}
\usepackage[font={sf}]{subcaption}
\usepackage{adjustbox}

\usepackage[font={scriptsize,sf}]{caption}
\newcommand{\etal}{et~al.}
\newcommand{\tr}{\mathrm{tr}}
\newcommand{\name}{ULCA}
\newcommand{\fullname}{unified linear comparative analysis}

\preprinttext{}
\ieeedoi{10.1109/TVCG.2021.3114807}

\onlineid{1570}
\vgtccategory{Research}
\vgtcpapertype{Analytics \& Decisions}

\title{Interactive Dimensionality Reduction for Comparative Analysis
\vspace{-9pt}
}

\author{Takanori Fujiwara, Xinhai Wei, Jian Zhao, and Kwan-Liu Ma}
\authorfooter{
\vspace{-0pt}
\item
 Takanori Fujiwara and Kwan-Liu Ma are with University of California, Davis. E-mail: \{tfujiwara, klma\}@ucdavis.edu.
\item
 Xinhai Wei and Jian Zhao are with University of Waterloo. E-mail: \{x67wei, jianzhao\}@uwaterloo.ca.
}

\shortauthortitle{Fujiwara \MakeLowercase{\textit{et al.}}: Interactive Dimensionality Reduction for Comparative Analysis}

\abstract{
Finding the similarities and differences between groups of datasets is a fundamental analysis task. For high-dimensional data, dimensionality reduction (DR) methods are often used to find the characteristics of each group. However, existing DR methods provide limited capability and flexibility for such comparative analysis as each method is designed only for a narrow analysis target, such as identifying factors that most differentiate groups. This paper presents an interactive DR framework where we integrate our new DR method, called ULCA (unified linear comparative analysis), with an interactive visual interface. ULCA unifies two DR schemes, discriminant analysis and contrastive learning, to support various comparative analysis tasks. To provide flexibility for comparative analysis, we develop an optimization algorithm that enables analysts to interactively refine ULCA results. Additionally, the interactive visualization interface facilitates interpretation and refinement of the ULCA results. We evaluate ULCA and the optimization algorithm to show their efficiency as well as present multiple case studies using real-world datasets to demonstrate the usefulness of this framework. 
\vspace{-12pt}
}

\keywords{Dimensionality\,reduction,\,discriminant\,analysis,\,contrastive\,learning,\,comparative\,analysis,\,interpretability,\,visual\,analytics.\vspace{-3pt}}

\nocopyrightspace

\vgtcinsertpkg

\begin{document}

\firstsection{Introduction}
\maketitle

\setlength{\abovedisplayskip}{2pt}
\setlength{\belowdisplayskip}{1pt}

\newcommand{\nInsts}{n}
\newcommand{\nInstsTg}{{n_\textrm{tg}}}
\newcommand{\nInstsBg}{{n_\textrm{bg}}}
\newcommand{\nAttrs}{d}
\newcommand{\nLatFeats}{{d'}}

\newcommand{\OrgMat}{\mathbf{X}}
\newcommand{\OrgMatTg}{{\mathbf{X}_\textrm{tg}}}
\newcommand{\OrgMatBg}{{\mathbf{X}_\textrm{bg}}}
\newcommand{\ReprMat}{\mathbf{Z}}
\newcommand{\ReprMatTg}{{\mathbf{Z}_\textrm{tg}}}
\newcommand{\ReprMatBg}{{\mathbf{Z}_\textrm{bg}}}
\newcommand{\ProjMat}{\mathbf{M}}

\newcommand{\Cov}{\mathbf{C}}
\newcommand{\CovTg}{{\mathbf{C}_\textrm{tg}}}
\newcommand{\CovBg}{{\mathbf{C}_\textrm{bg}}}
\newcommand{\CovWithin}{{\mathbf{C}_\textrm{wi}}}
\newcommand{\CovBetween}{{\mathbf{C}_\textrm{bw}}}
\newcommand{\CovWithinEach}[1]{{\mathbf{C}_{\textrm{wi}_#1}}}
\newcommand{\CovBetweenEach}[1]{{\mathbf{C}_{\textrm{bw}_#1}}}

\newcommand{\Label}{y}
\newcommand{\LabelVec}{\mathbf{y}}

\newcommand{\InstVec}{\mathbf{x}}
\newcommand{\MeanVec}{\bm{\mu}}

\newcommand{\Weight}{w}
\newcommand{\WeightTg}{{\mathbf{w}_\textrm{tg}}}
\newcommand{\WeightBg}{{\mathbf{w}_\textrm{bg}}}
\newcommand{\WeightBw}{{\mathbf{w}_\textrm{bw}}}
\newcommand{\WeightTgEach}[1]{{w_{\textrm{tg}_#1}}}
\newcommand{\WeightBgEach}[1]{{w_{\textrm{bg}_#1}}}
\newcommand{\WeightBwEach}[1]{{w_{\textrm{bw}_#1}}}

\newcommand{\ContParam}{\alpha}
\newcommand{\RegConst}{\gamma}

\newcommand{\Params}{\theta}
\newcommand{\Dist}{l}
\newcommand{\Area}{a}
\newcommand{\TargetClass}{k}
\newcommand{\WeightDist}{r_{\Dist}}
\newcommand{\WeightArea}{r_{\Area}}
\newcommand{\CostDist}{J_{\Dist}(\Params)}
\newcommand{\CostArea}{J_{\Area}(\TargetClass, \Params)}
\newcommand{\DistIdeal}[1]{{\Dist'_{#1}}}
\newcommand{\DistNew}[1]{{\hat{\Dist}_{#1}}}
\newcommand{\AreaIdeal}[1]{{\Area'_{#1}}}
\newcommand{\AreaNew}[1]{{\hat{\Area}_{#1}}}

\newcommand{\norm}[1]{\left\lVert #1 \right\rVert}

The comparison of two or more groups of datasets is a common analysis task to identify factors that make the groups different from or similar to each other. 
For example, by comparing gut microbiota between cohorts of colorectal cancer patients and healthy subjects, we can identify which bacteria composition highly contributes to inducing cancer formation or keeping a healthy gut environment~\cite{sobhani2019colorectal,hofseth2020early}.
Also, when analyzing the general public's political opinions, political scientists may want to find unique characteristics in some political party's supporters when comparing them with others~\cite{fujiwara2020contrastive,hare2015using}. 
This type of comparative analysis is universal and can be found in numerous domains, such as researches in healthcare~\cite{guo2020comparative,jin2020carepre}, biomedicine~\cite{yasir2015comparison,kvam2012comparison}, and sociology~\cite{moore2015social,hill2015real}.

Various approaches are available for comparison tasks, including statistical hypothesis tests~\cite{du2010choosing} and visual  comparison~\cite{gleicher2011visual,gleicher2018considerations}. 
Among others, dimensionality reduction (DR) methods, such as principal component analysis (PCA)~\cite{hotelling1933analysis,jolliffe1986principal} and linear discriminant analysis (LDA)~\cite{izenman2008modern}, play an important role for comparative analysis, especially when datasets have a large number of attributes~\cite{legendre2013beta}. 
From many attributes, DR generates a small number of latent features, with which the similarities of data points across groups can be represented as their spatial proximities in a lower-dimensional (or embedding) space.  
By referring to the ``similarity $\approx$ proximity''~\cite{wenskovitch2018towards} relationship, we can visually identify useful patterns, including subgroups and outliers, while considering the combinational effects from multiple attributes.
Also, linear DR methods (e.g., PCA), which provide a linear mapping from the original attributes to latent features, produce interpretable axes in the embedding space.
By interpreting the axes, we can further identify highly influential attributes on,\,for\,example,\,differences\,among\,groups~\cite{brehmer2014visualizing,fujiwara2019supporting}. 

However, existing DR methods provide limited capability and flexibility for comparative analysis.
General-purpose DR methods that produce embeddings without group information (e.g., PCA) do not prioritize extracting patterns that highly relate to group differences or similarities, which is vital for comparative analysis. 
Only a handful of DR methods such as LDA and contrastive PCA (cPCA)~\cite{abid2018exploring} are specifically designed for comparative analysis; however, even these methods are only tailored for a narrow analysis target (e.g., LDA is for finding factors that most differentiate groups).
Additionally, these DR methods do not provide the functionality that allows analysts to perform hypothetical changes on an embedding result (e.g., changing data points' positions) and then link these changes to the parameters of the DR algorithms in order to produce a new result that resembles the hypothetical changes~\cite{endert2011observation}. 
This functionality is important to interactively adjust an embedding result to intuitively find certain patterns of analysts' target interest while actively involving human-in-the-loop~\cite{lage2018human}.

To address the aforementioned problems, we introduce a novel visual analytics framework for comparative analysis, which consists of a new DR method, called \textit{\name{} (\fullname{})}, an interactive parameter optimization algorithm, and a visual interface. 
\name{} is an exploratory data analysis tool that unifies two DR schemes, discriminant analysis and contrastive learning, to support comparisons that cannot be achieved when using only one of these schemes.
More specifically, \name{} is a linear DR method that not only comprises the functionalities of PCA, cPCA, and LDA but also fills the gaps among analysis targets of these methods.
For instance, unlike ordinary LDA, \name{} can be used to find latent features that distinguish multiple groups while, at the same time, producing a higher variance for a particular group.
In this way, we can, for example, find a political stance that clearly separates the supporters of each political party while still encompassing the diverse opinions of the supporters of a certain party.
\name{} also provides detailed control of each group's contribution to the embedding, which allows flexible comparisons based on the analyst's interest.
Additionally, to help analysts intuitively adjust the related parameters, we develop a backward optimization algorithm that updates the embedding result by finding optimal parameters to achieve the analyst's demonstrated changes in the embedding result.

Within our framework, we develop a visual interface that provides the essential functionalities to visualize, interpret, and interact with the results of \name{}. 
To support the wide, ever-changing analysis needs, rather than developing a tool supporting all possible analysis tasks, we design our interface to be easily integrated with existing analysis and visualization libraries.
Specifically, our interface can be directly used with Python and the Jupyter Notebook~\cite{jupyter} (which supports the interactive execution of Python scripts), resulting in broader and easier adoptions.
Consequently, analysts can effortlessly apply any analytical processing (e.g., normalization) with existing libraries before applying \name{} or even utilize the interactively refined \name{} result for further analysis (e.g., reuse of obtained latent features for other data). 

To demonstrate the efficiency of the algorithms of \name{} and the backward optimization, we conduct a performance evaluation. 
The results show that the algorithms have a reasonable computational cost for interactive use.
Also, the results guide appropriate settings for the backward optimization based on a desirable balance of latency and accuracy.
We also demonstrate the effectiveness of our framework with multiple case studies using publicly available datasets. 
We provide source code of the framework, a demonstration video of the interface, and a comprehensive qualitative comparison between ULCA and other DR methods in the supplementary material at \href{https://takanori-fujiwara.github.io/s/ulca/}{\texttt{https://takanori-fujiwara.github.io/s/ulca/}}.

In summary, our main contributions include:
\begin{compactitem}
    \item A new linear DR method, \name{}, which unifies and enhances two DR schemes, discriminant analysis and contrastive learning.  
    \item A backward optimization algorithm that converts a manipulation on an embedding result into \name{}'s parameters to produce a new embedding similar to the manipulated result.
    \item A visual interface that allows analysts to not only visualize and interact with a \name{} result but also to use \name{} with existing analysis environments and visualization libraries.
    \item Performance evaluation and case studies with real-world datasets to assess the efficacy of our framework for comparative analysis.
\end{compactitem}

\section{Background and Related Work}
We provide the relevant background and works in DR methods and their enhancement in interactive usage.

\subsection{Dimensionality Reduction Methods}
\label{sec:dr_methods}

DR is an essential tool to analyze high-dimensional data~\cite{liu2017visualizing,sacha2017visual} as it can provide a succinct low-dimensional overview while preserving the essential information of the original data (e.g., data variance when using PCA~\cite{hotelling1933analysis,jolliffe1986principal}).
DR methods can be categorized as either linear or nonlinear DR based on how they produce embeddings. 
Linear DR, such as PCA and classical multidimensional scaling (MDS)~\cite{torgerson1952}, can be defined as DR that produces a linear transformation matrix (or projection matrix) $\smash{\ProjMat \in \mathbb{R}^{\nAttrs \times \nLatFeats}}$ where $\nAttrs$ and $\smash{\nLatFeats}$ are the numbers of dimensions in original and embedding spaces~\cite{cunningham2015linear}.
A projection matrix $\ProjMat$ is derived by solving each DR's optimization problem.
By using $\ProjMat$, from the original dataset $\smash{\OrgMat \in \mathbb{R}^{\nInsts \times \nAttrs}}$ ($\nInsts$ is the number of data points), a linear DR method can produce an embedding result $\smash{\ReprMat \in \mathbb{R}^{\nInsts \times \nLatFeats}}$ with $\ReprMat = \OrgMat \ProjMat$.\,When\,the\,use\,of\,$\ReprMat$\,aims\,for\,exploratory\,data\,analysis,\,linear DR is sometimes called projection pursuit (PP)~\cite{huber1985projection,jones1987projection}. 
For PP, the optimization problem is finding an embedding space that captures structures of the user's interest.
For example, PCA can be considered a PP method that shows directions containing high data variance~\cite{jones1987projection}. 
Several methods have been developed under the concept of PP~\cite{pires2010projection,posse1992projection,zhou2010stable}.

While linear DR can only preserve the linear structure of $\OrgMat$ in $\ReprMat$, nonlinear DR aims to capture the nonlinear structure of $\OrgMat$. 
For example, many nonlinear DR methods used for visualization, such as t-SNE~\cite{maaten2008visualizing} and UMAP~\cite{mcinnes2018umap}, aim to preserve local neighborhoods for each data point, which is often difficult when relying only on a linear transformation. 
These methods first generate a neighbor graph where each edge represents a dissimilarity of nodes (i.e., data points); then, they maximally preserve local neighborhoods of each data point in an embedding space. 
While nonlinear DR has the advantage to capture the nonlinear structure, many nonlinear DR methods do not provide a parametric mapping from $\OrgMat$ to $\ReprMat$. 
Consequently, the interpretation of embedding results is often difficult~\cite{fujiwara2019supporting}.
On the other hand, embedding results by linear DR can be interpreted from $\ProjMat$, which shows how embedding axes are derived from the original dimensions.
Comprehensive information of DR methods can be found in several surveys~\cite{van2009dimensionality,cunningham2015linear,espadoto2021toward}.

Several DR methods can be used for comparative analysis of multiple groups of data points.
Discriminant analysis~\cite{mika1999fisher,hastie1996discriminant,izenman2008modern}, such as LDA, is designed to differentiate multiple groups by finding an embedding space where the separation of each group is maximized. 
Canonical correlations analysis (CCA)~\cite{hotelling1992relations,yang2019survey} finds latent features for each of two different datasets such that the correlation between each latent feature is maximized. 
The result informs which combination of attributes can better explain the relationships between two datasets. 
Recently, contrastive learning~\cite{zou2013contrastive} is introduced to find salient patterns in one dataset compared to another.
For example, cPCA~\cite{ge2016rich,abid2018exploring} is the extended version of PCA for contrastive learning and produces an embedding space where one group has a high variance but another group does not. 

For comparative analysis, our work utilizes linear DR as the interpretability of the embedding result is vital to support human-in-the-loop analysis~\cite{lage2018human} for gaining new insights through interactive analysis. 
We introduce a new linear DR method, \name{}, which unifies and enhances LDA and cPCA to perform flexible comparative analysis.

\subsection{Interactive Dimensionality Reduction}

Researchers have studied how effectively DR methods can be used with interactive visualizations~\cite{nonato2018multidimensional,sacha2017visual,jiang2019recent}.
A comprehensive survey of the related studies is provided by Sacha \etal~\cite{sacha2017visual}.
The survey also reveals common interaction scenarios, such as refinement of DR results by tuning parameters or selecting a subset of data.
Here, we focus on discussing interactive DR using \textit{parametric interaction} and \textit{observation-level interaction}~\cite{self2018observation}. 
Self \etal~\cite{self2018observation} defined that parametric interaction is to directly adjust parameters of a DR method while observation-level interaction is to manipulate data points in an embedding result (e.g., changing their positions) and interpret the semantic meaning of manipulation in order to update the embedding result accordingly.

As parametric interactions, for example, iPCA~\cite{jeong2009ipca} supports adjustment of each attribute's contribution to a PCA result.
P{\'e}rez et al.~\cite{perez2015interactive} enabled de-cluttering DR results by controlling how strongly DR places data points close to their cluster centers.
Wang et al.~\cite{wang2017linear} visualized an LDA result with star coordinates~\cite{kandogan2000star} and allowed analysts to update each attribute's contribution to the LDA result by interactively adjusting the length of the star coordinate axes.
Coimbra et al.~\cite{coimbra2016explaining} developed enhanced star coordinates to help understand a 3D embedding result.
With their method, analysts can find an optimal viewpoint, with which the distribution of selected attributes can be easily observed.
Johansson and Johansson~\cite{johansson2009interactive} designed a quality measure that consists of a weighted combination of correlation, outlier detection, and cluster detection qualities. 
Analysts can adjust the weights and extract a set of attributes that maximizes the defined quality measure. 
Explainers~\cite{gleicher2013explainers} also generate projection functions based on the user-defined tradeoffs among correctness, simplicity, and diversity of resultant projections.

Observation-level interactions are often designed to manipulate a few data points' positions in an embedding result.
For instance, Endert \etal~\cite{endert2011observation} introduced user-guided weighted MDS (WMDS). 
This method automatically updates its algorithm parameters based on the rearranged data points' positions.
Dis-function~\cite{brown2012disfunction} and Andromeda~\cite{self2016bridging} employ similar approaches with the user-guided WMDS. 
Joia et al.~\cite{joia2011local} and Mamani et al.~\cite{mamani2013user} also utilizes user-specified data points as control points to generate desired DR results~\cite{joia2011local,mamani2013user}. 
SIRIUS~\cite{dowling2019sirius} extends the user-guided WMDS to handle the manipulation of both data points and attributes. 
Pollex~\cite{wenskovitch2019pollux} supports a cluster-centric interaction to change a cluster assignment by moving a data point from a current cluster to the other. 
A few methods such as InterAxis~\cite{kim2016interaxis} and AxiSketcher~\cite{kwon2017axisketcher} take an approach that generates embedding axes by directly indicating how data points should be arranged along the axes.

Unlike the above works, our work focuses on the usage of parametric and observation-level interactions for comparative analysis. 
In addition, while the existing observation-level interactions are designed for manipulating individual data points, we provide interactions that can be performed on a group of data points. 

\begin{figure*}[tb]
    \centering
    \includegraphics[width=\linewidth]{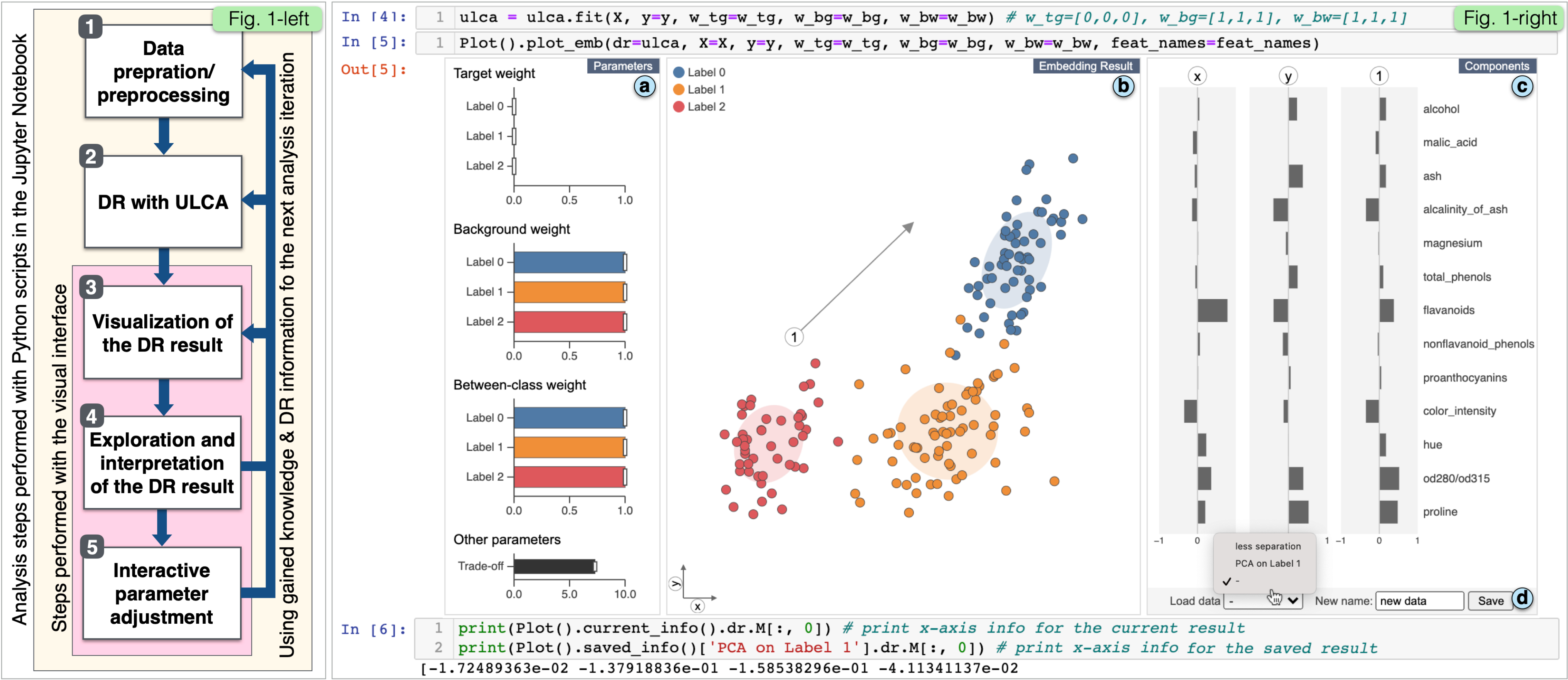}
    \caption{A typical comparative analysis workflow with our framework (left). Following the analysis workflow, the analyst is analyzing the Wine dataset~\cite{uci_mlr} with our framework in the Jupyter Notebook (right). 
    The result produced with \name{} is visualized with the UI of our framework. (a) The parameter view shows the used parameters. The analyst can interactively adjust parameters with this view. (b) The embedding result view depicts a lower-dimensional representation of the dataset. The ellipse represents each group's confidence ellipse~\cite{jolliffe1986principal} (by default, 50\% confidence ellipse). The analyst can directly manipulate the centroid or scatteredness of each group by moving or scaling the corresponding confidence ellipse to trigger the backward parameter selection.
    (c) The component view informs a numerical mapping from the original attributes to each component (i.e., the axis in the embedding result view). 
    (d) The analyst can store the current state of visualizations and parameters by using a saving function.}
    \label{fig:ui}
\end{figure*}

\vspace{-4pt}
\section{Analysis Workflow and Example}
\label{sec:workflow_and_example}
\vspace{-2pt}

We introduce a typical comparative analysis workflow (\autoref{fig:ui}-left) with our framework while demonstrating it with an analysis example. 
Here, we analyze the Wine dataset~\cite{uci_mlr} by using Python scripts and our framework in the Jupyter Notebook.
The dataset includes 178 data points/wines with 13 attributes (e.g., alcohol percentage) and consists of three predefined groups (corresponding to three different cultivars).

The workflow starts from (1) \textit{data preparation and preprocessing}, such as the assignment of data groups, data cleaning, and normalization~\cite{garcia2015data}. 
For example, we load the Wine dataset and their predefined group information, and apply normalization to each attribute.
Then, we can (2) \textit{apply DR using \name{}}, which we introduce in \autoref{sec:methodology}, to the processed dataset to find latent features that capture characteristics specific to each group or similar to each other. 
For this analysis example, to know whether or not there exist differentiating factors among the groups, we apply \name{} to the dataset with parameters that produce the same result when applying LDA, as shown in `\texttt{In [4]}' in \autoref{fig:ui}-right.

For the following steps (3--5), we can utilize the visual interface.
We can first (3) \textit{visualize the DR result}.
In \autoref{fig:ui}-right, we invoke the visual interface in `\texttt{In [5]}', and `\texttt{Out [5]}' shows the \name{} result.
Afterward, we can (4) \textit{explore and interpret the DR result}.
For example, based on the embedding result in \autoref{fig:ui}b, we can find the three groups (Labels 0--2) are well separated. 
Additionally, to understand factors related to this separation, we review the information of $x$- and $y$-axes in \autoref{fig:ui}c, where the bar charts show a linear mapping from the original attributes to each axis.
As the absolute value of the bar chart approaches 1, the corresponding attribute has a higher influence on the axis.
From \autoref{fig:ui}c, we can expect that the three groups have clearly different values in, for example, \texttt{flavanoids} ($x$-axis) and \texttt{proline} ($y$-axis).

\begin{figure}[tb]
    \centering
    \includegraphics[width=0.95\linewidth]{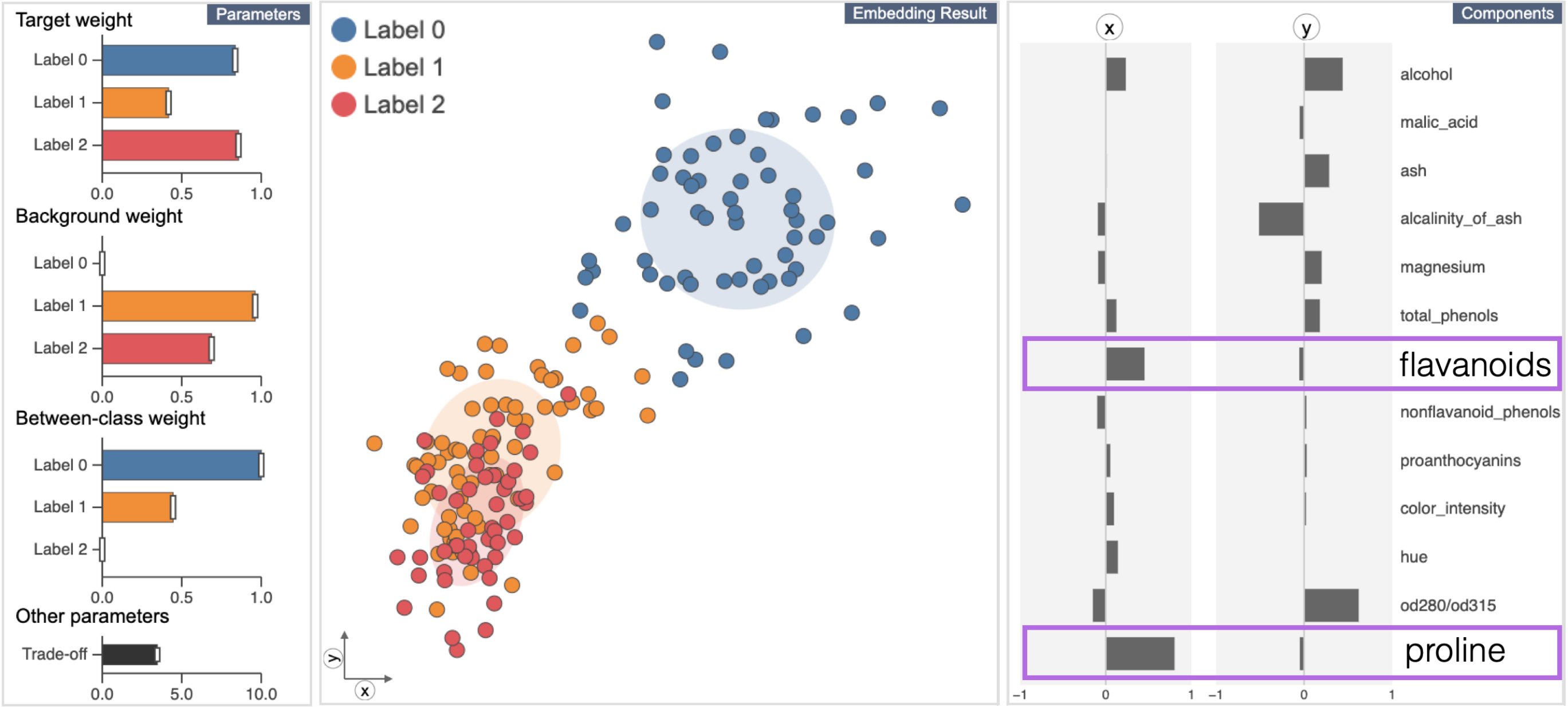}
    \caption{
    The \name{} result after the backward parameter selection.
    }
    \label{fig:example1}
\end{figure}

\begin{figure}[tb]
    \centering
    \includegraphics[width=0.95\linewidth]{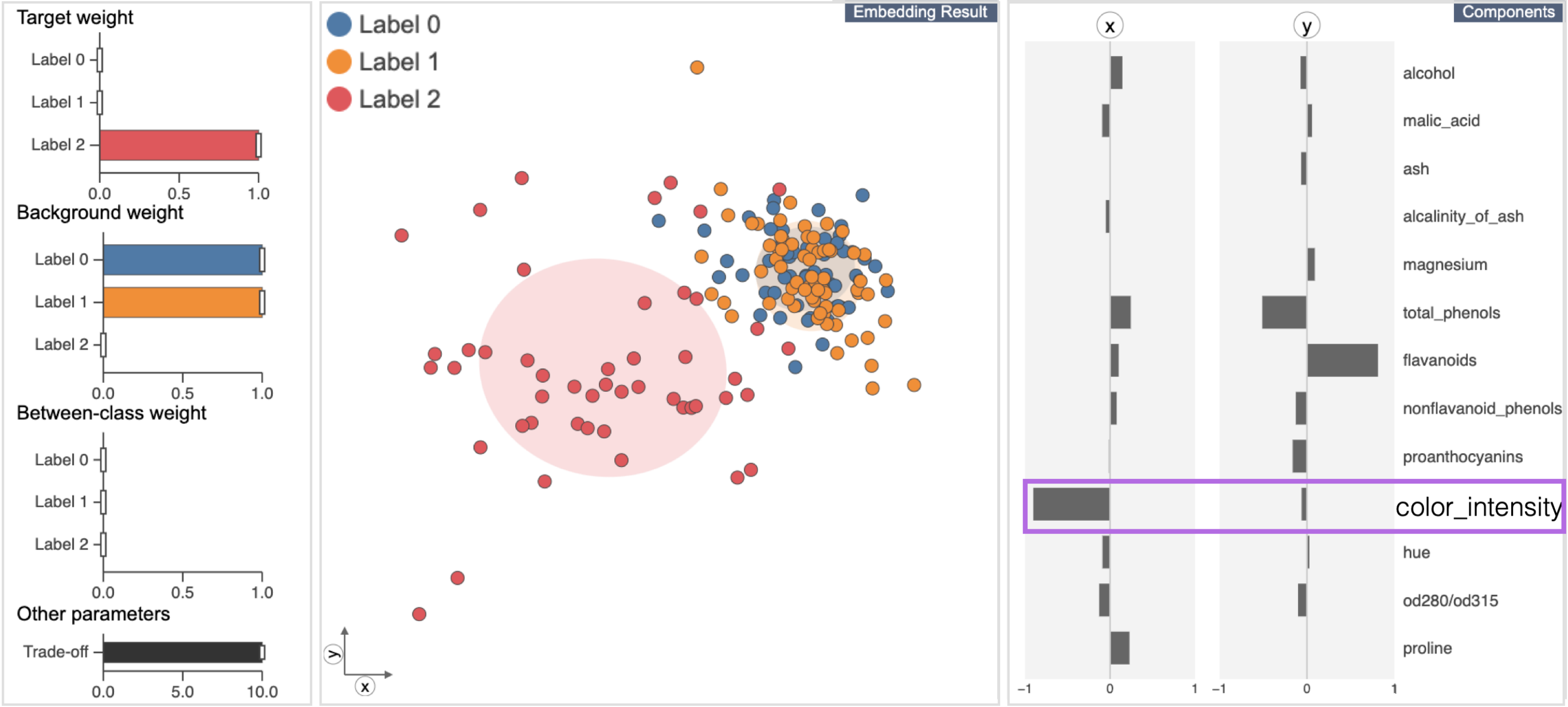}
    \caption{
    The \name{} result after the parameter adjustment using \autoref{fig:ui}a.
    }
    \label{fig:example2}
\end{figure}

Lastly, to refine the DR result or analyze the data from a different perspective, we can (5) \textit{interactively set different parameters} using sliders in \autoref{fig:ui}a or changing a group's position or scatteredness in \autoref{fig:ui}b.
The change triggers rerunning of \name{}.
Also, we can utilize the gained knowledge and DR results for the next analysis loop (the backward arrows in \autoref{fig:ui}-left).
For example, as we have already identified the differentiating factors of the groups, we next want to know factors common between Labels 1 and 2 but different from Label 0.
In \autoref{fig:ui}b, by dragging the confidence ellipse of Label 2 (shown as a red ellipse), we place a centroid of Label 2 to a close position of a centroid of Label 1. 
The backward parameter selection we introduce in \autoref{sec:param_optim}  finds parameters to resemble the indicated change, and updates the result, as shown in \autoref{fig:example1}.
\name{} has successfully produced an embedding result where the points in Labels 1 and 2 are placed close to each other but Label 0 is placed far away from the others.
By looking at the axis information in \autoref{fig:example1}, we notice that \texttt{proline} and \texttt{flavanoids} (annotated with purple) have strong influences on $x$-axis.

We further want to know factors that wines in Label 2 have a high variety than the others. 
To achieve this, we update parameters with the sliders in \autoref{fig:ui}a so that the variance of Label 2 is maximized while minimizing the variances of Labels 0 and 1. 
The result is instantaneously updated, as shown in \autoref{fig:example2}, where we can see Label 2 has a much higher variance than the others along $x$-axis.
From \autoref{fig:example2}, we observe that \texttt{color\_intensity} has a prominent influence on $x$-axis; thus, Label 2 seems to consist of the wines with various color brightness. 

The above analysis with the typical workflow presents that our framework provides flexible analysis and helps identify multiple characteristics of groups.
Such flexibility and capability are not supported by existing methods, such as LDA and cPCA.

\vspace{-2pt}
\section{Methodology}
\label{sec:methodology}
\vspace{-1pt}

This section introduces the design and derivation of \name{} in detail. 
\autoref{table:notation} summarizes the notations used throughout the paper.

\vspace{-2pt}
\subsection{Existing Linear DR Used for Comparative Analysis}
\label{sec:ldr_for_ca}
\vspace{-1pt}

Since \name{} unifies several linear DR methods, we start from a brief introduction to the related methods, specifically, PCA, cPCA, and LDA.

\noindent\textbf{PCA.}\,PCA~\cite{hotelling1933analysis}\,finds\,latent\,features\,which\,maximally\,capture\,the variance of\,a\,whole\,dataset\,in\,an\,embedding\,space.\,The optimization to identify such $\nLatFeats$ latent features from $\nAttrs$ attributes ($\nLatFeats\!\!\leq\!\nAttrs$) can be written as:
\begin{align}
    \max_{\ProjMat^\top \ProjMat = I_\nLatFeats} ~ \tr(\ProjMat^\top \Cov \ProjMat)
\end{align}
where $\smash{\Cov \in \mathbb{R}^{\nAttrs \times \nAttrs}}$ is a covariance matrix of the original data $\smash{\OrgMat \in \mathbb{R}^{\nInsts \times \nAttrs}}$ ($\nInsts$: the number of data points) and $\smash{\ProjMat \in \mathbb{R}^{\nAttrs \times \nLatFeats}}$ is a projection matrix (note: $\smash{I_\nLatFeats}$ is a $\smash{\nLatFeats \times \nLatFeats}$ identify matrix). 
This optimization is often solved with singular value decomposition, eigenvalue decomposition (EVD), or manifold optimization~\cite{absil2009optimization} (we describe the details in \autoref{sec:optimization}).

\noindent\textbf{cPCA.} cPCA~\cite{ge2016rich,abid2018exploring} is a variant of PCA for contrastive learning. 
Contrastive learning~\cite{zou2013contrastive} aims to find salient features in one group (target group) by comparing it with another group (background group). 
Within this scheme, cPCA specifically finds latent features, with which a target group has a high variance but a background has a low variance (i.e., the saliency is identified in variance).  
This optimization can be written as: 
\begin{align}
\label{eq:cpca}
    \max_{\ProjMat^\top \ProjMat = I_\nLatFeats} ~ \tr \bigl( \ProjMat^\top (\CovTg - \ContParam \CovBg ) \ProjMat \bigr)
\end{align}
where $\smash{\CovTg}$, $\smash{\CovBg \in \mathbb{R}^{\nAttrs \times \nAttrs}}$ are covariance matrices of target and background groups  $\smash{\OrgMatTg \in \mathbb{R}^{\nInstsTg \times \nAttrs}}$, $\smash{\OrgMatBg \in \mathbb{R}^{\nInstsBg \times \nAttrs}}$, respectively ($\smash{\nInstsTg}$, $\smash{\nInstsBg}$: the number of data points in each group).  
$\ContParam$ $(0 \leq \ContParam \leq \infty)$ is a hyperparameter, called a contrast parameter, which controls the trade-off between having a high target variance and a low background variance. 
When $\ContParam = 0$, cPCA only maximizes the variance of $\smash{\OrgMatTg}$ (i.e., PCA on $\smash{\OrgMatTg}$).
As $\ContParam$ increases, cPCA more focuses on reducing the variance of $\smash{\OrgMatBg}$.

cPCA is used for finding patterns hidden by dominant variance from features that do not relate to analysis interests (examples are provided by Abid and Zhang \etal~\cite{abid2018exploring}).
In addition to this original usage (i.e., eliminating the influence from uninterested variations), cPCA can be used to compare two groups and to find more variety patterns in one group, such as political opinions diverse in supporters of a certain political party but uniform in the other supporters~\cite{fujiwara2020contrastive}. 

\noindent\textbf{LDA.} LDA~\cite{izenman2008modern} uses predefined group/class information to find an embedding that maximizes separation among groups. 
To do so, LDA minimizes data variance within each group while maximizes the separation of each group's centroid. 
The\,optimization\,of\,LDA\,can\,be\,written\,as: 
\begin{align}
\label{eq:lda}
    \max_{\ProjMat^\top \ProjMat = I_\nLatFeats} ~ \frac{\tr(\ProjMat^\top \CovBetween \ProjMat)}{\tr(\ProjMat^\top \CovWithin \ProjMat)}
\end{align}
where $\smash{\CovWithin}$ and $\smash{\CovBetween}$ are \textit{within-class} and \textit{between-class} covariance matrices. 
These are computed with 
$\smash{\CovWithin = n^{-1} \sum_{i=1}^{n} (\InstVec_i - \MeanVec_{\Label_i})(\InstVec_i - \MeanVec_{\Label_i})^\top}$,  
$\smash{\CovBetween = n^{-1} \sum_{i=1}^{n} (\MeanVec_{\Label_i} - \MeanVec)(\MeanVec_{\Label_i} - \MeanVec)^\top}\!$ where $\smash{\InstVec_i \in \mathbb{R}^\nAttrs}$ is the $i$-th row of the input data $\OrgMat$, $\smash{\MeanVec \in \mathbb{R}^\nAttrs}$ is the column means of $\OrgMat$, and $\smash{\MeanVec_{\Label_i} \in \mathbb{R}^\nAttrs}$ 
is the column means of data points in a class that $\smash{\InstVec_i}$ belongs to.
Here, $\smash{\Label_i \in \{1, \ldots, c\}}$ ($c$: the number of classes or groups) is the $i$-th element of $\LabelVec$, a vector containing a group label for each data point.

\vspace{-0pt}
\subsection{Unified Linear Comparative Analysis (\name{})}
\label{sec:ulca}
\vspace{-0pt}

Here, we introduce \name{}, which unifies and enhances cPCA and LDA.
\name{} embraces functionalities of various linear DR methods, including PCA, cPCA, LDA, among others. 
As the intermediate product of \name{}, we introduce a generalized version of cPCA (gcPCA), which enables analysts to apply cPCA to any number of groups. 

\vspace{-2pt}
\subsubsection{Generalization of cPCA}
\vspace{-2pt}

We generalize cPCA, which originally compares only two groups. 
Assume we have $c$ groups and let \scalebox{0.9}{$\smash{\CovWithinEach{j}}$} be a covariance matrix of data points in group $j$, i.e., \scalebox{0.9}{$\smash{\CovWithinEach{j} = (\sum_{i=1}^{n} \delta_{\Label_i}^{\scriptscriptstyle j})^{-1} \sum_{i=1}^{n} \delta_{\Label_i}^{\scriptscriptstyle j} (\InstVec_i - \MeanVec_{\Label_i})(\InstVec_i - \MeanVec_{\Label_i})^\top}$} where \scalebox{0.9}{$\smash{\delta_{\Label_i}^{\scriptscriptstyle j} = 1}$} when $\smash{\Label_i = j}$, otherwise \scalebox{0.9}{$\smash{\delta_{\Label_i}^{\scriptscriptstyle j} = 0}$}.  
With weights $\smash{\WeightTg}$ and $\smash{\WeightBg}$, of which $j$-th elements $\smash{\WeightTgEach{j}}$ and $\smash{\WeightBgEach{j}}$ ($\smash{0 \leq \WeightTgEach{j} , \WeightBgEach{j} \leq 1}$) represent contributions of group $j$'s covariance to target and background variances, the  optimization problem of gcPCA can be written as:
\begin{align}
\label{eq:gcpca}
    \max_{\ProjMat^\top \ProjMat = I_\nLatFeats} ~ \tr \biggl( \ProjMat^\top \Bigl( \sum_{j=1}^{c} \WeightTgEach{j} \CovWithinEach{j} -  \ContParam \sum_{j=1}^{c} \WeightBgEach{j} \CovWithinEach{j} \Bigr) \ProjMat \biggr).
\end{align}
Using $\smash{\WeightTg}$ and $\smash{\WeightBg}$, gcPCA allows any groups to be target or background. 
For example, when $c=2$, $\smash{\WeightTg = (1, 0)}$, and $\smash{\WeightBg = (0, 1)}$, \autoref{eq:gcpca} reduces to cPCA (\autoref{eq:cpca}), where groups 1 and 2 are target and background, respectively.
It is noteworthy that when $c=2$, $\smash{\WeightTg = (1, 1)}$, and $\smash{\WeightBg = (0, 1)}$, \autoref{eq:gcpca} reduces to ccPCA~\cite{fujiwara2019supporting}, which is an enhanced version of cPCA developed for characterizing clusters identified in the DR result.
Moreover, with decimal weights, gcPCA can precisely control the effect from each group variance to an embedding. 

\begin{table}[t]
  \renewcommand{\arraystretch}{0.8}
  \small
  \centering
  \caption{Summary of notation.\vspace{-2pt}}
  \label{table:notation}
  \makeatletter
\def\thickhline{%
  \noalign{\ifnum0=`}\fi\hrule \@height \thickarrayrulewidth \futurelet
   \reserved@a\@xthickhline}
\def\@xthickhline{\ifx\reserved@a\thickhline
               \vskip\doublerulesep
               \vskip-\thickarrayrulewidth
             \fi
      \ifnum0=`{\fi}}
\makeatother

\newlength{\thickarrayrulewidth}
\setlength{\thickarrayrulewidth}{2\arrayrulewidth}

\begin{tabular}{rl}
\thickhline{}
 $\nInsts$, $\nAttrs$, $\smash{\nLatFeats}$, $c$ & \# of data points, original attributes, latent features, groups/classes.\\
 $\OrgMat$, $\LabelVec$, $\ReprMat$ & Original dataset, group labels of data points, embedding result.\\
 $\ProjMat$ & Projection matrix. \\
 $\smash{\CovWithinEach{j}}$, $\smash{\CovBetweenEach{j}}$ & $j$-th group's within-class and between-class covariance matrices.\\
 $\smash{\WeightTg}$, $\smash{\WeightBg}$, $\smash{\WeightBw}$ & Weights for target, background,  between-class covariance matrices.\\
 $\ContParam$ & Trade-off parameter (or contrast parameter).\\ 
\thickhline{}
\end{tabular}
  \vspace{-4mm}
\end{table}

\vspace{-2pt}
\subsubsection{Trace-Ratio Form of gcPCA}
\label{sec:trace_ratio_form_gcpa}
\vspace{-2pt}

LDA and gcPCA have a different form of the optimization problem with each other (refer to \autoref{eq:lda} and \autoref{eq:gcpca}). 
To enable integration of gcPCA and LDA, we introduce the trace-ratio form~\cite{jia2009trace} of gcPCA.

As shown in \autoref{eq:cpca} and \ref{eq:gcpca}, cPCA and gcPCA's optimization problems are written as the trace-difference problem (i.e., maximizing the difference of matrix traces)~\cite{jia2009trace}. 
As discussed by Fujiwara \etal~\cite{fujiwara2020interpretable}, when we want to maximize the variance of the target matrix while simultaneously minimizing the variance of the background matrix, the optimization problem of cPCA can be converted into the maximization of $\smash{\tr(\ProjMat^\top \CovTg \ProjMat) / \tr(\ProjMat^\top \CovBg \ProjMat)}$, which is the trace-ratio problem (i.e., maximizing the ratio of matrix traces).
Similarly, for the same purpose, gcPCA can be written as the following trace-ratio problem:
\begin{align}
\label{eq:gcpca_trace_ratio}
    \max_{\ProjMat^\top \ProjMat = I_\nLatFeats} ~ \frac{ \tr \Bigl( \ProjMat^\top \bigl(\sum_{j=1}^{c} \WeightTgEach{j} \CovWithinEach{j} \bigr) \ProjMat \Bigr)} {\tr \Bigl( \ProjMat^\top \bigl( \sum_{j=1}^{c} \WeightBgEach{j} \CovWithinEach{j} \bigr) \ProjMat \Bigr)}.
\end{align}
As proved by Guo \etal~\cite{guo2003generalized}, the optimization problem of \autoref{eq:gcpca_trace_ratio} is equivalent to find $\ContParam$ that produces zero as the optimum value of \autoref{eq:gcpca}.
We describe an algorithm to find such $\ContParam$ in \autoref{sec:optimization}.
Here, we want to note that we can regard gcPCA with \autoref{eq:gcpca} as a relaxed problem that finds $\ProjMat$ with the user-specified $\ContParam$. 
Now, we can handle both LDA (\autoref{eq:lda}) and gcPCA (\autoref{eq:gcpca_trace_ratio}) as the trace-ratio problem.

\vspace{-1pt}
\subsubsection{Integration of gcPCA and LDA}
\label{sec:integration_gcpca_lda}
\vspace{-1pt}

We introduce the optimization problem of \name{} by integrating gcPCA and LDA.
By comparing \autoref{eq:lda} and \autoref{eq:gcpca_trace_ratio}, we can see that gcPCA and LDA share the same denominator when $\smash{\WeightBgEach{j} = 1}$ for all $j$.
However, gcPCA and LDA have slightly different numerators, where gcPCA and LDA have within-class and between-class covariance matrices, respectively. 
We can fill the gap between gcPCA and LDA by setting the following optimization problem:
\begin{equation}
\label{eq:ulca}
    \max_{\ProjMat^\top \ProjMat = I_\nLatFeats} \frac{ \tr ( \ProjMat^\top \Cov_0 \ProjMat )}
    {\tr ( \ProjMat^\top \Cov_1 \ProjMat ) },
\end{equation}
\vspace{-5pt}
\begin{equation}
    \Cov_0 = \sum_{j=1}^{c} \WeightTgEach{j} \CovWithinEach{j} + \sum_{j=1}^{c} \WeightBwEach{j} \CovBetweenEach{j} + \RegConst_0 I_\nAttrs,
\end{equation}
\vspace{-12pt}
\begin{equation}
    \Cov_1 = \sum_{j=1}^{c} \WeightBgEach{j} \CovWithinEach{j}  + \RegConst_1 I_\nAttrs.
    \vspace{-3pt}
\end{equation}
where $\smash{\CovBetweenEach{j}}$ is a between-class covariance matrix related to group $j$, i.e., \scalebox{0.9}{$\smash{\CovBetweenEach{j} = (\sum_{i=1}^{n} \delta_{\Label_i}^{\scriptscriptstyle j})^{-1} \sum_{i=1}^{n} \delta_{\Label_i}^{\scriptscriptstyle j} (\MeanVec_{\Label_i} - \MeanVec)(\MeanVec_{\Label_i} - \MeanVec)^\top}$};   
$\smash{\WeightBwEach{j}}$ ($\smash{0 \leq \WeightBwEach{j} \leq 1}$) is the $j$-th element of a vector $\smash{\WeightBw}$; $\smash{\RegConst_0}$ and $\smash{\RegConst_1}$ are non-negative numbers.

We use $\smash{\RegConst_0}$ and $\smash{\RegConst_1}$ to avoid the case where either matrix trace of $\smash{\Cov_0}$ or $\smash{\Cov_1}$ is always zero. 
While \name{} uses $\smash{\RegConst_0 = 0}$ and $\smash{\RegConst_1 = 0}$ by default, $\smash{\RegConst_0 = 1}$ is used when \scalebox{0.9}{$\smash{\sum_{j=1}^{c} \WeightTgEach{j} \CovWithinEach{j} + \sum_{j=1}^{c} \WeightBwEach{j} \CovBetweenEach{j} = 0}$} and $\smash{\RegConst_1 = 1}$ is used when \scalebox{0.9}{$\smash{\sum_{j=1}^{c} \WeightBgEach{j} \CovWithinEach{j} = 0}$}.
With this way, \autoref{eq:ulca} can handle either case where both $\smash{\WeightTg}$ and $\smash{\WeightBw}$ are zero vectors or $\smash{\WeightBg}$ is a zero vector. 
For example, when $\smash{\WeightTg = (1, 0, \ldots, 0)}$, $\smash{\WeightBg = \textbf{0}}$, $\smash{\WeightBw = \textbf{0}}$, \autoref{eq:ulca} only maximizes the within-class variance of group 1 (i.e., PCA to group 1).
We use $\smash{\RegConst_0}$ and $\smash{\RegConst_1}$ mainly for the above purpose; however, similar to regularized LDA~\cite{guo2007regularized}, these values can be used to add regularization terms for the case when $\smash{\CovWithin}$ is (close to) singular, which is usually caused when $\smash{n \leq d}$. 
Similar to other machine learning algorithms (e.g., linear regression), $\smash{\RegConst_0}$ and $\smash{\RegConst_1}$ can be used to add the bias into $\Cov_0$ and $\Cov_1$ to avoid overfitting.

By using \name{}, we can perform comparative analysis utilizing the strengths of both discriminant analysis and contrastive learning. 
For example, when $\smash{\WeightTg = (1, 0, 0)}$, $\smash{\WeightBg = (0, 1, 1)}$, and $\smash{\WeightBw = (1, 1, 1)}$, \name{} produces the result where group 1's variance and a distance between each group are maximized while the other groups' variances are minimized. 
We demonstrate concrete analysis examples in \autoref{sec:case_studies}.

Similar to the discussion in \autoref{sec:trace_ratio_form_gcpa}, a relaxed ULCA, where $\ProjMat$ is found with the user-specified $\ContParam$, also can be written as:
\begin{equation}
    \label{eq:ulca_trace_diff}
    \max_{\ProjMat^\top \ProjMat = I_\nLatFeats} \tr \bigl( \ProjMat^\top (\Cov_0 - \ContParam \Cov_1) \ProjMat \bigr). 
\end{equation}
After obtaining $\ProjMat$ by solving \autoref{eq:ulca} or \autoref{eq:ulca_trace_diff}, embedding result $\ReprMat$ can be generated from the original dataset $\OrgMat$ with $\ReprMat = \OrgMat \ProjMat$.

In summary, the optimization problems in \autoref{eq:ulca} and \ref{eq:ulca_trace_diff} are designed for \name{}, where the problems of gcPCA and LDA are integrated. 
Consequently, \name{} encompasses functionalities of PCA, cPCA, ccPCA, gcPCA, and (regularized) LDA while enabling flexible comparative analysis using discriminant analysis and contrastive learning together.

\vspace{-1pt}
\subsubsection{Optimization}
\label{sec:optimization}
\vspace{-1pt}

We explain two approaches to solve \autoref{eq:ulca} and \ref{eq:ulca_trace_diff}: using eigenvalue decomposition (EVD) and manifold optimization~\cite{townsend2016pymanopt}.

\vspace{2pt}
\noindent
\textbf{EVD-based approach.}
EVD is commonly used to solve the trace-difference problem~\cite{jia2009trace}. 
$\ProjMat$ that satisfies \autoref{eq:ulca_trace_diff} can be obtained by first applying EVD to $\smash{(\Cov_0 - \ContParam \Cov_1)}$ and then taking the top $\smash{\nLatFeats}$ eigenvectors. 

To heuristically solve the trace-ratio problem of \autoref{eq:ulca}, we can use an iterative algorithm due to the work by Dinkelbach~\cite{dinkelbach1967nonlinear}.
The algorithm consists of two steps. 
Given $\smash{\ProjMat_t}$ at iteration step $t$, we perform
\vspace{-2pt}
\begin{align}
    \mathrm{\mathbf{Step 1.}}\ \ \ \  &  \alpha_t \gets \dfrac{\tr(\ProjMat^\top_t \Cov_0 \ProjMat_t)}{\tr(\ProjMat^\top_t \Cov_1 \ProjMat_t)},
    \label{eq:iter_step1}
    \\[-1.2ex]
    \mathrm{\mathbf{Step 2.}}\ \ \ \  &  \ProjMat_{t+1} \gets \arg\max\limits_{\ProjMat^\top \ProjMat = I_\nLatFeats} \tr(\ProjMat^\top (\Cov_0 -\alpha_t \Cov_1) \ProjMat).
    \label{eq:iter_step2}
\end{align}
At $t=0$, because the computed $\smash{\ProjMat_0}$ does not exist, we self-define $\smash{\ContParam_0 = 0}$ as a default solution to Step 1. 
As demonstrated, $\smash{\ContParam_t}$ in \autoref{eq:iter_step1} is an objective value of \autoref{eq:ulca}, which is computed with the current $\smash{\ProjMat_t}$.
The second step (\autoref{eq:iter_step2}) is to derive  $\smash{\ProjMat_{t+1}}$ for the next iteration.
This step just solves the relaxed problem (\autoref{eq:ulca_trace_diff}) based on the current parameter $\smash{\ContParam_t}$ with EVD. 
With this iterative algorithm, $\alpha_t$ monotonically increases till `$\smash{\max \tr(\ProjMat^\top (\Cov_0 -\alpha_t \Cov_1) \ProjMat)}$' reaches approximately zero and usually converges quickly (e.g., in less than 10 iterations)~\cite{fujiwara2020interpretable,jia2009trace}. 

\vspace{0pt}
\noindent
\textbf{Manifold-optimization-based approach.} Another way to solve \autoref{eq:ulca} and \autoref{eq:ulca_trace_diff} is using a generic solver designed for optimization over manifolds (or often called manifold optimization)~\cite{absil2009optimization}.
According to Cunningham and Ghahramani~\cite{cunningham2015linear}, linear DR can be considered as solving a manifold optimization problem. 
We can also directly solve both \autoref{eq:ulca} and \autoref{eq:ulca_trace_diff} with a manifold optimization solver available in existing libraries, such as Manopt~\cite{boumal2014manopt}.
More specifically, using the Riemannian Trust Regions (RTR) method~\cite{absil2007trust} over the Grassmann manifold can be used as a solver to achieve the best performance due to the evaluation by Cunningham and Ghahramani~\cite{cunningham2015linear}.
Refer to their work~\cite{cunningham2015linear} for details of manifold optimization and solvers. 

When compared with the EVD-based approach, manifold optimization has two main benefits. 
First, manifold optimization is generic for linear DR methods, including methods that cannot be represented as the trace-ratio or trace-difference problem.
As a result, when adding some enhancement to \name{} in the future, we just need to design a new optimization problem and do not need to find a solution specific to the new problem.
Also, while EVD captures more information in the top eigenvectors than others (e.g., the first eigenvector preserves the original data variance than the second eigenvector), manifold optimization equally treats each embedding axis.
This is especially beneficial when we use DR for a visualization purpose as we can fully utilize a 2D space to convey the preserved information.
Therefore, our implementation of \name{}, by default, uses manifold optimization with the RTR method~\cite{absil2007trust} while we provide the EVD-based approach as an option. 

\name{}'s embedding axes (i.e., columns of $\ProjMat$) obtained via manifold optimization are always orthogonal but could have a different rotation at every execution (even with the same dataset and parameters).
To produce consistent axes, by default, we apply the varimax rotation~\cite{kaiser1958varimax} to $\ProjMat$.
This rotates $\ProjMat$ to minimize the number of attributes that have a high contribution to the axes; thus, it can also improve the interpretability of axes, as often used in factor analysis~\cite{jolliffe1986principal}.
However, the varimax rotation still does not produce consistent axes in terms of their order and sign.
Thus, we adjust the signs to make each column sum of $\ProjMat$ positive, and reorder the axes by the maximum column value of $\ProjMat$.

\vspace{-1pt}
\subsection{Parameter Selection}
\vspace{-1pt}

\name{} allows analysts to adjust multiple important parameters: $\smash{\WeightTg}$, $\smash{\WeightBg}$, $\smash{\WeightBw}$, and $\ContParam$.
Here, we provide general guidance on how to chose desired parameters. 
Also, as a convenient way to select parameters, we introduce the backward parameter selection, which finds the best parameters to resemble the changes indicated in an embedding result. 

\begin{figure}[tb]
    \centering
    \includegraphics[width=0.99\linewidth]{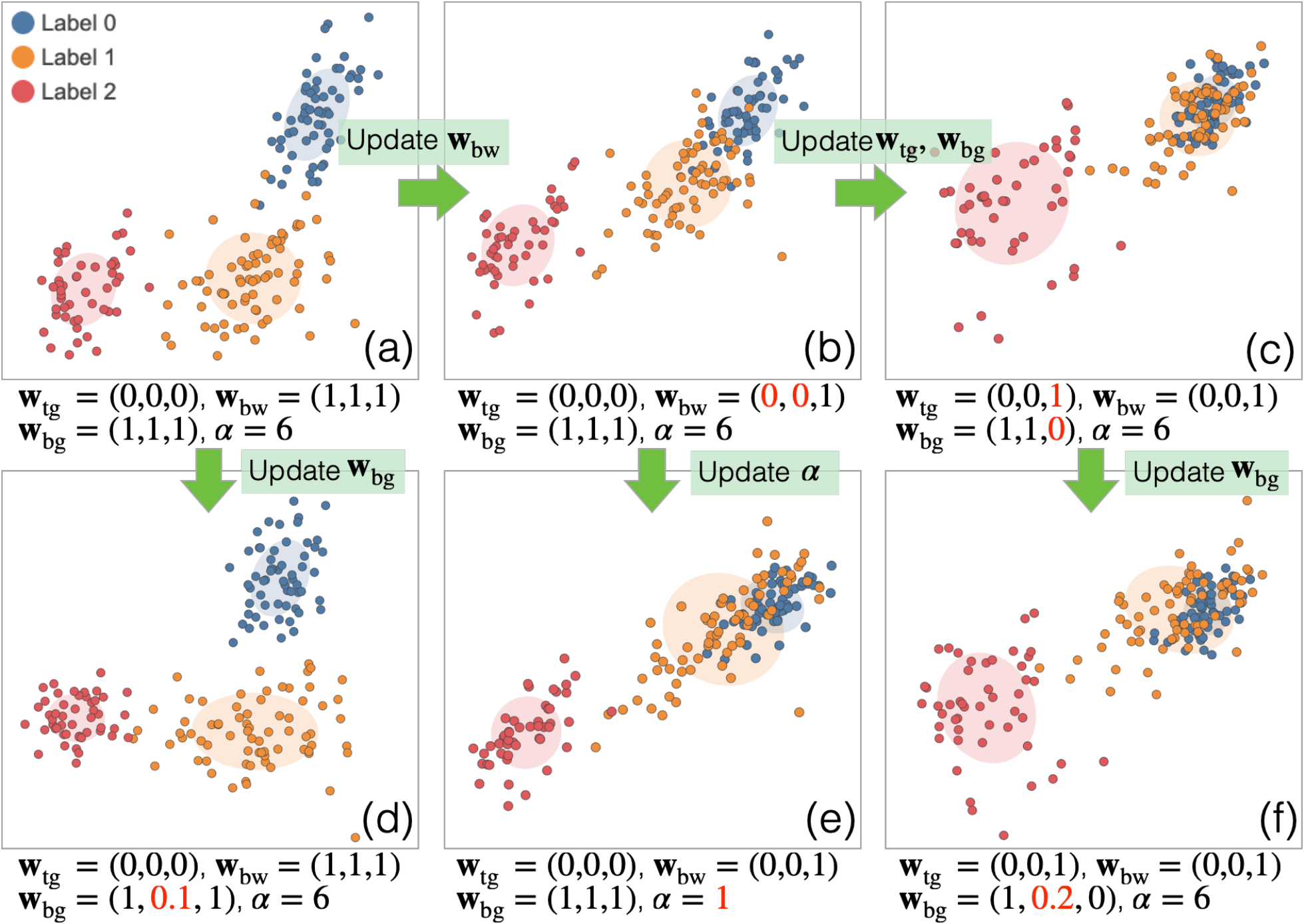}
    \caption{Parameter adjustment examples with the Wine dataset. The updated parameters in each embedding are highlighted in red.}
    \label{fig:param_select}
\end{figure}

\vspace{-2pt}
\subsubsection{General Guidance}
\vspace{-2pt}

A general rule of parameter selection is simple. When we want to observe higher variances in some groups, for the corresponding groups, we should set larger weights in $\smash{\WeightTg}$ and smaller weights in $\smash{\WeightBg}$, and vice versa.
When we want to increase a distance between groups, we should assign larger values to the corresponding weights in $\smash{\WeightBw}$.
$\ContParam$ can be selected automatically by solving \autoref{eq:ulca}.
However, as demonstrated in the analyses using cPCA by Abid and Zhang et al.~\cite{abid2018exploring}, using a different $\ContParam$ in \autoref{eq:ulca_trace_diff} may lead to the discovery of latent patterns that could not be found with an automatically selected $\ContParam$ value.
$\ContParam$ can be used to control how much the background variance should be reduced in an embedding result (i.e., larger $\ContParam$, smaller background variance). 

In \autoref{fig:param_select}, we demonstrate several examples using different parameters. 
\autoref{fig:param_select}a uses parameters that produce the same result with LDA and $\ContParam$ is automatically selected. 
From \autoref{fig:param_select}a, for example, by decreasing $\smash{\WeightBgEach{2}}$, we can weaken the reduction of the orange group's variance relative to others, as shown in \autoref{fig:param_select}d. 
Another example in \autoref{fig:param_select}b is generated by reducing $\smash{\WeightBwEach{1}}$ and $\smash{\WeightBwEach{2}}$; consequently, only the red group has clear separation from others. 
\autoref{fig:param_select}e uses smaller $\ContParam$ than \autoref{fig:param_select}b, and it changes the variance relationships.
In \autoref{fig:param_select}c, $\smash{\WeightTgEach{3}}$ is increased and $\smash{\WeightBgEach{3}}$ is reduced. 
As a result, the red group gets a much larger variance. 
Lastly, we set small $\smash{\WeightBgEach{2}}$ relative to $\smash{\WeightBgEach{1}}$ in \autoref{fig:param_select}f. 
The orange group's variance is now larger than that of the blue group.

\vspace{-2pt}
\subsubsection{Backward Parameter Selection}
\label{sec:param_optim}
\vspace{-1pt}

As discussed above, once we know how each parameter influences an embedding result, it is not difficult to select proper parameters based on analysis purposes. 
However, to further aid parameter selection, we develop a backward algorithm that finds parameters to resemble a user-demonstrated change in a visualized result~\cite{saket2017visualization}. 
The backward algorithm supports two types of changes or interactions: (1) moving a group centroid and (2) scaling a group variance in an embedding result. 
These two interactions are closely connected to possible user intents~\cite{self2016bridging,self2018observation,wenskovitch2020respect} in comparative analysis. 
The first interaction infers that the analyst wants to change distance relationships among groups, for example, to find better group separation. 
With the second interaction, the analyst intends to change relative variances among groups, for example, to increase a certain group's variance relative to the others.   

\begin{figure}[tb]
    \centering
    \includegraphics[width=0.99\linewidth]{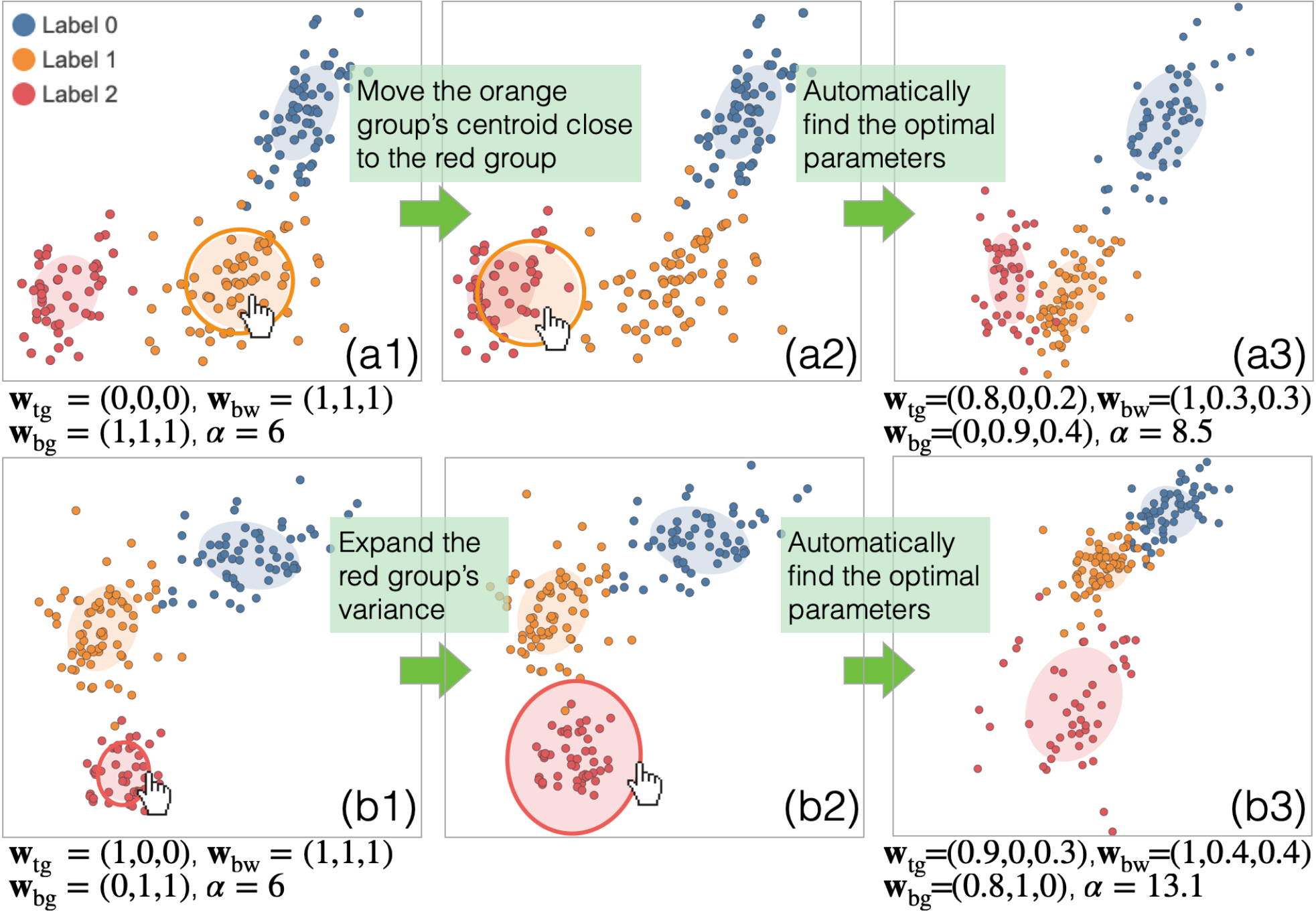}
    \caption{Backward parameter selection on the Wine dataset when moving a group centroid (a1--a3) and scaling a confidence ellipse (b1--b3).}
    \label{fig:backward_param_select}
\end{figure}

\autoref{fig:backward_param_select} shows examples of the backward parameter selection when the two interactions are performed. 
For scaling of a group variance, instead of adjusting each group variance directly, we change the variance through uniform-scaling of the confidence ellipse as it is commonly used to visualize the scatteredness of data points. 
After either interaction is performed, the backward algorithm should find parameters so that both relative centroid distances and relative variance differences of each group in the demonstrated changes (e.g., \autoref{fig:backward_param_select}a2, b2) are preserved as much as possible. 
Such optimization can be written as:
\begin{equation}
\label{eq:backward_algorithm}
    \min_\Params ~ \WeightDist \CostDist + \WeightArea \CostArea
\end{equation}
\vspace{-3pt}
\begin{equation}
\label{eq:cost_dist}
    \CostDist = \sqrt{\sum_{1 \leq i, j \leq c} (\DistIdeal{i,j} - \DistNew{i,j})^2 \Big/ \sum_{1 \leq i,j \leq c} \DistIdeal{i,j}^2}
\end{equation}
\vspace{-8pt}
\begin{equation}    
\label{eq:cost_area}
    \CostArea = \frac{1}{c} \sum_{i=1}^{c} \biggl( \Big| \frac{\AreaIdeal{\TargetClass}}{\AreaIdeal{i}} - \frac{\AreaNew{\TargetClass}}{\AreaNew{i}} \bigg| \ \Big/  \frac{\AreaIdeal{\TargetClass}}{\AreaIdeal{i}} \biggr) 
    \vspace{-3pt}
\end{equation}
where $\Params$ consists of parameters, $\smash{\WeightTg}$, $\smash{\WeightBg}$, $\smash{\WeightBw}$, and $\ContParam$; $\smash{\CostDist}$ and $\smash{\CostArea}$ ($\smash{0 \leq \CostDist, \CostArea \leq 1}$) are cost functions related to the distance and variance relationships, respectively; $\TargetClass$ is a label of moved/scaled group ($1 \leq k \leq c$); $\smash{\WeightDist}$ and $\smash{\WeightArea}$ ($\smash{\WeightDist + \WeightArea = 1}$, $\smash{\WeightDist, \WeightArea \geq 0}$) are parameters that control the weights of $\smash{\CostDist}$ and $\smash{\CostArea}$ to the total cost. And, $0 \leq \WeightDist \CostDist + \WeightArea \CostArea \leq 1$.

$\smash{\CostDist}$ can be computed with the user-indicated centroid distances \scalebox{0.9}{$\smash{\DistIdeal{i,j}}$} ($\smash{i,j = \{1, \ldots, c\}}$) and new centroid distances \scalebox{0.9}{$\smash{\DistNew{i,j}}$}, which are generated with parameters $\Params$.  
As shown in \autoref{eq:cost_dist}, we follow classical MDS's \textit{strain}~\cite{torgerson1952} to design $\smash{\CostDist}$, where the root sum squared (RSS) of distance differences is computed and scaled  to a range of $[0, 1]$.
We compute $\smash{\CostArea}$ with $\smash{\AreaIdeal{i}}$ ($\smash{i = \{1, \ldots, c \}}$), the areas of the user-indicated confidence ellipses, and $\smash{\AreaNew{i}}$, the areas of the confidence ellipses in a new embedding space.
In \autoref{eq:cost_area}, we first compute $\smash{\AreaIdeal{\TargetClass} / \AreaIdeal{i}}$ and $\smash{\AreaNew{\TargetClass} / \AreaNew{i}}$ (the moved/scaled group's area ratios to the other) and then take their sum of absolute differences (SAD). 
Afterward, we scale the SAD to a range of $[0, 1]$.
From $\smash{\CostDist}$ and $\smash{\CostArea}$, we compute the total cost with $\smash{\WeightDist}$ and $\smash{\WeightArea}$.
We can set different $\smash{\WeightDist}$ and $\smash{\WeightArea}$ for each interaction. 
When moving a centroid, we expect that the analyst mainly wants to refine the distance relationships, and thus, we use $\smash{\WeightDist = 0.8}$ and $\smash{\WeightArea = 0.2}$ by default. 
On the other hand, for the scaling, we use $\smash{\WeightDist = 0.2}$ and $\smash{\WeightArea = 0.8}$.

Now, we need to solve \autoref{eq:backward_algorithm} to find the optimal $\Params$.
Since \scalebox{0.9}{$\smash{\DistNew{i,j}}$} and $\smash{\AreaNew{i}}$ are obtained after solving \autoref{eq:ulca_trace_diff}, this problem is optimizing $\Params$ over the optimization of \autoref{eq:ulca_trace_diff}. 
As a result, it may be difficult to find a direct solution, such as using EVD, or to derive a gradient. 
Thus, we use a generic gradient-free solver. 
Also, we need to constrain all the weights of $\smash{\WeightTg}$, $\smash{\WeightBg}$, $\smash{\WeightBw}$ within a range of $[0, 1]$ and $\ContParam$ within $[0, \infty)$.
To satisfy these requirements, we select COBYLA (or constrained optimization by linear approximations)~\cite{powell1998direct}.
For each iteration of the optimization with COBYLA, we solve \autoref{eq:ulca_trace_diff} with given $\Params$ and compute the cost in \autoref{eq:backward_algorithm}. 
Then, COBYLA selects a new $\Params$ from a trust region for the next iteration. 
COBYLA keeps iterating these steps until reaching the convergence or specified maximum number of iterations. 
We provide an evaluation of the optimization using COBYLA in \autoref{sec:perf_eval}.

\vspace{-2pt}
\subsection{Complexity Analysis}
\label{sec:complexity}
\vspace{-2pt}

We first analyze the time complexity of \name{} when using the EVD-based approach.
If $\smash{c \ll n, d}$, which is a reasonable assumption for a practical usage, the relaxed version of \name{} in \autoref{eq:ulca_trace_diff} has two major calculations: covariance matrix calculation ($\smash{\mathcal{O}(\nInsts \nAttrs^2)}$) and EVD ($\smash{\mathcal{O}(\nAttrs^3)}$). 
Thus, the relaxed version of \name{} has $\smash{\mathcal{O}(\nInsts \nAttrs^2 + \nAttrs^3)}$ time complexity, which is compatible with PCA and cPCA. 
As mentioned, solving \autoref{eq:ulca} with the iterative algorithm converges quickly. 
Thus, the complexity of the non-relaxed version of \name{} is similar to $\smash{\mathcal{O}(\nInsts \nAttrs^2 + \nAttrs^3)}$ .

After the covariance matrix calculation ($\smash{\mathcal{O}(\nInsts \nAttrs^2)}$), the manifold-optimization-based approach iteratively solves \autoref{eq:ulca} or \autoref{eq:ulca_trace_diff} with a manifold optimization solver.
Each iteration performs matrix multiplication of $\ProjMat$ and covariance matrices, which has the time complexity of $\smash{\mathcal{O}(\nLatFeats \nAttrs^2)}$, and then computes partial derivatives for the next iteration.
When using the RTR method~\cite{absil2007trust}, at each iteration, the second-order partial derivatives (or the Hessian matrix) are computed with $\smash{\mathcal{O}(\nLatFeats^2 \nAttrs^2)}$ time complexity.
When $\smash{\nLatFeats \ll \nAttrs}$, we can consider both of the above time complexities are $\smash{\mathcal{O}(\nAttrs^2)}$.
Now, the time complexity depends on the number of iterations till the convergence. 
Based on the performance evaluation by Cunningham and Ghahramani~\cite{cunningham2015linear}, the manifold optimization runtime for various linear DR methods follows approximately three orders of $\nAttrs$. 
From this, we can consider that the manifold-optimization-based approach practically has a similar runtime to the EVD-based approach.
However, in the worst case, as studied by Boumal \etal~\cite{boumal2019global}, the RTR method needs to iterate $\smash{\mathcal{O}(1 / \epsilon^3)}$ where $\epsilon$ ($0 \leq \epsilon \leq 1$) is the convergence threshold.
To avoid a numerous number of iterations, we can also set the maximum number of iterations. 
For example, to complete the optimization in similar runtime with the EVD-based approach, we can use $\nAttrs$ as the maximum number of iterations.

The backward parameter selection in \autoref{sec:param_optim} solves \autoref{eq:ulca_trace_diff} at each iteration with parameters $\Params$ selected by COBYLA~\cite{powell1998direct}.
Thus, the time complexity is the multiplication of the user-specified maximum number of iterations and \name{}'s time complexity. 

\vspace{-1pt}
\section{Visual Interface}
\label{sec:visual_interface}
\vspace{-1pt}

Our framework provides a visual interface to visualize and interact with a \name{} result.
As shown in \autoref{fig:ui}, the interface provides three views with simple visualizations: (a) a parameter view, where bar charts display parameters of $\smash{\WeightTg}$, $\smash{\WeightBg}$, $\smash{\WeightBw}$, and $\ContParam$; (b) an embedding result view, where a scatterplot depicts an embedding result with the confidence ellipse of each group; (c) a component view, where bar charts show the information of the axes of the embedding result view. 
All views are fully linked, and each is updated based on an interaction performed in the other views. 

\vspace{2pt}
\noindent
\textbf{Supported interactions.} 
In the parameter view (\autoref{fig:ui}a), the analyst can adjust each parameter by changing the corresponding bar length. 
This interaction instantly reruns \name{} in \autoref{eq:ulca_trace_diff}. 
In the embedding result view (\autoref{fig:ui}b), the analyst can move each group's centroid and scale each confidence ellipse by dragging the corresponding confidence ellipse's center and outline, respectively. 
These changes induce the backward parameter selection. 
The analyst also can draw a new axis, as\,demonstrated\,with\,axis\,\textcircled{\small 1}\,in \autoref{fig:ui}b,\,to\,see\,how\,each\,attribute contributes\,to\,directions\,of\,interest.\,A\,linear\,mapping\,from\,the\,original attributes\,to\,the\,new\,axis\,can\,be\,computed\,with $\smash{\ProjMat \mathbf{v} / ||\mathbf{v}||}$ where $\mathbf{v}$ is a vector of the new axis.
Then, this information is added to the component view.
When hovering over a certain attribute name in the component view (\autoref{fig:ui}c), to see the distribution of the attribute values,  each point's size in the embedding result view is updated based on its attribute value.

\vspace{2pt}
\noindent
\textbf{Mental map preservation.} 
When updating an embedding, linear DR methods that use EVD or manifold optimization, including \name{}, cause arbitrary sign flipping and rotations of embedding axes~\cite{fujiwara2019incremental}. 
Thus, an embedding result would be drastically changed, and it would be difficult to follow the changes as the analyst easily loses their mental map~\cite{purchase2006important}.
To mitigate this problem, we take an approach similar to one developed to use linear DR in a streaming setting~\cite{fujiwara2019incremental} (note that several works addressed mental map preservation for nonlinear DR~\cite{rauber2016dynamic,cantareira2020generic}).
We use the rotation matrix obtained through the Procrustes analysis~\cite{gower2004procrustes} of the previous and new embedding results.
This rotation matrix adjusts the new embedding axes' signs and rotations to minimize the difference of data positions from the previous result.
Furthermore, we animate the changes in each view to help the analyst maintain their mental map.

\vspace{2pt}
\noindent
\textbf{Provenance support.} During the interactive analysis, the analyst may find interesting patterns and may want to record the (intermediate) analysis results to further investigate the results by comparing with other results later, apply the obtained embedding to other data, or share the results with others. 
To help the analyst keep the history of changes during the analysis (or called \textit{provenance}~\cite{ragan2016characterizing}), we provide a saving function. 
The analyst can name the result and save it through the text input field and the `save' button placed at \autoref{fig:ui}d. 
This function saves all necessary information to recover both visualizations and \name{} results.
The analyst can go back to the saved result by selecting the corresponding name from the drop-down list of the saved names (the left side of \autoref{fig:ui}d). 
Also, the saved results can be directly referred from Python and the Jupyter Notebook due to the seamless integration of the interface with them, as described below. 

\begin{figure}[tb]
    \centering
    \includegraphics[width=\linewidth]{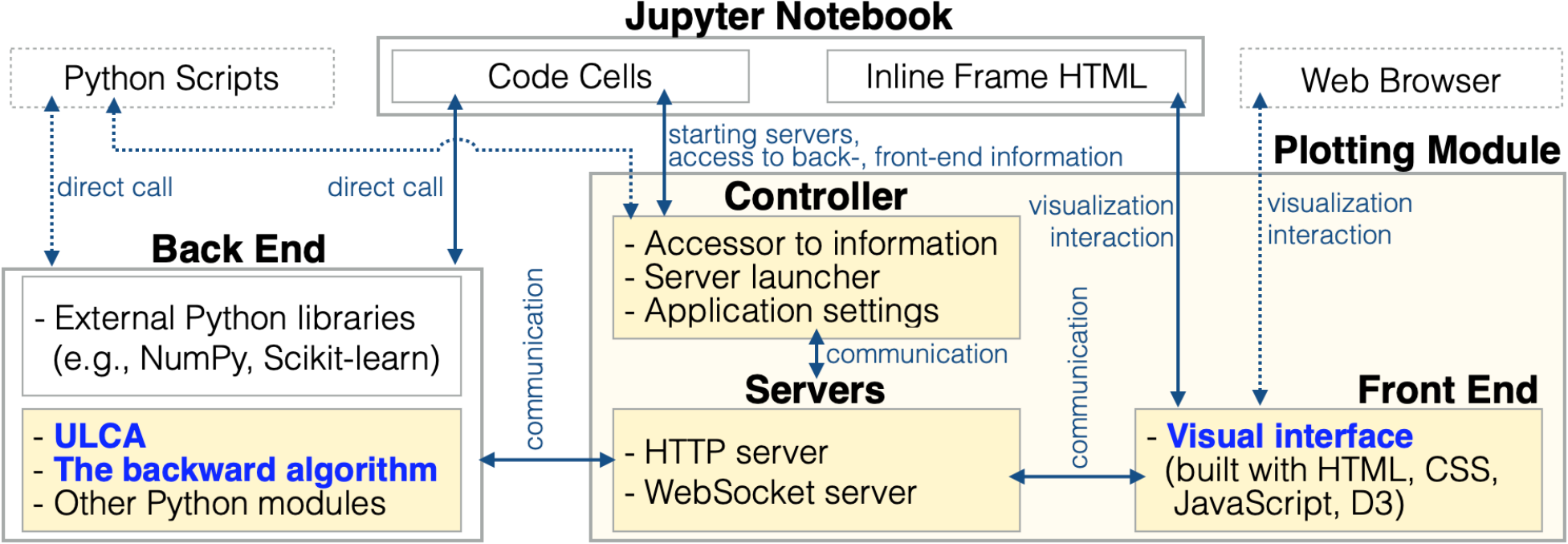}
    \caption{
    The system architecture of the framework. The yellow boxes are the implemented modules for the framework. 
    }
    \label{fig:system_architecture}
\end{figure}

\vspace{-2pt}
\subsection{Implementation}
\vspace{-1pt}

The knowledge discovery process often requires many different analytical components~\cite{fayyad1996data,keim2008visual}.
To allow for use of \name{} together with all necessary algorithms, visualizations, and interactions, 
we design the visual interface to be seamlessly used with Python and the Jupyter Notebook~\cite{jupyter}.
In this way, when analyzing datasets with our framework, the analyst can utilize various existing analysis and visualization libraries, such as scikit-learn~\cite{pedregosa2011scikit} (for machine learning), Matplotlib~\cite{hunter2007matplotlib} (for visualization), or their own analysis methods. 
Also, the analyst can employ various analytical processing, such as assigning new labels to data points and sorting the attributes in the component view based on their contributions to each embedding axis.
In \autoref{fig:ui}, we visualize the result obtained by applying \name{} to the processed dataset in the Jupyter Notebook, interact with the result in the visual interface, and access the information gained through interactive analysis.  

\autoref{fig:system_architecture} shows our framework's system architecture.
While the front end is implemented as a web-based application with JavaScript and D3~\cite{bostock2011d3}, the other parts are implemented with Python. 
For the implementation of optimization solvers, the framework uses Pymanopt~\cite{townsend2016pymanopt} and SciPy~\cite{virtanen2020scipy}. 
To make the front end callable from the Jupyter Notebook's \textit{code cells} where we can write and run Python scripts, we implement a plotting module that consists of a controller, servers, and the front end.

The analyst can use the controller to adjust settings and call the visual interface, as demonstrated in `\texttt{In\,[5]}' code cell in \autoref{fig:ui}. 
Then, the controller starts HTTP and WebSocket servers in localhost to establish communications.
Afterward, the visual interface is shown by using an HTML inline frame, as shown in `\texttt{Out\,[5]}' in \autoref{fig:ui}. 
All the information shown in the visual interface is stored in the plotting module as Python objects, and can be accessed via the controller.
For example, `\texttt{In\,[6]}' in \autoref{fig:ui} refers to $x$-axis information of the current result shown in `\texttt{Out\,[5]}' and that of the saved result named with `\texttt{PCA on Label 1}'.
The controller provides an option to show the visual interface in an individual webpage (the dashed lines and boxes in \autoref{fig:system_architecture}). 
The framework also can be used without using the Jupyter Notebook.

\vspace{-3pt}
\section{Performance Evaluation}
\label{sec:perf_eval}
\vspace{-1pt}

We evaluate the performance of \name{} and the backward parameter selection.
As an experimental platform, we use the MacBook Pro (16-inch, 2019) with 2.3 GHz 8-Core Intel Core i9 and 64 GB 2,667 MHz DDR4. 
The experiment details are available online~\cite{supp}.

\vspace{2pt}
\noindent
\textbf{Data.}
We generate datasets with various numbers of data points, attributes, and groups. 
From\,the\,documents\,(i.e., data points) in the 20 Newsgroups dataset~\cite{uci_mlr}, we extract arbitrary numbers of topics (i.e., attributes) by utilizing the Latent Dirichlet Allocation~\cite{blei2003latent}. 
To obtain various numbers of data points and groups, we randomly sample documents and apply $k$-means clustering~\cite{hartigan1979algorithm} to the sampled documents.

\begin{figure}[t]
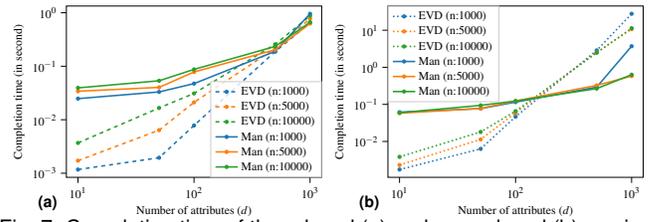

    \centering
    \begin{subfigure}[t]{0.475\linewidth}
        \begin{adjustbox}{width=\linewidth}
\begin{tikzpicture}

\input{plots/plot_adjustment}
\node at (-0.5, -0.7) {\textbf{\textsf{\Large (a)}}};

\definecolor{color0}{rgb}{0.12156862745098,0.466666666666667,0.705882352941177}
\definecolor{color1}{rgb}{1,0.498039215686275,0.0549019607843137}
\definecolor{color2}{rgb}{0.172549019607843,0.627450980392157,0.172549019607843}

\begin{axis}[
width=\axisdefaultwidth,
height=0.85*\axisdefaultheight,
legend cell align={left},
legend style={
  fill opacity=0.8,
  draw opacity=1,
  text opacity=1,
  at={(1.0,0.0)},
  anchor=south east,
  draw=white!80!black
},
log basis x={10},
log basis y={10},
tick align=outside,
tick pos=left,
x grid style={white!69.0196078431373!black},
xlabel={Number of attributes (\(\displaystyle d\))},
xmin=7.94328234724282, xmax=1258.92541179417,
xmode=log,
xtick style={color=black},
xtick={0.1,1,10,100,1000,10000,100000},
xticklabels={
  \(\displaystyle {10^{-1}}\),
  \(\displaystyle {10^{0}}\),
  \(\displaystyle {10^{1}}\),
  \(\displaystyle {10^{2}}\),
  \(\displaystyle {10^{3}}\),
  \(\displaystyle {10^{4}}\),
  \(\displaystyle {10^{5}}\)
},
y grid style={white!69.0196078431373!black},
ylabel={Completion time (in second)},
ymin=0.000837773138479868, ymax=1.33607619073583,
ymode=log,
ytick style={color=black},
ytick={1e-05,0.0001,0.001,0.01,0.1,1,10,100},
yticklabels={
  \(\displaystyle {10^{-5}}\),
  \(\displaystyle {10^{-4}}\),
  \(\displaystyle {10^{-3}}\),
  \(\displaystyle {10^{-2}}\),
  \(\displaystyle {10^{-1}}\),
  \(\displaystyle {10^{0}}\),
  \(\displaystyle {10^{1}}\),
  \(\displaystyle {10^{2}}\)
}
]
\addplot [very thick, color0, dashed, mark=*, mark size=3, mark options={solid}]
table {%
10 0.0011713964999998
50 0.0019478446999997
100 0.0078072720000001
500 0.187839930299999
1000 0.649588143399998
};
\addlegendentry{EVD (n:1000)}
\addplot [very thick, color1, dashed, mark=*, mark size=3, mark options={solid}]
table {%
10 0.0017155440000003
50 0.0063807505999989
100 0.0211167428000003
500 0.216320875799999
1000 0.775196269599999
};
\addlegendentry{EVD (n:5000)}
\addplot [very thick, color2, dashed, mark=*, mark size=3, mark options={solid}]
table {%
10 0.0036851503999997
50 0.0166010057000008
100 0.0310110564999973
500 0.2572197184
1000 0.874548407599997
};
\addlegendentry{EVD (n:10000)}
\addplot [very thick, color0, mark=*, mark size=3, mark options={solid}]
table {%
10 0.0248119696999999
50 0.0330850641000004
100 0.0473462502999993
500 0.189700846799999
1000 0.955550698300001
};
\addlegendentry{Man (n:1000)}
\addplot [very thick, color1, mark=*, mark size=3, mark options={solid}]
table {%
10 0.0340688223999983
50 0.0402411102000002
100 0.0781642849
500 0.203441041400001
1000 0.628029037700001
};
\addlegendentry{Man (n:5000)}
\addplot [very thick, color2, mark=*, mark size=3, mark options={solid}]
table {%
10 0.0393897011999982
50 0.0534920947000003
100 0.0873279041000032
500 0.234495619500001
1000 0.665290380399998
};
\addlegendentry{Man (n:10000)}
\end{axis}

\end{tikzpicture}
        \end{adjustbox}
    \end{subfigure}
    \begin{subfigure}[t]{0.475\linewidth}
        \begin{adjustbox}{width=\linewidth}
\begin{tikzpicture}

\input{plots/plot_adjustment}
\node at (-0.5, -0.7) {\textbf{\textsf{\Large (b)}}};

\definecolor{color0}{rgb}{0.12156862745098,0.466666666666667,0.705882352941177}
\definecolor{color1}{rgb}{1,0.498039215686275,0.0549019607843137}
\definecolor{color2}{rgb}{0.172549019607843,0.627450980392157,0.172549019607843}

\begin{axis}[
width=\axisdefaultwidth,
height=0.85*\axisdefaultheight,
legend cell align={left},
legend style={
  fill opacity=0.8,
  draw opacity=1,
  text opacity=1,
  at={(0.0,1.0)},
  anchor=north west,
  draw=white!80!black
},
log basis x={10},
log basis y={10},
tick align=outside,
tick pos=left,
x grid style={white!69.0196078431373!black},
xlabel={Number of attributes (\(\displaystyle d\))},
xmin=7.94328234724282, xmax=1258.92541179417,
xmode=log,
xtick style={color=black},
xtick={0.1,1,10,100,1000,10000,100000},
xticklabels={
  \(\displaystyle {10^{-1}}\),
  \(\displaystyle {10^{0}}\),
  \(\displaystyle {10^{1}}\),
  \(\displaystyle {10^{2}}\),
  \(\displaystyle {10^{3}}\),
  \(\displaystyle {10^{4}}\),
  \(\displaystyle {10^{5}}\)
},
y grid style={white!69.0196078431373!black},
ylabel={Completion time (in second)},
ymin=0.00107901875350161, ymax=44.9127214489647,
ymode=log,
ytick style={color=black},
ytick={0.0001,0.001,0.01,0.1,1,10,100,1000},
yticklabels={
  \(\displaystyle {10^{-4}}\),
  \(\displaystyle {10^{-3}}\),
  \(\displaystyle {10^{-2}}\),
  \(\displaystyle {10^{-1}}\),
  \(\displaystyle {10^{0}}\),
  \(\displaystyle {10^{1}}\),
  \(\displaystyle {10^{2}}\),
  \(\displaystyle {10^{3}}\)
}
]
\addplot [very thick, color0, dotted, mark=*, mark size=3, mark options={solid}]
table {%
10 0.0017498427000077
50 0.0063177710000047
100 0.0464817253999967
500 2.8538428397
1000 27.6948714956
};
\addlegendentry{EVD (n:1000)}
\addplot [very thick, color1, dotted, mark=*, mark size=3, mark options={solid}]
table {%
10 0.0023184198000194
50 0.0113845121999929
100 0.0589789300999996
500 2.51406618740001
1000 10.6924467389
};
\addlegendentry{EVD (n:5000)}
\addplot [very thick, color2, dotted, mark=*, mark size=3, mark options={solid}]
table {%
10 0.0038471789000141
50 0.0180627602000072
100 0.0657691612999997
500 2.47003171160001
1000 11.3394712815
};
\addlegendentry{EVD (n:10000)}
\addplot [very thick, color0, mark=*, mark size=3, mark options={solid}]
table {%
10 0.0615687009999931
50 0.076889544699992
100 0.11501538490001
500 0.292040414500013
1000 3.6993734305
};
\addlegendentry{Man (n:1000)}
\addplot [very thick, color1, mark=*, mark size=3, mark options={solid}]
table {%
10 0.0579750097000214
50 0.0772844864000148
100 0.123781149299987
500 0.325321055199981
1000 0.579752862100008
};
\addlegendentry{Man (n:5000)}
\addplot [very thick, color2, mark=*, mark size=3, mark options={solid}]
table {%
10 0.059418838099998
50 0.0940752570999734
100 0.123042725399989
500 0.265897391099997
1000 0.630426402299997
};
\addlegendentry{Man (n:10000)}
\end{axis}

\end{tikzpicture}
        \end{adjustbox}
    \end{subfigure}
    \vspace{-2pt}
    \caption{Completion time of the relaxed (a) and non-relaxed (b) versions of \name{}, using EVD (\texttt{EVD}) and manifold-optimization (\texttt{Man}) approaches.}
     \label{fig:perf_eval_ulca}
\end{figure}

\vspace{2pt}
\noindent
\textbf{Evaluation of ULCA.}
We evaluate the efficiency of the relaxed (\autoref{eq:ulca_trace_diff}) and non-relaxed (\autoref{eq:ulca}) versions of \name{}, using EVD (\texttt{EVD}) and manifold optimization (\texttt{Man}). 
For each different combination of the numbers of data points ($n$) and attributes ($d$), we execute \name{} 10 times with random values of $\WeightTg$, $\WeightBg$, $\WeightBw$ and compute the average completion time. 
We use a fixed number of groups ($c=3$) because it does not have a strong influence on the computational cost.
For the relaxed version of \name{}, we use $\ContParam = 1$.
The result is shown in \autoref{fig:perf_eval_ulca}. 

From \autoref{fig:perf_eval_ulca}, no matter what version and approach are used, the completion time mainly depends on the number of attributes.
However, even for a large number of attributes, \name{} can be solved quickly. 
For example, the relaxed and non-relaxed versions are solved about 1 second when $d=1,000$ and $d=500$, respectively.
From the comparison of \texttt{EVD} and \texttt{Man}, \texttt{EVD} is faster when the number of attributes is relatively small while \texttt{Man} becomes similar or faster than \texttt{EVD} after $d=500$. 
This tendency is more apparent in \autoref{fig:perf_eval_ulca}b.
As described in \autoref{sec:complexity}, \texttt{EVD}'s time complexity heavily depends on the number of attributes (scale of $d^3$).
\texttt{Man} is less sensitive to it (scale of $d^2$).
One notable point is the dramatic time increase when $n=1,000$ and $d=1,000$ for both \texttt{EVD} and \texttt{Man} in \autoref{fig:perf_eval_ulca}b.
We consider that this is due to the instability of LDA's optimization when $\smash{n \leq d}$, as studied by Guo \etal~\cite{guo2007regularized}. 
Since \name{} inherits the characteristics of LDA, \name{} also suffers from the same problem.
Consequently, the iterative steps used in both \texttt{EVD} and \texttt{Man} cannot reach the convergence quickly. 
To avoid this, similar to the work by Guo \etal~\cite{guo2007regularized}, we can apply regularization with $\RegConst_0$ and $\RegConst_1$.

\begin{figure}[t]
    \centering
    \begin{subfigure}[t]{0.47\linewidth}
        \begin{adjustbox}{width=\linewidth}
\begin{tikzpicture}

\tikzstyle{every plot}=[mark size=1pt]
\node at (-0.5, -0.7) {\textbf{\textsf{\Large (a)}}};

\definecolor{color0}{rgb}{0.890196078431372,0.466666666666667,0.76078431372549}
\definecolor{color1}{rgb}{0.737254901960784,0.741176470588235,0.133333333333333}
\definecolor{color2}{rgb}{0.0901960784313725,0.745098039215686,0.811764705882353}

\begin{axis}[
width=\axisdefaultwidth,
height=0.74*\axisdefaultheight,
legend cell align={left},
legend style={
  fill opacity=0.8,
  draw opacity=1,
  text opacity=1,
  at={(0.0,1.0)},
  anchor=north west,
  draw=white!80!black
},
tick align=outside,
tick pos=left,
x grid style={white!69.0196078431373!black},
xlabel={Maximum number of iterations ($m$)},
xmin=1.25, xmax=83.75,
xtick style={color=black},
y grid style={white!69.0196078431373!black},
ylabel={Completion time (second)},
ymin=-0.0886035384393377, ymax=4.55864959070224,
ytick style={color=black}
]
\addplot [very thick, color0, mark=*, mark size=3, mark options={solid}]
table {%
5 0.202941921425326
10 0.545573107079401
20 1.29251722627298
40 2.14439546172883
80 4.3474108121049
};
\addlegendentry{d:100, c:2}
\addplot [very thick, color1, mark=*, mark size=3, mark options={solid}]
table {%
5 0.20684156129459
10 0.38638157153487
20 0.681873571225084
40 1.78500045163356
80 2.76152384928815
};
\addlegendentry{d:100, c:4}
\addplot [very thick, color2, mark=*, mark size=3, mark options={solid}]
table {%
5 0.170709972919735
10 0.438147081939996
20 0.88588396986106
40 1.83588468559489
80 3.97022424281026
};
\addlegendentry{d:100, c:6}
\addplot [very thick, color0, dashed, mark=*, mark size=3, mark options={solid}]
table {%
5 0.14759288938791
10 0.236247139234278
20 0.61468715432594
40 1.07440097020479
80 2.42908600216434
};
\addlegendentry{d:10, c:2}
\addplot [very thick, color1, dashed, mark=*, mark size=3, mark options={solid}]
table {%
5 0.142187146415048
10 0.246813244384262
20 0.542906605854379
40 0.980197992073877
80 1.97922582359078
};
\addlegendentry{d:10, c:4}
\addplot [very thick, color2, dashed, mark=*, mark size=3, mark options={solid}]
table {%
5 0.122635240158007
10 0.239624921532059
20 0.62488901598383
40 1.01518761201687
80 2.20108279280883
};
\addlegendentry{d:10, c:6}
\end{axis}

\end{tikzpicture}
        \end{adjustbox}
    \end{subfigure}
    \begin{subfigure}[t]{0.48\linewidth}
        \begin{adjustbox}{width=\linewidth}
\begin{tikzpicture}

\tikzstyle{every plot}=[mark size=1pt]

\node at (-0.5, -0.7) {\textbf{\textsf{\Large (b)}}};

\definecolor{color0}{rgb}{0.890196078431372,0.466666666666667,0.76078431372549}
\definecolor{color1}{rgb}{0.737254901960784,0.741176470588235,0.133333333333333}
\definecolor{color2}{rgb}{0.0901960784313725,0.745098039215686,0.811764705882353}

\begin{axis}[
width=\axisdefaultwidth,
height=0.74*\axisdefaultheight,
legend cell align={left},
legend style={
  fill opacity=0.8,
  draw opacity=1,
  text opacity=1,
  at={(1.0,0.0)},
  anchor=south east,
  draw=white!80!black
},
tick align=outside,
tick pos=left,
x grid style={white!69.0196078431373!black},
xlabel={Maximum number of iterations ($m$)},
xmin=1.25, xmax=83.75,
xtick style={color=black},
y grid style={white!69.0196078431373!black},
ylabel={Accuracy},
ymin=0.0718459435054414, ymax=0.962784874243602,
ytick style={color=black},
ytick={0,0.1,0.2,0.3,0.4,0.5,0.6,0.7,0.8,0.9,1},
yticklabels={0.0,0.1,0.2,0.3,0.4,0.5,0.6,0.7,0.8,0.9,1.0}
]
\addplot [very thick, color0, mark=*, mark size=3, mark options={solid}]
table {%
5 0.45335118345224
10 0.263943497719433
20 0.806235306793965
40 0.886045158160727
80 0.92228765011914
};
\addlegendentry{d:100, c:2}
\addplot [very thick, color1, mark=*, mark size=3, mark options={solid}]
table {%
5 0.112343167629903
10 0.416060641512461
20 0.660915689767656
40 0.770112811792365
80 0.865421950058898
};
\addlegendentry{d:100, c:4}
\addplot [very thick, color2, mark=*, mark size=3, mark options={solid}]
table {%
5 0.184034553933027
10 0.345378857992927
20 0.493882485572426
40 0.652008074050254
80 0.918748226844718
};
\addlegendentry{d:100, c:6}
\addplot [very thick, color0, dashed, mark=*, mark size=3, mark options={solid}]
table {%
5 0.329926955731155
10 0.464605163841315
20 0.805649185153396
40 0.840589211348049
80 0.858919692022108
};
\addlegendentry{d:10, c:2}
\addplot [very thick, color1, dashed, mark=*, mark size=3, mark options={solid}]
table {%
5 0.25481392777807
10 0.434902174326793
20 0.549765895672574
40 0.824292734151709
80 0.920055994719629
};
\addlegendentry{d:10, c:4}
\addplot [very thick, color2, dashed, mark=*, mark size=3, mark options={solid}]
table {%
5 0.194208961899597
10 0.318224957088428
20 0.382440318219065
40 0.718842307130843
80 0.860155352896545
};
\addlegendentry{d:10, c:6}
\end{axis}

\end{tikzpicture}
        \end{adjustbox}
    \end{subfigure}
    \vspace{-2pt}
    \caption{Completion time (a) and accuracy (b) of the backward parameter selection with the manifold-optimization-based approach.}
     \label{fig:perf_eval_backward}
\end{figure}
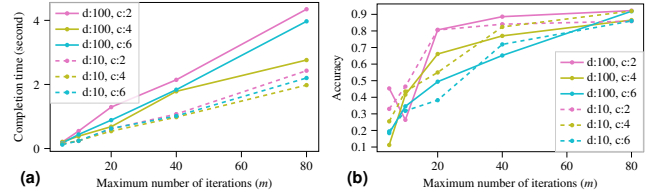

\begin{figure*}[tb]
    \centering
    \includegraphics[width=\linewidth]{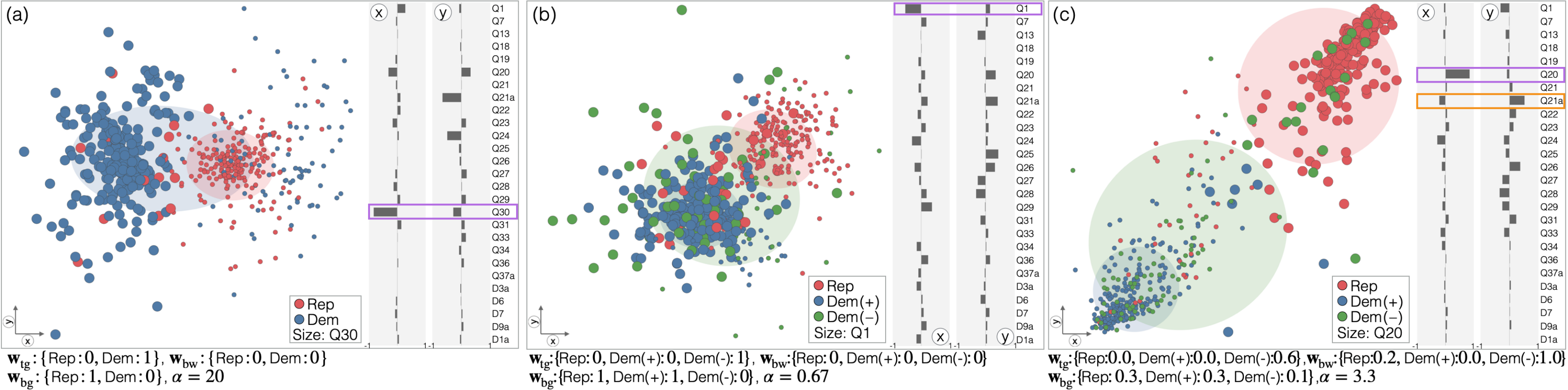}
    \caption{Case Study 1. Each of the results (a--c) is generated with the parameters listed at the bottom. The size of each point represents an answer for the question annotated with the purple box (big circle: `yes', small circle: `no').}
    \label{fig:cs1}
\end{figure*}

\vspace{0pt}
\noindent
\textbf{Evaluation of the backward parameter selection.}
We test the completion time and accuracy of the backward parameter selection. 
We use only the manifold-optimization-based approach and a fixed number of data points (i.e., $n=1,000$) but various values for the maximum number of iterations for COBYLA (here, we represent it with $m$), number of groups $c$, and number of attributes $d$.
Other parameters are set with default values. 
For each setting, we generate an initial embedding result with $\ContParam=1$ and random values of $\WeightTg$, $\WeightBg$, $\WeightBw$.
Then, we mimic the user-demonstrated change in the initial embedding result by randomly selecting a group and an interaction (e.g., moving group 1's centroid to coordinate $(0.2, 0.8)$ or expanding group 2's confidence ellipse $1.2$ times) and the backward parameter selection is performed for this mimicked change. 
We generate 500 sets of changes for each setting and compute the average completion time and accuracy. 
The accuracy is calculated with $\smash{(e_{\textrm{init}} - e) / (e_{\textrm{init}} - e_{\textrm{opt}})}$ where $\smash{e_{\textrm{init}}}$, $\smash{e_{\textrm{opt}}}$, and $e$ are the cost of \autoref{eq:backward_algorithm} when using the initial embedding, fully optimized parameters, and parameters produced with $m$, respectively. 
We set the objective value obtained with $m=1,000$ as $\smash{e_{\textrm{opt}}}$.
The case where $e_{\textrm{opt}} = e_{\textrm{init}}$ indicates there is no way to refine the embedding; thus, we discard such cases. 
The result is shown in \autoref{fig:perf_eval_backward}.

From \autoref{fig:perf_eval_backward}a, the completion time linearly increases by $m$.
In \autoref{fig:perf_eval_backward}b, the accuracy is clearly improved till $m$ reaches $20$ or $40$; however, afterward, clear improvement cannot be seen.
From this, we can consider that the optimization can often produce a sufficient solution at around 40th iterations.
Also, the optimization is finished within about 1 second when $m \leq 20$ and 2 seconds when $m \leq 40$. 
Thus, we can set $m$ about 20 when we need more interactivity, otherwise, we can set it about $40$. 
However, when the number of groups is large (e.g., $c=6$), the accuracy could be low (e.g., 0.6), as shown in \autoref{fig:perf_eval_backward}b.
In such cases, we can consider using larger $m$ while sacrificing the interactivity.

In summary, our evaluation shows the high efficiency of \name{}: the completion time is about 1 second even when $n=10,000$ and $d=1,000$.
Thus, we can instantaneously update \name{} when adjusting parameters in the visual interface.
The background parameter selection also can provide a sufficient result in a considerably short time (e.g., 80\% of accuracy in 2 seconds). 
The results guide selection of solvers (e.g., \texttt{EVD} vs \texttt{Manopt}) and parameters (especially, the maximum number of iterations in COBYLA) for better performance and interactive usage.

\vspace{-4pt}
\section{Case Studies}
\label{sec:case_studies}
\vspace{-2pt}

We have demonstrated the usefulness of our framework by analyzing the Wine dataset~\cite{uci_mlr} in \autoref{sec:workflow_and_example}. 
With publicly available data, we perform two additional case studies.
All detailed information of the data (e.g., details of the survey questions in \autoref{sec:cs1}), analysis processes, used parameters, and results are available at our website~\cite{supp}.

\subsection{Case Study 1: Analysis of Political Groups}
\label{sec:cs1}

We\,analyze\,the\,PPIC\,Statewide\,Survey,\,October 2018~\cite{ppic_survey}.
This survey contains California residents' political opinions on, for example, political parties and expansion of the Mexico-U.S. barrier. 
By comparing groups within this dataset, such as supporters of different political parties, we can identify their opinion differences or reveal subgroups within each group.\,As\,a\,representative\,analysis,\,we look for a subgroup within the Democrat supporters (\texttt{Dem}) by comparing them with the Republican supporters (\texttt{Rep}), and review the characteristics of the subgroup.

In the Jupyter Notebook, we first preprocess the dataset to select data points (i.e., residents) and attributes (i.e., survey questions) of interest, discard missing values, and apply normalization to each attribute.
Note that while most of the attributes are either binary, ordinal, or numerical, we drop nominal data because it is not suitable to be analyzed with \name{}.
In the end, 548 data points and 27 attributes remain.

We then use \name{} with parameters that produce the same result when applying cPCA to \texttt{Dem} and \texttt{Rep} as target and background groups, respectively. 
This setting reveals opinions that are varied in \texttt{Dem} but uniform in \texttt{Rep}.   
With the visual interface, we initially display the \name{} result with $\ContParam = 0$, and interactively increase the value until we find interesting patterns, resulting \autoref{fig:cs1}a where $\ContParam = 20$. 
In \autoref{fig:cs1}a, \texttt{Dem} is separated into both left and right sides, while \texttt{Rep} is mostly placed at the right-hand side.
As shown in the component view, $x$-axis is dominantly constructed from \texttt{Q30}, which asks if they have a favorable impression of the current Democrat party.
By hovering over \texttt{Q30} in the view, we can see that \texttt{Dem} has both supporters who do and do not have a favorable impression of the Democrats. 
Note that we easily found this result because of exploratory data analysis using DR.
For example,\,the\,Mann-Whitney\,U\,rank\,test\,on \texttt{Dem} and \texttt{Rep} for each question reveals that most questions (20 of 27) have a statistically significant difference.
This information is not useful to find the subgroups of \texttt{Dem}.
Also, without interactively adjusting $\ContParam$ and looking at \autoref{fig:cs1}a, finding the \texttt{Dem}'s subgroup overlapping with \texttt{Rep} is difficult.

To investigate more of the residents who support the Democrats but do not have a favorable impression, we write a script in the Jupyter Notebook to separate \texttt{Dem} based on the answer for \texttt{Q30}: \texttt{Dem(+)} (the answer is `yes') and \texttt{Dem(-)} (`no'), and drop \texttt{Q30} from the attributes.
Then, we apply \name{} with parameters that highlight opinions that are more varied in \texttt{Dem(-)} than \texttt{Dem(+)} and \texttt{Rep}.
The result in \autoref{fig:cs1}b successfully reveals such opinions (i.e., \texttt{Dem(-)} has a higher variance than others).
Unlike \autoref{fig:cs1}a, many attributes significantly contribute to the axes, such as \texttt{Q1} (whether Jerry Brown's performance as the California governor is appropriate or not) and \texttt{Q25} (whether the gun restriction should be more strict or not).
As an example, we select \texttt{Q1} in \autoref{fig:cs1}b, where \texttt{Dem(+)} (blue) and \texttt{Rep} (red) have almost uniform opinions on Brown's performance (\texttt{Dem(+)}: positive, \texttt{Rep}: negative opinions) while \texttt{Dem(-)} consists of both opinions. 
Through this analysis, we can expect that those in \texttt{Dem(-)} have objections to some Democrats' policies, leading to an unfavorable impression of the current party.

Next, we identify political opinions that can clearly distinguish \texttt{Dem(+)} and \texttt{Rep} but are diverse within \texttt{Dem(-)}. 
To achieve this, we interactively move away \texttt{Rep} and \texttt{Dem(+)}'s centroids from each other in the interface. 
The backward parameter selection then automatically finds the proper parameters to refine the result, as shown in \autoref{fig:cs1}c.
At this time, \texttt{Q20} (annotated with a purple box) and \texttt{Q21a} (orange box) most contribute to $x$- and $y$-axes, respectively. 
\texttt{Q20} asks if they approve of Donald Trump's performance as the president while \texttt{Q21a} is about their opinion on Trump's nomination of Brett Kavanaugh to the U.S. Supreme Court. 
Thus, both questions are highly related to how the residents think about Donald Trump.
We select \texttt{Q20} from the component view. 
As shown in \autoref{fig:cs1}c, the majority of \texttt{Dem(+)} does not approve Trump's performance. 
However, a considerable amount of people in \texttt{Dem(-)} appreciate the work by Donald Trump.

\begin{figure*}[tb]
    \centering
    \includegraphics[width=\linewidth]{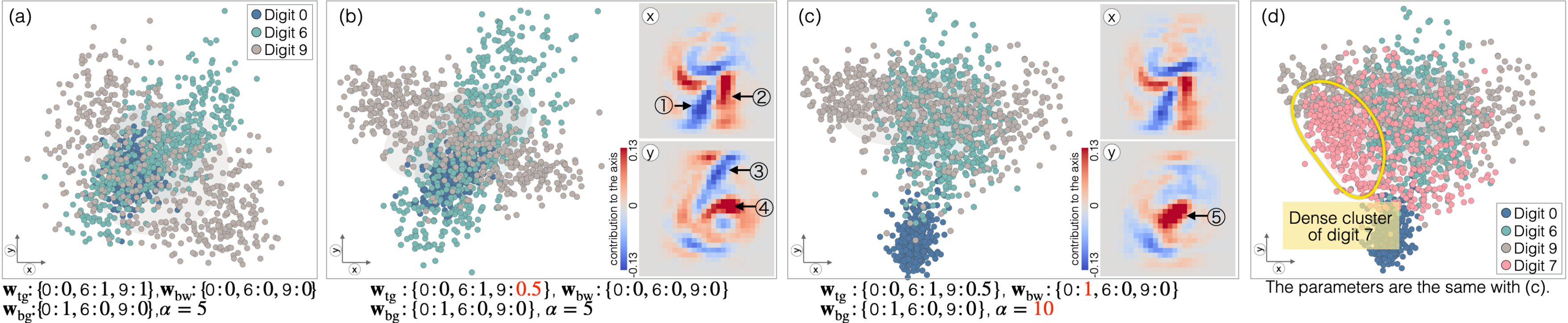}
    \caption{Case Study 2. Each of the results (a--d) is generated with the parameters listed at the bottom. In (b) and (c), we use 2D heatmaps to show each pixel's contribution to each axis. Dark blue and red pixels highly contribute to the negative and positive directions of the axes, respectively.}
    \label{fig:cs2}
\end{figure*}

\vspace{-1pt}
\subsection{Case Study 2: Characterizing Handwritten Digits}
\vspace{-1pt}

We analyze the MNIST handwritten digits dataset~\cite{mnist}.
This dataset contains 70,000 handwritten digits (i.e., data points) stored in $28 \times 28$ pixels (i.e, 784 attributes). 
Here, we review each digit's characteristics by comparing it to other digits. 
More specifically, we compare digits \texttt{0}, \texttt{6}, and \texttt{9}, all of which have a similar rounded structure. 

We first sample 500 images for each digit to moderate visual clutter in\,a\,resultant\,embedding\,space.\,To understand\,the\,various\,structures that people write for \texttt{6} and \texttt{9} but not for \texttt{0}, we use \name{} to maximize \texttt{6} and \texttt{9}'s variances while minimizing \texttt{0}'s variance in the embedding space. 
The result in \autoref{fig:cs2}a shows that digit \texttt{9} has a much higher variance than \texttt{6}. This implies the embedding mainly captures the information related to \texttt{9}. To produce similar variances for both \texttt{6} and \texttt{9}, we interactively reduce \texttt{9}'s weight in $\WeightTg$. In \autoref{fig:cs2}b, when we halve the corresponding weight, digits \texttt{6} and \texttt{9} achieve a similar variance. Also, we can see that \texttt{9} and \texttt{6} are widely distributed along $x$- and $y$-axes, respectively.

To understand structures highly related to the spread along each axis, we refer to each pixel's contribution to the construction of each axis.
However, since we have many pixels and want to see the contribution information in the context of digit shapes, the component view is not suitable for this analysis.
As we can seamlessly access all the information in the visual interface, we extract the axis information and depict it in 2D heatmaps (the right side of \autoref{fig:cs2}b) with an external visualization library.
From the heatmaps, we can see how people tend to write each digit differently.
For example, as annotated with \textcircled{\small 1} and \textcircled{\small 2}, digit \texttt{9} is written typically in two ways with a straight line, which cannot be seen in digit \texttt{0}.
For digit \texttt{6}, we observe that people tend to use either of the strokes annotated with \textcircled{\small 3} and \textcircled{\small 4} to make \texttt{6} different from \texttt{0}.
However, \texttt{6} is overlapped with \texttt{0} at the bottom left in the embedding space, where we expect \texttt{6} is written with the stroke annotated with \textcircled{\small 3}.

We move on to finding strokes that clearly differentiate \texttt{6} and \texttt{9} from \texttt{0} but are still written variously for \texttt{6} and \texttt{9}.
These strokes characterize the uniqueness of \texttt{6} and \texttt{9} relative to \texttt{0}.
To see if we can separate \texttt{0} from the others in the embedding, we set 1 to \texttt{0}'s weight in $\WeightBw$.
Then, by gradually increasing $\ContParam$, we find the desired embedding (\autoref{fig:cs2}c), where \texttt{0} has almost no overlap with the others. 
As seen in the heatmaps in\,\autoref{fig:cs2}c, while\,the\,new\,$x$-axis is similar to the previous one, the new $y$-axis is highly influenced by the pixels annotated with \textcircled{\small{5}}.
This implies that the corresponding pixels tend to be used only when writing digit \texttt{6} or \texttt{9}, but the amount of these pixels used for \texttt{6} or \texttt{9} differs by person.

Lastly, because we expect that the strokes annotated with \textcircled{\small 1} and \textcircled{\small 2} are also used for other digits, such as \texttt{1}, \texttt{4}, and \texttt{7}, we are interested in whether the patterns similar to \texttt{9} can be found (or not found) for these other digits.
By using the projection matrix used for \autoref{fig:cs2}c, we plot the sampled 500 data points of digit \texttt{7} onto the same embedding space. 
From the result shown in \autoref{fig:cs2}d, we can see that \texttt{7} has a similar distribution as \texttt{9} along $x$-axis in general; however, \texttt{7} is clustered more on the left-hand side of the embedding space, as annotated with yellow.
This annotated part tends to use the diagonal stroke \textcircled{\small{1}} in \autoref{fig:cs2}b. 
Thus, when writing \texttt{7}, people often use the diagonal stroke \textcircled{\small{1}} more and the vertical stroke \textcircled{\small{2}} less when compared to writing \texttt{9}.

This case study demonstrates how the embedding results can guide the analysis (e.g., the parameter adjustment after looking at the variance difference between \texttt{6} and \texttt{9} in \autoref{fig:cs2}a) and how the component/axis information can help analysts find patterns rooted in the combination of multiple attributes (e.g., the strokes shown in \autoref{fig:cs2}b).

\begin{figure}[tb]
    \centering
    \includegraphics[width=\linewidth]{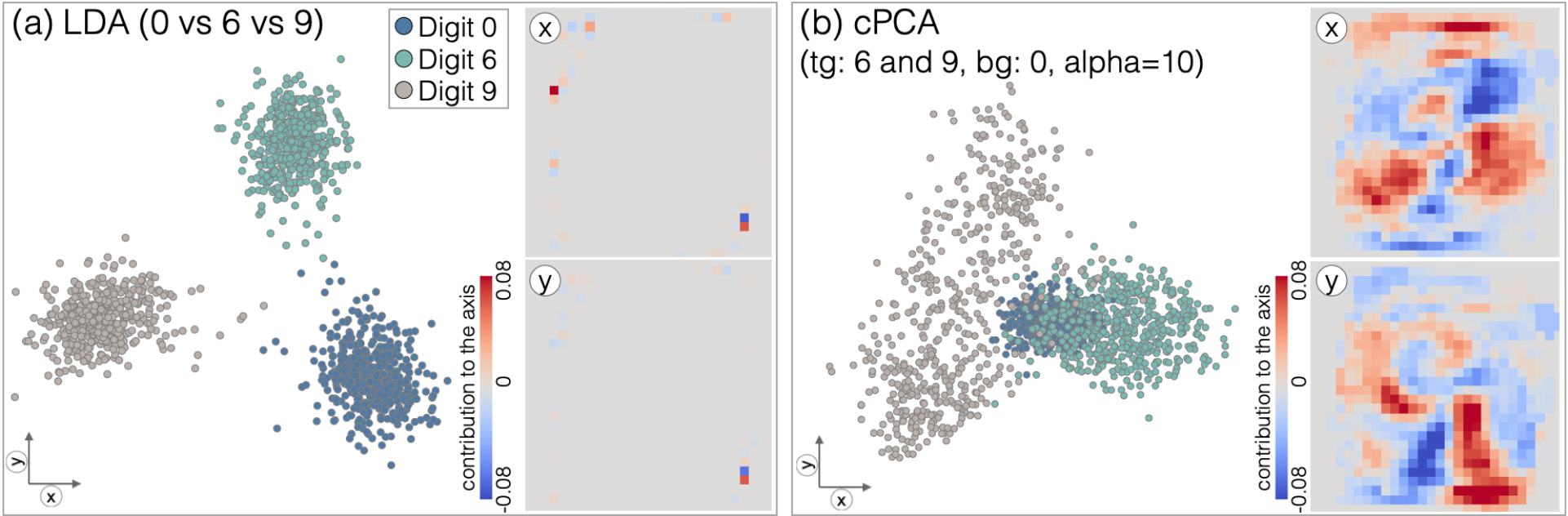}
    \caption{LDA and cPCA results of the dataset used in Case Study 2.}
    \label{fig:lda_cpca}
\end{figure}

\section{Discussion}

We demonstrate the usefulness of our framework for comparative analysis with our case studies, where we use the collective power of discriminant analysis, contrastive learning, and interactive refinement of DR results. 
These demonstrative analyses highlight the potential usage of\,our\,framework\,for\,a\,variety\,of\,applications.\,The patterns found in the case studies are difficult to identify with other DR methods, such as PCA,\,LDA,\,and\,cPCA.\,For example,\,as shown in \autoref{fig:lda_cpca},\,while applying LDA and cPCA to the dataset used in Case Study 2 can reveal clusters\,for\,digits\,\texttt{0},\,\texttt{6},\,and \texttt{9},\,we\,cannot\,find\,interesting patterns\,from the information of components unlike what ULCA provides.
We provide a comprehensive comparison among ULCA and other methods in the supplementary material~\cite{supp}.
Our algorithms' performance is also discussed in \autoref{sec:perf_eval}.
Here, we further discuss other aspects of our framework.

\vspace{2pt}
\noindent
\textbf{Visualization design.} 
The design of our visual interface aims to keep visualizations as simple as possible and only use visualizations that most analysts are likely to be familiar with.
However, at the same time, this design choice may cause problems, such as visual scalability. 
For example, in the parameter view, we assign an individual bar chart for each group's weight and a different color for each group; consequently, the related visualizations are not scalable to a large number of groups. 
Similarly, the component view is not scalable for a dataset with a large number of attributes.
However, due to the seamless integration with Python, we can easily use alternative visualizations, as demonstrated in the case study.
Ideally, the visual interface should allow the analysts or visualization developers to customize views based on their analysis needs. 
We plan to extend our implementation to be applicable to any visualizations developed with D3 into the visual interface.

\vspace{2pt}
\noindent
\textbf{Parameter suggestion.} 
The backward parameter selection finds parameters that resemble the user-demonstrated change as close as possible. 
However, the analyst may try infeasible refinements. 
For example, they might try to increase some group's variance even when its variance is already maximized.
To avoid this, similar to the feasibility map by Cavallo and Demiralp~\cite{cavallo2018visual}, we could precompute embeddings derived with various potential interactions and inform which interactions are feasible. 
However, this approach has a drawback in its computational cost as we need to check the feasibility from the wide searching space. 
Another interesting approach is suggesting which parameter values can result in significantly different embedding results when compared to other cases. 
For this approach, we can employ several existing methods~\cite{abid2018exploring,lehmann2016optimal}.
For example, for cPCA, Abid and Zhang \etal~\cite{abid2018exploring} suggested a small set of contrast parameter $\ContParam$ values (e.g., 4 values) by applying spectral clustering to the embedding results generated with many $\ContParam$ values (e.g., 40 values). 
This approach, though, also would have a large computational cost as \name{} has many adjustable parameters.
We will investigate how we can achieve the recommendation of interactions or parameters to find interesting embedding patterns in the future. 

\vspace{2pt}
\noindent
\textbf{Characteristics of ULCA.}
\name{} is a tool for exploratory data analysis. 
Similar to the discussion on cPCA by Abid et al.~\cite{abid2018exploring}, while \name{} uses group labels for embedding, ULCA is different from other supervised learning methods that focus on a task of classification/regression and multi-group statistical hypothesis tests whose primary goal is identifying an attribute that has a significantly different statistic value (e.g., mean) in each group. 
\name{} is a linear DR method, which can extract a latent linear structure from high-dimensional data.
\name{} benefits from the linearity to provide interpretable embedding axes; however, the linearity also might limit \name{}'s capability to find patterns when analyzing complex data.
We plan to investigate how to extend \name{} to a nonlinear DR method while retaining\,interpretability.\,Another\,limitation\,of\,\name{}\,is\,interpreting the result could be still difficult when analyzing a dataset with a very large number of attributes (e.g., 1,000 attributes) even with the axis information.
When applying ULCA to such a dataset, many of the attributes (e.g., 100 attributes) could significantly contribute to embedding axes. 
To avoid using many attributes when constructing the embedding axes, various sparse DR methods, such as sparse PCA~\cite{zou2006sparse}, sparse LDA~\cite{clemmensen2011sparse}, and space cPCA~\cite{boileau2020exploring}, have been developed.
These methods produce a sparse projection matrix by penalizing a case where an embedding uses many attributes.
We also plan to extend \name{} to a sparse DR method.

\vspace{-2pt}
\section{Conclusion}
\vspace{-1pt}

We have developed an interactive DR framework by introducing a new DR method, \name{}, which uses the collective capability of discriminant analysis and contrastive learning.
The framework further supports interactive analysis with a visual interface that can be seamlessly used with existing analysis libraries and also with an optimization algorithm that can interactively refine \name{} results.  
This new DR framework makes a tangible contribution to comparative analysis. 

\acknowledgments{The authors wish to thank Dr. Tzu-Ping Liu at the University of Taipei and Samuel Fuller at UC Davis for their guidance for the dataset used in Case Study 1. This work is supported in part by the U.S. National Science Foundation through grant IIS-1741536, the U.S. National Institute of Standards and Technology through grant 70NANB20H197, and the Natural Sciences and Engineering Research Council of Canada.}

\bibliographystyle{abbrv-doi}
\bibliography{00_references}

\begin{thebibliography}{10}

\bibitem{supp}
The supplementary materials: The demonstration video of the framework, source
  code, evaluation details, and supplemental evaluation.
\newblock \url{https://takanori-fujiwara.github.io/s/ulca/}.

\bibitem{abid2018exploring}
A.~Abid, M.~J. Zhang, V.~K. Bagaria, and J.~Zou.
\newblock Exploring patterns enriched in a dataset with contrastive principal
  component analysis.
\newblock {\em Nat. Commun.}, 9(1):2134, 2018.

\bibitem{absil2007trust}
P.-A. Absil, C.~G. Baker, and K.~A. Gallivan.
\newblock Trust-region methods on {Riemannian} manifolds.
\newblock {\em Found. of Comput. Math.}, 7(3):303--330, 2007.

\bibitem{absil2009optimization}
P.-A. Absil, R.~Mahony, and R.~Sepulchre.
\newblock {\em Optimization algorithms on matrix manifolds}.
\newblock Princeton University Press, 2009.

\bibitem{ppic_survey}
M.~Baldassare, D.~Bonner, A.~Dykman, and L.~Lopes.
\newblock {PPIC} statewide survey: Californians \& their government, {October}
  2018.
\newblock Report:
  \url{https://www.ppic.org/wp-content/uploads/ppic-statewide-survey-october-2018.pdf},
  Dataset:
  \url{https://www.ppic.org/data-set/ppic-statewide-survey-data-2018/}, 2018.
\newblock Accessed: 2021-3-21.

\bibitem{blei2003latent}
D.~M. Blei, A.~Y. Ng, and M.~I. Jordan.
\newblock Latent {Dirichlet} allocation.
\newblock {\em J. Mach. Learn. Res.}, 3:993--1022, 2003.

\bibitem{boileau2020exploring}
P.~Boileau, N.~S. Hejazi, and S.~Dudoit.
\newblock Exploring high-dimensional biological data with sparse contrastive
  principal component analysis.
\newblock {\em Bioinformatics}, 36(11):3422--3430, 2020.

\bibitem{bostock2011d3}
M.~Bostock, V.~Ogievetsky, and J.~Heer.
\newblock {D$^3$} data-driven documents.
\newblock {\em IEEE Trans. Vis. Comput. Graph.}, 17(12):2301--2309, 2011.

\bibitem{boumal2019global}
N.~Boumal, P.-A. Absil, and C.~Cartis.
\newblock Global rates of convergence for nonconvex optimization on manifolds.
\newblock {\em IMA J. Numer. Anal.}, 39(1):1--33, 2019.

\bibitem{boumal2014manopt}
N.~Boumal, B.~Mishra, P.-A. Absil, and R.~Sepulchre.
\newblock {Manopt}, a {Matlab} toolbox for optimization on manifolds.
\newblock {\em J. Mach. Learn. Res.}, 15(1):1455--1459, 2014.

\bibitem{brehmer2014visualizing}
M.~Brehmer, M.~Sedlmair, S.~Ingram, and T.~Munzner.
\newblock Visualizing dimensionally-reduced data: Interviews with analysts and
  a characterization of task sequences.
\newblock In {\em Proc. BELIV}, pp. 1--8, 2014.

\bibitem{brown2012disfunction}
E.~T. {Brown}, J.~{Liu}, C.~E. {Brodley}, and R.~{Chang}.
\newblock {Dis-Function}: Learning distance functions interactively.
\newblock In {\em Proc. VAST}, pp. 83--92, 2012.

\bibitem{cantareira2020generic}
G.~D. Cantareira and F.~V. Paulovich.
\newblock A generic model for projection alignment applied to neural network
  visualization.
\newblock In {\em Proc. EuroVA}, 2020.

\bibitem{cavallo2018visual}
M.~Cavallo and {\c{C}}.~Demiralp.
\newblock A visual interaction framework for dimensionality reduction based
  data exploration.
\newblock In {\em Proc. CHI}, pp. 1--13, 2018.

\bibitem{clemmensen2011sparse}
L.~Clemmensen, T.~Hastie, D.~Witten, and B.~Ersb{\o}ll.
\newblock Sparse discriminant analysis.
\newblock {\em Technometrics}, 53(4):406--413, 2011.

\bibitem{coimbra2016explaining}
D.~B. Coimbra, R.~M. Martins, T.~T. Neves, A.~C. Telea, and F.~V. Paulovich.
\newblock Explaining three-dimensional dimensionality reduction plots.
\newblock {\em Inform. Visual.}, 15(2):154--172, 2016.

\bibitem{cunningham2015linear}
J.~P. Cunningham and Z.~Ghahramani.
\newblock Linear dimensionality reduction: Survey, insights, and
  generalizations.
\newblock {\em J. Mach. Learn. Res.}, 16(1):2859--2900, 2015.

\bibitem{dinkelbach1967nonlinear}
W.~Dinkelbach.
\newblock On nonlinear fractional programming.
\newblock {\em Management Science}, 13(7):492--498, 1967.

\bibitem{dowling2019sirius}
M.~{Dowling}, J.~{Wenskovitch}, J.~T. {Fry}, S.~{Leman}, L.~{House}, and
  C.~{North}.
\newblock {SIRIUS}: Dual, symmetric, interactive dimension reductions.
\newblock {\em IEEE Trans. Vis. Comput. Graph.}, 25(1):172--182, 2019.

\bibitem{du2010choosing}
J.-B. du~Prel, B.~R{\"o}hrig, G.~Hommel, and M.~Blettner.
\newblock Choosing statistical tests: Part 12 of a series on evaluation of
  scientific publications.
\newblock {\em Deutsches {\"A}rzteblatt International}, 107(19):343, 2010.

\bibitem{uci_mlr}
D.~Dua and C.~Graff.
\newblock {UCI} machine learning repository.
\newblock \url{http://archive.ics.uci.edu/ml}, 2019.

\bibitem{endert2011observation}
A.~{Endert}, C.~{Han}, D.~{Maiti}, L.~{House}, S.~{Leman}, and C.~{North}.
\newblock Observation-level interaction with statistical models for visual
  analytics.
\newblock In {\em Proc. VAST}, pp. 121--130, 2011.

\bibitem{espadoto2021toward}
M.~{Espadoto}, R.~M. {Martins}, A.~{Kerren}, N.~S.~T. {Hirata}, and A.~C.
  {Telea}.
\newblock Toward a quantitative survey of dimension reduction techniques.
\newblock {\em IEEE Trans. Vis. Comput. Graph.}, 27(3):2153--2173, 2021.

\bibitem{fayyad1996data}
U.~Fayyad, G.~Piatetsky-Shapiro, and P.~Smyth.
\newblock From data mining to knowledge discovery in databases.
\newblock {\em AI Mag.}, 17(3):37--37, 1996.

\bibitem{fujiwara2019incremental}
T.~Fujiwara, J.-K. Chou, S.~Shilpika, P.~Xu, L.~Ren, and K.-L. Ma.
\newblock An incremental dimensionality reduction method for visualizing
  streaming multidimensional data.
\newblock {\em IEEE Trans. Vis. Comput. Graph.}, 26(1):418--428, 2020.

\bibitem{fujiwara2019supporting}
T.~Fujiwara, O.-H. Kwon, and K.-L. Ma.
\newblock Supporting analysis of dimensionality reduction results with
  contrastive learning.
\newblock {\em IEEE Trans. Vis. Comput. Graph.}, 26(1):45--55, 2020.

\bibitem{fujiwara2020contrastive}
T.~Fujiwara and T.-P. Liu.
\newblock Contrastive multiple correspondence analysis ({cMCA}): Using
  contrastive learning to identify latent subgroups in political parties.
\newblock {\em arXiv:2007.04540}, 2020.

\bibitem{fujiwara2020interpretable}
T.~Fujiwara, J.~Zhao, F.~Chen, Y.~Yu, and K.-L. Ma.
\newblock Interpretable contrastive learning for networks.
\newblock {\em arXiv:2005.12419}, 2020.

\bibitem{garcia2015data}
S.~Garc{\'\i}a, J.~Luengo, and F.~Herrera.
\newblock {\em Data Preprocessing in Data Mining}, vol.~72.
\newblock Springer, 2015.

\bibitem{ge2016rich}
R.~Ge and J.~Zou.
\newblock Rich component analysis.
\newblock In {\em Proc. ICML}, pp. 1502--1510, 2016.

\bibitem{gleicher2013explainers}
M.~{Gleicher}.
\newblock Explainers: Expert explorations with crafted projections.
\newblock {\em IEEE Trans. Vis. Comput. Graph.}, 19(12):2042--2051, 2013.

\bibitem{gleicher2018considerations}
M.~Gleicher.
\newblock Considerations for visualizing comparison.
\newblock {\em IEEE Trans. Vis. Comput. Graph.}, 24(1):413--423, 2018.

\bibitem{gleicher2011visual}
M.~Gleicher, D.~Albers, R.~Walker, I.~Jusufi, C.~D. Hansen, and J.~C. Roberts.
\newblock Visual comparison for information visualization.
\newblock {\em Inform. Visual.}, 10(4):289--309, 2011.

\bibitem{gower2004procrustes}
J.~C. Gower and G.~B. Dijksterhuis.
\newblock {\em Procrustes Problems}, vol.~30.
\newblock Oxford University Press on Demand, 2004.

\bibitem{guo2020comparative}
R.~Guo, T.~Fujiwara, Y.~Li, K.~M. Lima, S.~Sen, N.~K. Tran, and K.-L. Ma.
\newblock Comparative visual analytics for assessing medical records with
  sequence embedding.
\newblock {\em Visual Informatics}, 4(2):72--85, 2020.

\bibitem{guo2007regularized}
Y.~Guo, T.~Hastie, and R.~Tibshirani.
\newblock Regularized linear discriminant analysis and its application in
  microarrays.
\newblock {\em Biostatistics}, 8(1):86--100, 2007.

\bibitem{guo2003generalized}
Y.-F. Guo, S.-J. Li, J.-Y. Yang, T.-T. Shu, and L.-D. Wu.
\newblock A generalized {Foley--Sammon} transform based on generalized {Fisher}
  discriminant criterion and its application to face recognition.
\newblock {\em Pattern Recogn. Lett.}, 24(1-3):147--158, 2003.

\bibitem{hare2015using}
C.~Hare, D.~A. Armstrong, R.~Bakker, R.~Carroll, and K.~T. Poole.
\newblock Using bayesian {Aldrich-McKelvey} scaling to study citizens'
  ideological preferences and perceptions.
\newblock {\em Am. J. Polit. Sci.}, 59(3):759--774, 2015.

\bibitem{hartigan1979algorithm}
J.~A. Hartigan and M.~A. Wong.
\newblock A k-means clustering algorithm.
\newblock {\em J. R. Stat. Soc. C-Appl.)}, 28(1):100--108, 1979.

\bibitem{hastie1996discriminant}
T.~Hastie and R.~Tibshirani.
\newblock Discriminant analysis by {Gaussian} mixtures.
\newblock {\em J. R. Stat. Soc. B}, pp. 155--176, 1996.

\bibitem{hill2015real}
J.~Hill, W.~R. Ford, and I.~G. Farreras.
\newblock Real conversations with artificial intelligence: A comparison between
  human--human online conversations and human--chatbot conversations.
\newblock {\em Comput. Hum. Behav.}, 49:245--250, 2015.

\bibitem{hofseth2020early}
L.~J. Hofseth, J.~R. Hebert, A.~Chanda, H.~Chen, B.~L. Love, M.~M. Pena, E.~A.
  Murphy, M.~Sajish, A.~Sheth, P.~J. Buckhaults, et~al.
\newblock Early-onset colorectal cancer: Initial clues and current views.
\newblock {\em Nat. Rev. Gastro. Hepat.}, 17(6):352--364, 2020.

\bibitem{hotelling1933analysis}
H.~Hotelling.
\newblock Analysis of a complex of statistical variables into principal
  components.
\newblock {\em J. Educ. Psychol.}, 24(6):417, 1933.

\bibitem{hotelling1992relations}
H.~Hotelling.
\newblock Relations between two sets of variates.
\newblock In S.~Kotz and N.~L. Johnson, eds., {\em Breakthroughs in Statistics:
  Methodology and Distribution}, pp. 162--190. Springer New York, 1992.

\bibitem{huber1985projection}
P.~J. Huber.
\newblock Projection pursuit.
\newblock {\em Ann. Stat.}, pp. 435--475, 1985.

\bibitem{hunter2007matplotlib}
J.~D. Hunter.
\newblock Matplotlib: A {2D} graphics environment.
\newblock {\em Comput. Sci. Eng.}, 9(3):90--95, 2007.

\bibitem{izenman2008modern}
A.~J. Izenman.
\newblock Linear discriminant analysis.
\newblock In {\em Modern Multivariate Statistical Techniques}, pp. 237--280.
  Springer, 2013.

\bibitem{jeong2009ipca}
D.~H. Jeong, C.~Ziemkiewicz, B.~Fisher, W.~Ribarsky, and R.~Chang.
\newblock {iPCA}: An interactive system for {PCA}-based visual analytics.
\newblock {\em Comput. Graph. Forum}, 28(3):767--774, 2009.

\bibitem{jia2009trace}
Y.~{Jia}, F.~{Nie}, and C.~{Zhang}.
\newblock Trace ratio problem revisited.
\newblock {\em IEEE Trans. Neural Netw.}, 20(4):729--735, 2009.

\bibitem{jiang2019recent}
L.~Jiang, S.~Liu, and C.~Chen.
\newblock Recent research advances on interactive machine learning.
\newblock {\em J. Visualization}, 22(2):401--417, 2019.

\bibitem{jin2020carepre}
Z.~Jin, S.~Cui, S.~Guo, D.~Gotz, J.~Sun, and N.~Cao.
\newblock {CarePre}: An intelligent clinical decision assistance system.
\newblock {\em ACM Trans. Comput. Healthcare}, 1(1):1--20, 2020.

\bibitem{johansson2009interactive}
J.~Johansson and S.~Johansson.
\newblock Interactive dimensionality reduction through user-defined
  combinations of quality metrics.
\newblock {\em IEEE Trans. Vis. and Comput. Graphics}, 15(06), 2009.

\bibitem{joia2011local}
P.~{Joia}, D.~{Coimbra}, J.~A. {Cuminato}, F.~V. {Paulovich}, and L.~G.
  {Nonato}.
\newblock Local affine multidimensional projection.
\newblock {\em IEEE Trans. Vis. Comput. Graph.}, 17(12):2563--2571, 2011.

\bibitem{jolliffe1986principal}
I.~T. Jolliffe.
\newblock {\em Principal Component Analysis and Factor Analysis}, pp. 115--128.
\newblock Springer, 1986.

\bibitem{jones1987projection}
M.~C. Jones and R.~Sibson.
\newblock What is projection pursuit?
\newblock {\em J. R. Stat. Soc. A-G.}, 150(1):1--18, 1987.

\bibitem{kaiser1958varimax}
H.~F. Kaiser.
\newblock The varimax criterion for analytic rotation in factor analysis.
\newblock {\em Psychometrika}, 23(3):187--200, 1958.

\bibitem{kandogan2000star}
E.~Kandogan.
\newblock Star coordinates: A multi-dimensional visualization technique with
  uniform treatment of dimensions.
\newblock In {\em Proc. InfoVis}, pp. 9--12, 2000.

\bibitem{keim2008visual}
D.~Keim, G.~Andrienko, J.-D. Fekete, C.~G{\"o}rg, J.~Kohlhammer, and
  G.~Melan{\c{c}}on.
\newblock Visual analytics: Definition, process, and challenges.
\newblock In A.~Kerren, J.~T. Stasko, J.-D. Fekete, and C.~North, eds., {\em
  Information Visualization: Human-Centered Issues and Perspectives}, pp.
  154--175. Springer Berlin Heidelberg, 2008.

\bibitem{kim2016interaxis}
H.~{Kim}, J.~{Choo}, H.~{Park}, and A.~{Endert}.
\newblock {InterAxis}: Steering scatterplot axes via observation-level
  interaction.
\newblock {\em IEEE Trans. Vis. Comput. Graph.}, 22(1):131--140, 2016.

\bibitem{jupyter}
T.~Kluyver, B.~Ragan-Kelley, F.~P{\'e}rez, B.~Granger, M.~Bussonnier,
  J.~Frederic, K.~Kelley, J.~Hamrick, J.~Grout, S.~Corlay, P.~Ivanov, D.~Avila,
  S.~Abdalla, C.~Willing, and J.~development team.
\newblock Jupyter notebooks - a publishing format for reproducible
  computational workflows.
\newblock In F.~Loizides and B.~Scmidt, eds., {\em Positioning and Power in
  Academic Publishing: Players, Agents and Agendas}, pp. 87--90. IOS Press,
  2016.

\bibitem{kvam2012comparison}
V.~M. Kvam, P.~Liu, and Y.~Si.
\newblock A comparison of statistical methods for detecting differentially
  expressed genes from {RNA}-seq data.
\newblock {\em Am. J. Bot.}, 99(2):248--256, 2012.

\bibitem{kwon2017axisketcher}
B.~C. {Kwon}, H.~{Kim}, E.~{Wall}, J.~{Choo}, H.~{Park}, and A.~{Endert}.
\newblock {AxiSketcher}: Interactive nonlinear axis mapping of visualizations
  through user drawings.
\newblock {\em IEEE Trans. Vis. Comput. Graph.}, 23(1):221--230, 2017.

\bibitem{lage2018human}
I.~Lage, A.~Ross, S.~J. Gershman, B.~Kim, and F.~Doshi-Velez.
\newblock Human-in-the-loop interpretability prior.
\newblock In S.~Bengio, H.~Wallach, H.~Larochelle, K.~Grauman, N.~Cesa-Bianchi,
  and R.~Garnett, eds., {\em Proc. NIPS}, vol.~31, 2018.

\bibitem{mnist}
Y.~LeCun, C.~Cortes, and C.~J.C.~Burges.
\newblock The {MNIST} database of handwritten digits.
\newblock \url{http://yann.lecun.com/exdb/mnist/}, 1999.
\newblock Accessed: 2021-3-21.

\bibitem{legendre2013beta}
P.~Legendre and M.~De~C{\'a}ceres.
\newblock Beta diversity as the variance of community data: Dissimilarity
  coefficients and partitioning.
\newblock {\em Ecol. Lett.}, 16(8):951--963, 2013.

\bibitem{lehmann2016optimal}
D.~J. {Lehmann} and H.~{Theisel}.
\newblock Optimal sets of projections of high-dimensional data.
\newblock {\em IEEE Trans. Vis. Comput. Graph.}, 22(1):609--618, 2016.

\bibitem{liu2017visualizing}
S.~Liu, D.~Maljovec, B.~Wang, P.-T. Bremer, and V.~Pascucci.
\newblock Visualizing high-dimensional data: Advances in the past decade.
\newblock {\em IEEE Trans. Vis. Comput. Graph.}, 23(3):1249--1268, 2017.

\bibitem{mamani2013user}
G.~M. Mamani, F.~M. Fatore, L.~G. Nonato, and F.~V. Paulovich.
\newblock User-driven feature space transformation.
\newblock {\em Comput. Graph. Forum}, 32(3pt3):291--299, 2013.

\bibitem{mcinnes2018umap}
L.~McInnes, J.~Healy, and J.~Melville.
\newblock {UMAP}: Uniform manifold approximation and projection for dimension
  reduction.
\newblock {\em arXiv:1802.03426}, 2018.

\bibitem{mika1999fisher}
S.~Mika, G.~Ratsch, J.~Weston, B.~Scholkopf, and K.-R. Mullers.
\newblock {Fisher} discriminant analysis with kernels.
\newblock In {\em Proc. IEEE Signal Processing Society Workshop}, pp. 41--48,
  1999.

\bibitem{moore2015social}
J.~N. Moore, M.~A. Raymond, and C.~D. Hopkins.
\newblock Social selling: A comparison of social media usage across process
  stage, markets, and sales job functions.
\newblock {\em J. Mark. Theory Pract.}, 23(1):1--20, 2015.

\bibitem{nonato2018multidimensional}
L.~G. Nonato and M.~Aupetit.
\newblock Multidimensional projection for visual analytics: Linking techniques
  with distortions, tasks, and layout enrichment.
\newblock {\em IEEE Trans. Vis. Comput. Graph.}, 25(8):2650--2673, 2018.

\bibitem{pedregosa2011scikit}
F.~Pedregosa, G.~Varoquaux, A.~Gramfort, V.~Michel, B.~Thirion, O.~Grisel,
  M.~Blondel, P.~Prettenhofer, R.~Weiss, V.~Dubourg, J.~Vanderplas, A.~Passos,
  D.~Cournapeau, M.~Brucher, M.~Perrot, and E.~Duchesnay.
\newblock Scikit-learn: Machine learning in {P}ython.
\newblock {\em J. Mach. Learn. Res.}, 12:2825--2830, 2011.

\bibitem{perez2015interactive}
D.~P{\'e}rez, L.~Zhang, M.~Schaefer, T.~Schreck, D.~Keim, and I.~D{\'\i}az.
\newblock Interactive feature space extension for multidimensional data
  projection.
\newblock {\em Neurocomputing}, 150:611--626, 2015.

\bibitem{pires2010projection}
A.~M. Pires and J.~A. Branco.
\newblock Projection-pursuit approach to robust linear discriminant analysis.
\newblock {\em J. Multivariate Anal.}, 101(10):2464--2485, 2010.

\bibitem{posse1992projection}
C.~G. Posse.
\newblock Projection pursuit discriminant analysis for two groups.
\newblock {\em Communications in Statistics-Theory and Methods}, 21(1):1--19,
  1992.

\bibitem{powell1998direct}
M.~J. Powell.
\newblock Direct search algorithms for optimization calculations.
\newblock {\em Acta Numer.}, pp. 287--336, 1998.

\bibitem{purchase2006important}
H.~C. Purchase, E.~Hoggan, and C.~G{\"o}rg.
\newblock How important is the ``mental map''?--an empirical investigation of a
  dynamic graph layout algorithm.
\newblock In {\em Proc. GD}, pp. 184--195. Springer, 2006.

\bibitem{ragan2016characterizing}
E.~D. Ragan, A.~Endert, J.~Sanyal, and J.~Chen.
\newblock Characterizing provenance in visualization and data analysis: An
  organizational framework of provenance types and purposes.
\newblock {\em IEEE Trans. Vis. Comput. Graph.}, 22(1):31--40, 2016.

\bibitem{rauber2016dynamic}
P.~E. Rauber, A.~X. Falc\~{a}o, and A.~C. Telea.
\newblock Visualizing time-dependent data using dynamic {t-SNE}.
\newblock In {\em Proc. EuroVis}, p. 73–77, 2016.

\bibitem{sacha2017visual}
D.~{Sacha}, L.~{Zhang}, M.~{Sedlmair}, J.~A. {Lee}, J.~{Peltonen},
  D.~{Weiskopf}, S.~C. {North}, and D.~A. {Keim}.
\newblock Visual interaction with dimensionality reduction: A structured
  literature analysis.
\newblock {\em IEEE Trans. Vis. Comput. Graph.}, 23(1):241--250, 2017.

\bibitem{saket2017visualization}
B.~{Saket}, H.~{Kim}, E.~T. {Brown}, and A.~{Endert}.
\newblock Visualization by demonstration: An interaction paradigm for visual
  data exploration.
\newblock {\em IEEE Trans. Vis. Comput. Graph.}, 23(1):331--340, 2017.

\bibitem{self2018observation}
J.~Z. Self, M.~Dowling, J.~Wenskovitch, I.~Crandell, M.~Wang, L.~House,
  S.~Leman, and C.~North.
\newblock Observation-level and parametric interaction for high-dimensional
  data analysis.
\newblock {\em ACM Trans. Interact. Intell. Syst.}, 8(2):1--36, 2018.

\bibitem{self2016bridging}
J.~Z. Self, R.~K. Vinayagam, J.~Fry, and C.~North.
\newblock Bridging the gap between user intention and model parameters for
  human-in-the-loop data analytics.
\newblock In {\em Proc. HILDA}, pp. 1--6, 2016.

\bibitem{sobhani2019colorectal}
I.~Sobhani, E.~Bergsten, S.~Couffin, A.~Amiot, B.~Nebbad, C.~Barau,
  N.~de’Angelis, S.~Rabot, F.~Canoui-Poitrine, D.~Mestivier, et~al.
\newblock Colorectal cancer-associated microbiota contributes to oncogenic
  epigenetic signatures.
\newblock {\em Proc. Natl. Acad. Sci.}, 116(48):24285--24295, 2019.

\bibitem{torgerson1952}
W.~S. Torgerson.
\newblock Multidimensional scaling: I. theory and method.
\newblock {\em Psychometrika}, 17(4):401--419, 1952.

\bibitem{townsend2016pymanopt}
J.~Townsend, N.~Koep, and S.~Weichwald.
\newblock Pymanopt: A python toolbox for optimization on manifolds using
  automatic differentiation.
\newblock {\em J. Mach. Learn. Res.}, 17(137):1--5, 2016.

\bibitem{maaten2008visualizing}
L.~van~der Maaten and G.~Hinton.
\newblock Visualizing data using {t-SNE}.
\newblock {\em {J. Mach. Learn. Res.}}, 9(Nov):2579--2605, 2008.

\bibitem{van2009dimensionality}
L.~van~der Maaten, E.~Postma, and J.~van~den Herik.
\newblock Dimensionality reduction: A comparative review.
\newblock {\em J. Mach. Learn. Res.}, 10:66--71, 2009.

\bibitem{virtanen2020scipy}
P.~Virtanen, R.~Gommers, T.~E. Oliphant, M.~Haberland, T.~Reddy, D.~Cournapeau,
  E.~Burovski, P.~Peterson, W.~Weckesser, J.~Bright, et~al.
\newblock {{SciPy} 1.0: Fundamental Algorithms for Scientific Computing in
  Python}.
\newblock {\em Nat. Methods}, 17:261--272, 2020.

\bibitem{wang2017linear}
Y.~Wang, J.~Li, F.~Nie, H.~Theisel, M.~Gong, and D.~J. Lehmann.
\newblock Linear discriminative star coordinates for exploring class and
  cluster separation of high dimensional data.
\newblock {\em Comput. Graph. Forum}, 36(3):401--410, 2017.

\bibitem{wenskovitch2018towards}
J.~Wenskovitch, I.~Crandell, N.~Ramakrishnan, L.~House, and C.~North.
\newblock Towards a systematic combination of dimension reduction and
  clustering in visual analytics.
\newblock {\em IEEE Trans. Vis. Comput. Graph.}, 24(1):131--141, 2018.

\bibitem{wenskovitch2020respect}
J.~Wenskovitch, M.~Dowling, and C.~North.
\newblock With respect to what? simultaneous interaction with dimension
  reduction and clustering projections.
\newblock In {\em Proc. IUI}, pp. 177--188, 2020.

\bibitem{wenskovitch2019pollux}
J.~{Wenskovitch} and C.~{North}.
\newblock Pollux: Interactive cluster-first projections of high-dimensional
  data.
\newblock In {\em Proc. VDS}, pp. 38--47, 2019.

\bibitem{yang2019survey}
X.~Yang, L.~Weifeng, W.~Liu, and D.~Tao.
\newblock A survey on canonical correlation analysis.
\newblock {\em IEEE Trans. Knowl.Data Eng.}, 33(6):2349--2368, 2021.

\bibitem{yasir2015comparison}
M.~Yasir, E.~Angelakis, F.~Bibi, E.~Azhar, D.~Bachar, J.-C. Lagier, B.~Gaborit,
  A.~Hassan, A.~Jiman-Fatani, K.~Alshali, et~al.
\newblock Comparison of the gut microbiota of people in {France} and {Saudi
  Arabia}.
\newblock {\em Nutr. Diabetes}, 5(4):e153--e153, 2015.

\bibitem{zhou2010stable}
Z.~Zhou, X.~Li, J.~Wright, E.~Candes, and Y.~Ma.
\newblock Stable principal component pursuit.
\newblock In {\em Proc. ISIT}, pp. 1518--1522. IEEE, 2010.

\bibitem{zou2006sparse}
H.~Zou, T.~Hastie, and R.~Tibshirani.
\newblock Sparse principal component analysis.
\newblock {\em J. Comput. and Grap. Stat.}, 15(2):265--286, 2006.

\bibitem{zou2013contrastive}
J.~Y. Zou, D.~J. Hsu, D.~C. Parkes, and R.~P. Adams.
\newblock Contrastive learning using spectral methods.
\newblock In {\em Proc. NIPS}, pp. 2238--2246, 2013.

\end{thebibliography}

\end{document}


\maketitle

We present this supplementary document to compare results generated by different linear dimensionality reduction (DR) methods: PCA, LDA, cPCA, and ULCA. 
As described in Sect. 4 and demonstrated in the case studies, ULCA's strength is in its analysis capability utilizing both discriminant analysis (e.g., LDA) and contrastive learning (e.g., cPCA) and flexibility allowing adjustment of algorithm parameters based on analysis need.
Here, we apply PCA, LDA, and cPCA to the dataset used in Sect. 3 and 7, and compare the ULCA results shown in Fig. 1, 2, 3, 9, and 10. 
While we perform PCA for all the analysis cases, we apply LDA and cPCA only when they can (partially) fulfill the analysis purpose in each case. 
More specifically, LDA is designed for maximizing the separation among groups in an embedding space; thus, we only use LDA when we try to see the separation of groups during the analysis.
Similarly, cPCA is to find an embedding space where a target group's variance is higher than a background group's variance; thus, we only use cPCA when we want to increase or decrease some group's variance during the analysis.

\section{Results for the Wine Dataset}

\begin{figure*}[h]
    \captionsetup{farskip=0pt}
	\centering
    \includegraphics[width=1.0\linewidth]{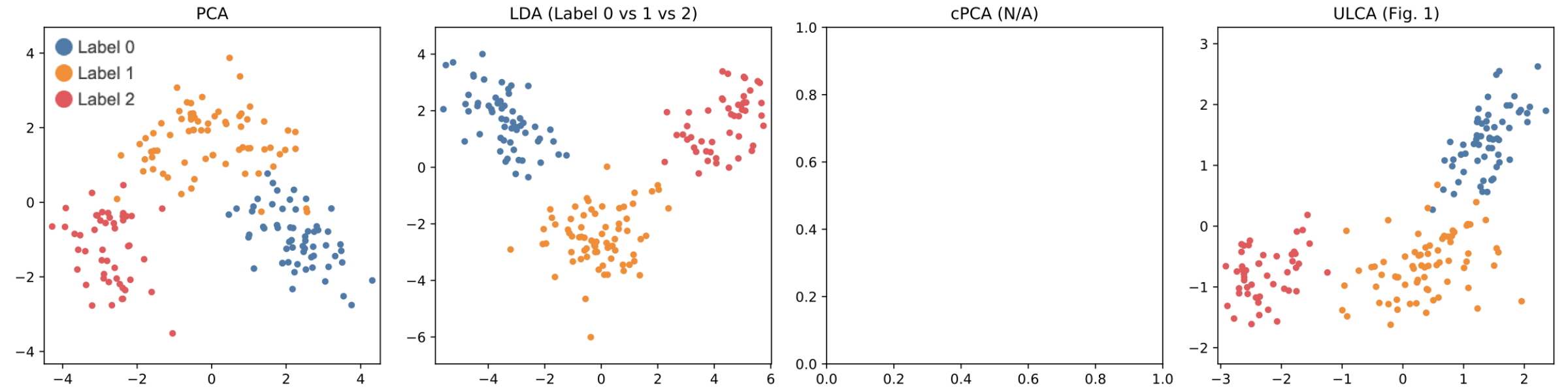}
  	\caption{The embedding results corresponding to Fig. 1.}
	\label{fig:wine1}
\end{figure*}

The results corresponding to Fig. 1 are shown below. 
While PCA is applied to the entire dataset without label information, LDA is used with label information. 
Since the analysis in Fig. 1 is for separating the wine groups, cPCA is not used here. 
All the embedding results (\autoref{fig:wine1}) with PCA, LDA, and ULCA show similar patterns; however, we can see LDA and ULCA have a better separation among groups. 
Also, from the axis information (\autoref{table:wine1}), we can see that both LDA and ULCA show \texttt{flavanoids} and \texttt{proline} have strong contribution to the embedding axes. 
These results demonstrate that ULCA embraces LDA's comparative analysis capabilities.

\begin{table}[h]
  \centering
  \caption{The axis information of \autoref{fig:wine1}.}
  \label{table:wine1}
  \begin{tabular}{lrrrrrrrr}
\toprule
{} &   PCA (x) &   PCA (y) &   LDA (x) &   LDA (y) &  cPCA (x) &  cPCA (y) &  ULCA (x) &  ULCA (y) \\
\midrule
alcohol              &  0.144329 & -0.483652 & -0.326569 &  0.705754 &         0 &         0 &  0.047908 &  0.222532 \\
malic\_acid           & -0.245188 & -0.224931 &  0.184094 &  0.340194 &         0 &         0 & -0.119098 & -0.001570 \\
ash                  & -0.002051 & -0.316069 & -0.100969 &  0.641759 &         0 &         0 & -0.068764 &  0.373212 \\
alcalinity\_of\_ash    & -0.239320 &  0.010591 &  0.515503 & -0.487473 &         0 &         0 & -0.140328 & -0.391263 \\
magnesium            &  0.141992 & -0.299634 & -0.030813 & -0.006591 &         0 &         0 &  0.020823 & -0.065924 \\
total\_phenols        &  0.394661 & -0.065040 &  0.385720 & -0.020104 &         0 &         0 & -0.052743 &  0.232090 \\
flavanoids           &  0.422934 &  0.003360 & -1.654628 & -0.490054 &         0 &         0 &  0.780194 & -0.385695 \\
nonflavanoid\_phenols & -0.298533 & -0.028779 & -0.185636 & -0.202407 &         0 &         0 &  0.062533 & -0.143723 \\
proanthocyanins      &  0.313429 & -0.039302 &  0.076533 & -0.175270 &         0 &         0 &  0.016385 &  0.054407 \\
color\_intensity      & -0.088617 & -0.529996 &  0.820805 &  0.585410 &         0 &         0 & -0.344212 & -0.115462 \\
hue                  &  0.296715 &  0.279235 & -0.186454 & -0.345456 &         0 &         0 &  0.227938 & -0.007595 \\
od280/od315          &  0.376167 &  0.164496 & -0.819544 &  0.036238 &         0 &         0 &  0.362277 &  0.386084 \\
proline              &  0.286752 & -0.364903 & -0.845097 &  0.895899 &         0 &         0 &  0.203277 &  0.514844 \\
\bottomrule
\end{tabular}

\end{table}

\clearpage

\begin{figure*}[h]
    \captionsetup{farskip=0pt}
	\centering
    \includegraphics[width=1.0\linewidth]{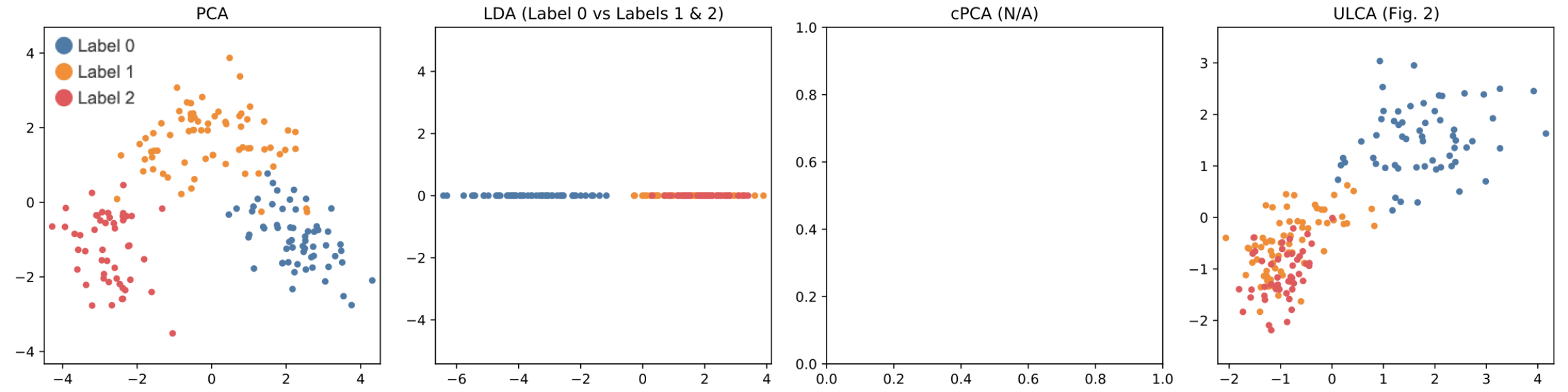}
  	\caption{The embedding results corresponding to Fig. 2.}
	\label{fig:wine2}
\end{figure*}

The results corresponding to Fig. 2 are shown below. 
In Fig. 2, we review the factors common between Labels 1 and 2 but different from Label 0.
Thus, we apply LDA to the dataset to distinguish Labels 1 and 2 from Label 0 (i.e., in LDA, we aggregate data points in Labels 1 and 2 into one class). 
Note that when using LDA, the number of embedding axes needs to be lower or equal to the number of classes; thus, \autoref{fig:wine2} shows 1D LDA result. 
Both LDA and ULCA show a clear separation between the groups and similar attributes' contribution to the embedding result, as shown in \autoref{table:wine2} (e.g., \texttt{proline} has the largest contribution to $x$-axis). 

\begin{table}[h]
  \centering
  \caption{The axis information of \autoref{fig:wine2}.}
  \label{table:wine2}
  \begin{tabular}{lrrrrrrrr}
\toprule
{} &   PCA (x) &   PCA (y) &   LDA (x) &  LDA (y) &  cPCA (x) &  cPCA (y) &  ULCA (x) &  ULCA (y) \\
\midrule
alcohol              &  0.144329 & -0.483652 & -0.721476 &      0.0 &         0 &         0 &  0.233373 &  0.437619 \\
malic\_acid           & -0.245188 & -0.224931 & -0.104383 &      0.0 &         0 &         0 &  0.004880 & -0.052372 \\
ash                  & -0.002051 & -0.316069 & -0.516378 &      0.0 &         0 &         0 &  0.012357 &  0.282481 \\
alcalinity\_of\_ash    & -0.239320 &  0.010591 &  0.704967 &      0.0 &         0 &         0 & -0.093801 & -0.512145 \\
magnesium            &  0.141992 & -0.299634 & -0.017396 &      0.0 &         0 &         0 & -0.090407 &  0.200818 \\
total\_phenols        &  0.394661 & -0.065040 &  0.288819 &      0.0 &         0 &         0 &  0.125364 &  0.175940 \\
flavanoids           &  0.422934 &  0.003360 & -0.839866 &      0.0 &         0 &         0 &  0.446139 & -0.047882 \\
nonflavanoid\_phenols & -0.298533 & -0.028779 &  0.007867 &      0.0 &         0 &         0 & -0.102017 &  0.027326 \\
proanthocyanins      &  0.313429 & -0.039302 &  0.175919 &      0.0 &         0 &         0 &  0.051957 &  0.027711 \\
color\_intensity      & -0.088617 & -0.529996 &  0.179579 &      0.0 &         0 &         0 &  0.097398 &  0.017415 \\
hue                  &  0.296715 &  0.279235 &  0.106346 &      0.0 &         0 &         0 &  0.141806 &  0.005798 \\
od280/od315          &  0.376167 &  0.164496 & -0.609172 &      0.0 &         0 &         0 & -0.152601 &  0.620854 \\
proline              &  0.286752 & -0.364903 & -1.222698 &      0.0 &         0 &         0 &  0.804762 & -0.052649 \\
\bottomrule
\end{tabular}

\end{table}

\clearpage

\begin{figure*}[h]
    \captionsetup{farskip=0pt}
	\centering
    \includegraphics[width=1.0\linewidth]{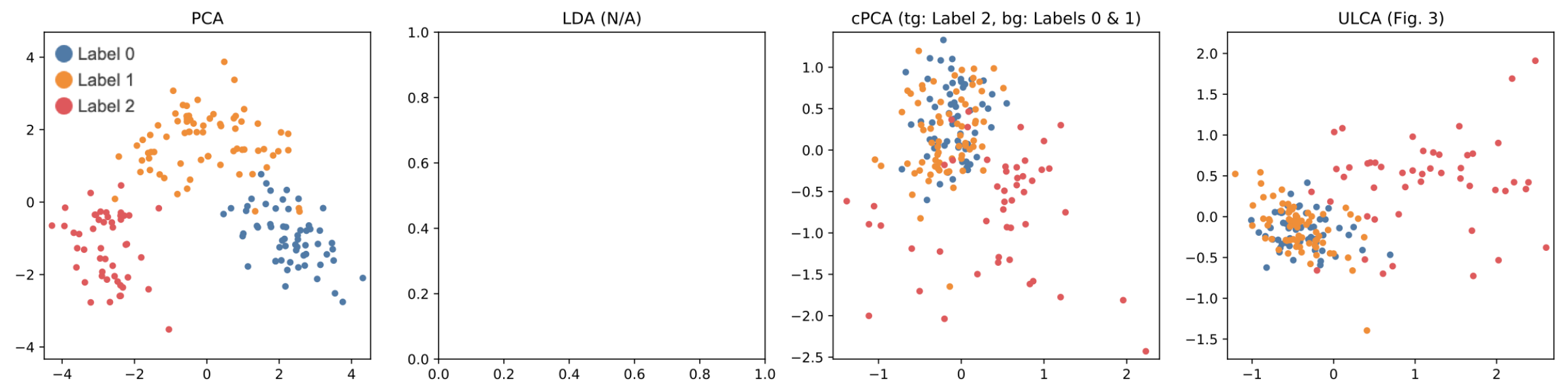}
  	\caption{The embedding results corresponding to Fig. 3.}
	\label{fig:wine3}
\end{figure*}

The results corresponding to Fig. 3 are shown below. 
In Fig. 3, we identify the factors for which wines in Label 2 have high variety than the others. 
Thus, for this analysis, we use cPCA but not LDA. 
Since cPCA only takes two groups as its inputs (i.e., target and background groups), we set data points in Label 2 as a target group and the other data points as a background group.
cPCA's $\alpha$ value is automatically selected by using the existing implementation available online\footnote{\url{https://github.com/takanori-fujiwara/ccpca}}.
Based on \autoref{fig:wine3} and \autoref{table:wine3}, cPCA and ULCA have similar embedding results and axes. 
Throughout the analyses in this section, we can observe that ULCA embraces both LDA and cPCA's analysis capabilities.

\begin{table}[h]
  \centering
  \caption{The axis information of \autoref{fig:wine3}.}
  \label{table:wine3}
  \begin{tabular}{lrrrrrrrr}
\toprule
{} &   PCA (x) &   PCA (y) &  LDA (x) &  LDA (y) &  cPCA (x) &  cPCA (y) &  ULCA (x) &  ULCA (y) \\
\midrule
alcohol              &  0.144329 & -0.483652 &        0 &        0 &  0.018234 & -0.473961 & -0.144189 &  0.073715 \\
malic\_acid           & -0.245188 & -0.224931 &        0 &        0 & -0.004813 &  0.085529 &  0.092141 & -0.057192 \\
ash                  & -0.002051 & -0.316069 &        0 &        0 &  0.027183 & -0.238400 & -0.003362 &  0.066421 \\
alcalinity\_of\_ash    & -0.239320 &  0.010591 &        0 &        0 &  0.012851 & -0.045151 &  0.049720 &  0.003406 \\
magnesium            &  0.141992 & -0.299634 &        0 &        0 & -0.163372 &  0.028401 &  0.008755 & -0.088388 \\
total\_phenols        &  0.394661 & -0.065040 &        0 &        0 &  0.442786 &  0.010927 & -0.243510 &  0.511967 \\
flavanoids           &  0.422934 &  0.003360 &        0 &        0 & -0.742596 &  0.071961 & -0.100889 & -0.811436 \\
nonflavanoid\_phenols & -0.298533 & -0.028779 &        0 &        0 &  0.257193 &  0.216337 & -0.077980 &  0.128603 \\
proanthocyanins      &  0.313429 & -0.039302 &        0 &        0 &  0.192331 & -0.135110 &  0.012373 &  0.163924 \\
color\_intensity      & -0.088617 & -0.529996 &        0 &        0 &  0.232810 & -0.238384 &  0.903389 &  0.061729 \\
hue                  &  0.296715 &  0.279235 &        0 &        0 &  0.026516 &  0.005303 &  0.089583 & -0.026993 \\
od280/od315          &  0.376167 &  0.164496 &        0 &        0 &  0.250281 &  0.197020 &  0.127810 &  0.102718 \\
proline              &  0.286752 & -0.364903 &        0 &        0 &  0.062119 &  0.736438 & -0.227985 & -0.004630 \\
\bottomrule
\end{tabular}

\end{table}

\clearpage

\section{Results for the PPIC Statewide Survey}

\begin{figure*}[h]
    \captionsetup{farskip=0pt}
	\centering
    \includegraphics[width=1.0\linewidth]{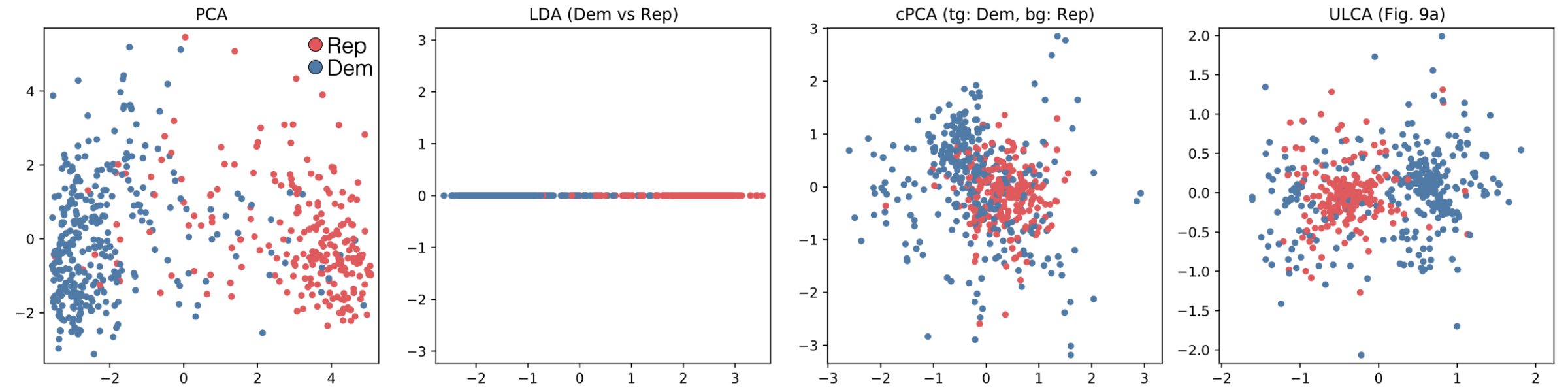}
  	\caption{The embedding results corresponding to Fig. 9a.}
	\label{fig:ppic1}
\end{figure*}

The results corresponding to Fig. 9a are shown below. 
The analysis in Fig. 9a aims to find subgroups from the Democrat supporters (\texttt{Dem}) by comparing them with the Republican supporters (\texttt{Rep}). 
Thus, we apply all DR methods to the dataset and cPCA's $\alpha$ is automatically selected.
From \autoref{fig:ppic1}, we can see that while PCA and LDA clearly separate \texttt{Dem} and \texttt{Rep}, they do not depict any clear subgroups. 
On the other hand, cPCA and ULCA show subgroups of \texttt{Dem}.
However, ULCA with manually selected $\alpha$ more clearly shows two distinct subgroups and the attribute with a dominant contribution to the embedding (i.e., \texttt{Q30}). 

\begin{table}[h]
  \centering
  \caption{The axis information of \autoref{fig:ppic1}.}
  \label{table:ppic1}
  \begin{tabular}{lrrrrrrrr}
\toprule
{} &   PCA (x) &   PCA (y) &   LDA (x) &  LDA (y) &  cPCA (x) &  cPCA (y) &  ULCA (x) &  ULCA (y) \\
\midrule
q1   & -0.262499 & -0.029540 & -0.135952 &      0.0 &  0.689553 &  0.162083 & -0.344023 & -0.050108 \\
q7   &  0.061919 & -0.412986 & -0.097428 &      0.0 & -0.013778 &  0.139085 &  0.036106 & -0.019798 \\
q13  & -0.064540 & -0.360961 &  0.018720 &      0.0 & -0.161281 &  0.015959 & -0.020793 & -0.004293 \\
q18  & -0.013561 & -0.076335 & -0.069628 &      0.0 &  0.095442 & -0.008482 &  0.001541 &  0.020085 \\
q19  & -0.078830 & -0.045852 & -0.178542 &      0.0 & -0.006679 &  0.055236 &  0.057741 & -0.037453 \\
q20  &  0.301285 & -0.056628 &  0.374915 &      0.0 &  0.168030 &  0.427085 &  0.183248 & -0.504519 \\
q21  &  0.172963 &  0.011879 & -0.009094 &      0.0 &  0.068641 &  0.063169 &  0.046290 & -0.021867 \\
q21a &  0.292849 & -0.010057 &  0.352594 &      0.0 &  0.119974 &  0.193656 &  0.051198 &  0.651416 \\
q22  &  0.264119 & -0.079989 &  0.052627 &      0.0 &  0.117895 &  0.118933 & -0.111115 &  0.027239 \\
q23  &  0.202948 & -0.056151 &  0.009101 &      0.0 & -0.014212 &  0.194740 &  0.123681 & -0.181099 \\
q24  & -0.242087 & -0.080452 & -0.101145 &      0.0 &  0.093145 & -0.157120 &  0.064125 &  0.459837 \\
q25  & -0.238140 & -0.035276 &  0.018476 &      0.0 &  0.223884 & -0.015996 & -0.000025 &  0.047087 \\
q26  &  0.282375 & -0.043620 & -0.144249 &      0.0 &  0.293312 &  0.204617 & -0.006155 &  0.053229 \\
q27  & -0.259116 &  0.005183 & -0.087271 &      0.0 & -0.277763 &  0.489935 &  0.053455 & -0.122329 \\
q28  & -0.253654 & -0.045732 & -0.318113 &      0.0 & -0.259232 &  0.232197 &  0.147868 &  0.002127 \\
q29  & -0.264244 & -0.044850 & -0.033479 &      0.0 & -0.133985 &  0.002701 & -0.127783 & -0.110617 \\
q30  & -0.244617 & -0.073278 & -0.328858 &      0.0 & -0.018518 &  0.143548 &  0.851559 &  0.024652 \\
q31  &  0.265555 & -0.000986 &  0.334397 &      0.0 & -0.123369 & -0.229083 & -0.161481 & -0.046006 \\
q33  & -0.257920 & -0.076160 & -0.274269 &      0.0 & -0.038854 &  0.320887 & -0.040543 & -0.118824 \\
q34  &  0.017638 & -0.452132 & -0.002937 &      0.0 & -0.157850 &  0.156949 &  0.028108 &  0.094607 \\
q36  &  0.028359 & -0.385510 &  0.041148 &      0.0 & -0.063469 &  0.196644 & -0.047058 & -0.067060 \\
q37a & -0.003487 & -0.078765 &  0.013674 &      0.0 &  0.047617 &  0.169870 & -0.012261 & -0.010002 \\
d3a  &  0.023643 & -0.250583 & -0.011744 &      0.0 &  0.125461 & -0.001833 & -0.000281 &  0.027467 \\
d6   & -0.000859 & -0.114435 & -0.029078 &      0.0 & -0.114432 &  0.109959 &  0.050348 &  0.017560 \\
d7   & -0.033413 & -0.259972 &  0.123900 &      0.0 & -0.112893 &  0.159850 &  0.067566 & -0.047822 \\
d9a  &  0.043446 & -0.120803 &  0.038419 &      0.0 & -0.116516 &  0.081919 &  0.018063 & -0.056948 \\
d1a  &  0.037022 & -0.367580 &  0.024154 &      0.0 &  0.130929 & -0.041311 & -0.043981 &  0.041030 \\
\bottomrule
\end{tabular}

\end{table}

\clearpage

\begin{figure*}[h]
    \captionsetup{farskip=0pt}
	\centering
    \includegraphics[width=1.0\linewidth]{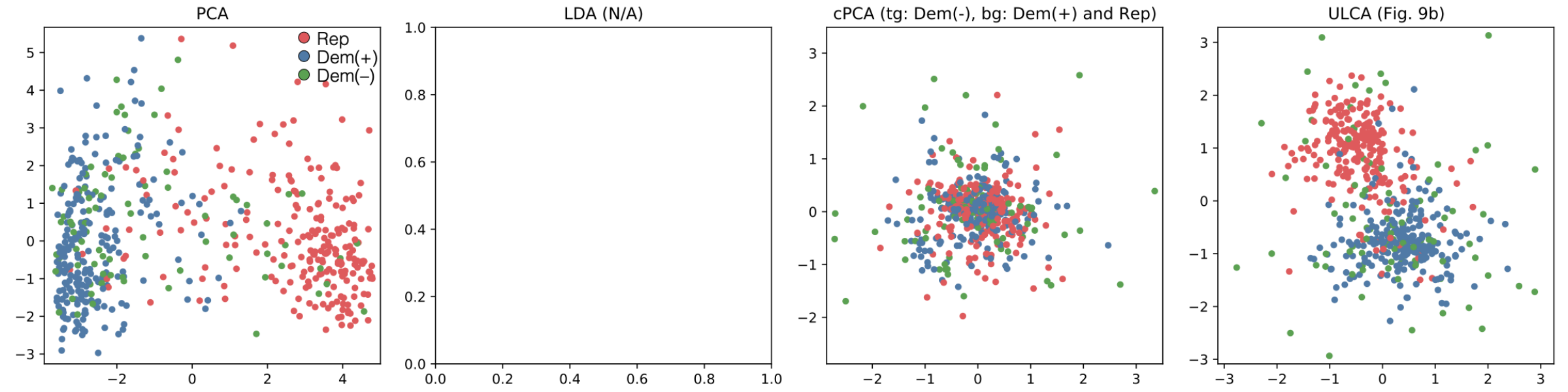}
  	\caption{The embedding results corresponding to Fig. 9b.}
	\label{fig:ppic2}
\end{figure*}

The results corresponding to Fig. 9b are shown below. 
The analysis in Fig. 9b is to find opinions more varied in \texttt{Dem(-)} than the other groups.  
Thus, only PCA and cPCA are related to this analysis.
For cPCA, we set \texttt{Dem(-)} as a target group and the other two groups as an aggregated background group, and select $\alpha$ automatically.
From \autoref{fig:ppic2}, when compared with the result with PCA, \texttt{Dem(-)} is more widely distributed than others when using cPCA and ULCA. 
Note that here we discarded \texttt{Q30} from attributes used in each DR; however, PCA has the result highly similar to \autoref{fig:ppic1}.
This implies \texttt{Q30} has a limited influence on PCA's embedding, and the subgroup we found in \autoref{fig:ppic1} with ULCA or cPCA is difficult to find with PCA.
When comparing cPCA and ULCA, ULCA shows separation between \texttt{Dem(+)} and \texttt{Rep}. 
We can consider that this is because ULCA can individually handle each of the target groups, which is infeasible with cPCA. 
When computing covariance matrices, ULCA applies centering to each of the groups individually; as a result, in the embedding space, each group's variance is condensed around its centroid, unlike cPCA where the common centroid of the aggregated group is used. 
ULCA can provide more precise controls to embed datasets than cPCA, especially when we have more than two groups.

\begin{table}[h]
  \centering
  \caption{The axis information of \autoref{fig:ppic2}.}
  \label{table:ppic2}
  \begin{tabular}{lrrrrrrrr}
\toprule
{} &   PCA (x) &   PCA (y) &  LDA (x) &  LDA (y) &  cPCA (x) &  cPCA (y) &  ULCA (x) &  ULCA (y) \\
\midrule
q1   & -0.272006 & -0.025406 &        0 &        0 &  0.472050 &  0.373206 &  0.558180 &  0.159270 \\
q7   &  0.051561 & -0.415685 &        0 &        0 & -0.085978 & -0.005874 & -0.183562 &  0.071507 \\
q13  & -0.068367 & -0.347333 &        0 &        0 & -0.065163 &  0.004781 &  0.015901 & -0.270734 \\
q18  & -0.015867 & -0.121445 &        0 &        0 &  0.001138 & -0.024359 & -0.015735 & -0.003108 \\
q19  & -0.086224 & -0.067838 &        0 &        0 &  0.055548 &  0.118862 &  0.093673 & -0.043103 \\
q20  &  0.307895 & -0.044009 &        0 &        0 &  0.199424 &  0.494824 & -0.132921 &  0.339701 \\
q21  &  0.180770 &  0.015282 &        0 &        0 &  0.032325 &  0.045913 &  0.086195 & -0.030779 \\
q21a &  0.296603 &  0.011459 &        0 &        0 &  0.197613 & -0.489857 & -0.222388 &  0.412679 \\
q22  &  0.270156 & -0.062986 &        0 &        0 &  0.194321 & -0.017552 &  0.118428 &  0.050037 \\
q23  &  0.203250 & -0.043077 &        0 &        0 & -0.073007 &  0.009687 & -0.157588 &  0.100453 \\
q24  & -0.253813 & -0.053852 &        0 &        0 &  0.271304 & -0.120468 &  0.317152 &  0.080041 \\
q25  & -0.252807 & -0.008591 &        0 &        0 &  0.386441 & -0.096710 & -0.013817 &  0.433229 \\
q26  &  0.288913 & -0.046929 &        0 &        0 &  0.373558 & -0.115602 &  0.225641 &  0.314264 \\
q27  & -0.266678 &  0.006934 &        0 &        0 & -0.153651 &  0.278042 &  0.069138 & -0.276877 \\
q28  & -0.261923 & -0.055755 &        0 &        0 & -0.340273 &  0.052775 & -0.192458 & -0.321027 \\
q29  & -0.273050 & -0.041173 &        0 &        0 & -0.090726 & -0.413597 & -0.381342 & -0.012470 \\
q31  &  0.276355 & -0.021581 &        0 &        0 & -0.258279 &  0.108430 &  0.000249 & -0.175618 \\
q33  & -0.270587 & -0.074903 &        0 &        0 &  0.163379 & -0.122102 &  0.000337 &  0.086607 \\
q34  &  0.013855 & -0.446806 &        0 &        0 &  0.118189 & -0.073985 &  0.152876 &  0.019129 \\
q36  &  0.024034 & -0.365329 &        0 &        0 &  0.005342 & -0.101658 & -0.247427 &  0.155303 \\
q37a & -0.002479 & -0.094320 &        0 &        0 &  0.032020 &  0.027129 &  0.095491 & -0.030879 \\
d3a  &  0.013835 & -0.259995 &        0 &        0 &  0.010438 &  0.113828 &  0.159899 & -0.062622 \\
d6   & -0.010729 & -0.079336 &        0 &        0 & -0.056727 & -0.018921 & -0.046021 & -0.178001 \\
d7   & -0.039018 & -0.279497 &        0 &        0 & -0.035465 &  0.047200 & -0.087423 &  0.133693 \\
d9a  &  0.043838 & -0.115847 &        0 &        0 & -0.058634 & -0.100970 & -0.192951 &  0.056562 \\
d1a  &  0.032435 & -0.393590 &        0 &        0 &  0.113705 &  0.035856 &  0.163867 &  0.043843 \\
\bottomrule
\end{tabular}

\end{table}

\clearpage

\begin{figure*}[h]
    \captionsetup{farskip=0pt}
	\centering
    \includegraphics[width=1.0\linewidth]{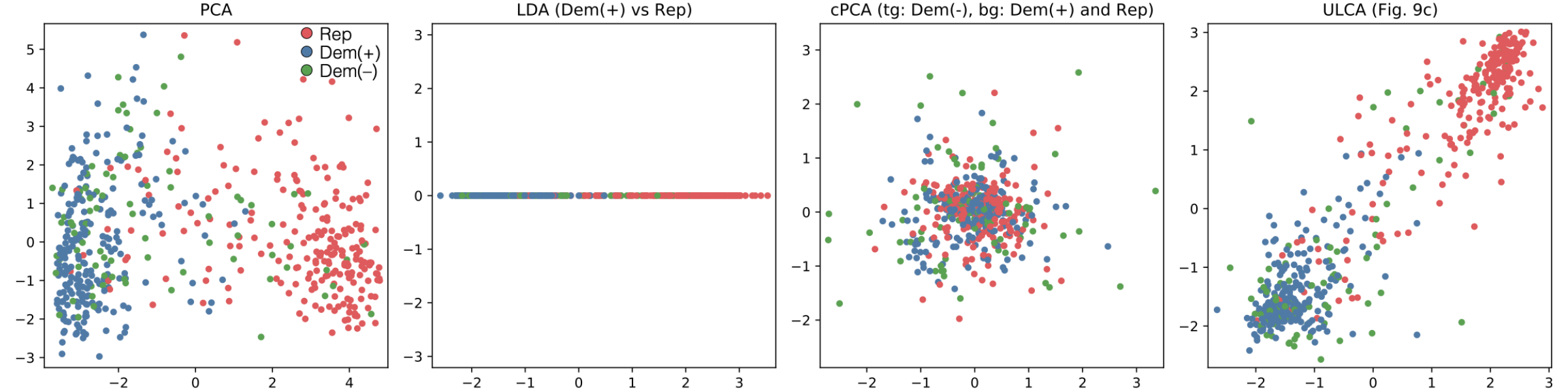}
  	\caption{The embedding results corresponding to Fig. 9c.}
	\label{fig:ppic3}
\end{figure*}

The results corresponding to Fig. 9c are shown below. 
The analysis in Fig. 9c is to find opinions more varied in \texttt{Dem(-)} than the other groups but at the same time clearly distinguish \texttt{Dem(+)} and \texttt{Rep}.  
Thus, we apply all the DR methods (LDA relates to finding the separation while cPCA related to finding more diverse opinions in \texttt{Dem(-)}).
For cPCA, we use the same setting with \autoref{fig:ppic2}.
For LDA, to focus on distinguishing \texttt{Dem(+)} and \texttt{Rep}, we only use data points in \texttt{Dem(+)} and \texttt{Rep} while learning a projection matrix, and utilize this projection matrix to plot \texttt{Dem(-)} in the same embedding space.
From \autoref{fig:ppic3}, unlike the other methods, ULCA shows the clear separation between \texttt{Dem(+)} and \texttt{Rep} while producing high variance for \texttt{Dem(-)}.

\begin{table}[h]
  \centering
  \caption{The axis information of \autoref{fig:ppic3}.}
  \label{table:ppic3}
  \begin{tabular}{lrrrrrrrr}
\toprule
{} &   PCA (x) &   PCA (y) &   LDA (x) &  LDA (y) &  cPCA (x) &  cPCA (y) &  ULCA (x) &  ULCA (y) \\
\midrule
q1   & -0.272006 & -0.025406 & -0.181956 &      0.0 &  0.472050 &  0.373206 & -0.191644 & -0.124360 \\
q7   &  0.051561 & -0.415685 & -0.113320 &      0.0 & -0.085978 & -0.005874 &  0.107413 &  0.004499 \\
q13  & -0.068367 & -0.347333 & -0.011961 &      0.0 & -0.065163 &  0.004781 & -0.075113 & -0.083833 \\
q18  & -0.015867 & -0.121445 & -0.059593 &      0.0 &  0.001138 & -0.024359 & -0.062652 & -0.075154 \\
q19  & -0.086224 & -0.067838 & -0.165659 &      0.0 &  0.055548 &  0.118862 & -0.071863 & -0.086774 \\
q20  &  0.307895 & -0.044009 &  0.449496 &      0.0 &  0.199424 &  0.494824 &  0.790036 & -0.142629 \\
q21  &  0.180770 &  0.015282 &  0.003009 &      0.0 &  0.032325 &  0.045913 &  0.002213 &  0.031159 \\
q21a &  0.296603 &  0.011459 &  0.346843 &      0.0 &  0.197613 & -0.489857 &  0.023907 &  0.676308 \\
q22  &  0.270156 & -0.062986 &  0.047464 &      0.0 &  0.194321 & -0.017552 &  0.032788 &  0.171582 \\
q23  &  0.203250 & -0.043077 &  0.022052 &      0.0 & -0.073007 &  0.009687 &  0.169506 &  0.060299 \\
q24  & -0.253813 & -0.053852 & -0.159578 &      0.0 &  0.271304 & -0.120468 & -0.347739 &  0.057961 \\
q25  & -0.252807 & -0.008591 &  0.014950 &      0.0 &  0.386441 & -0.096710 &  0.028889 &  0.156015 \\
q26  &  0.288913 & -0.046929 & -0.113532 &      0.0 &  0.373558 & -0.115602 & -0.144949 &  0.331211 \\
q27  & -0.266678 &  0.006934 & -0.087042 &      0.0 & -0.153651 &  0.278042 &  0.016656 & -0.235233 \\
q28  & -0.261923 & -0.055755 & -0.341739 &      0.0 & -0.340273 &  0.052775 & -0.120233 & -0.436055 \\
q29  & -0.273050 & -0.041173 & -0.080050 &      0.0 & -0.090726 & -0.413597 & -0.046404 & -0.123246 \\
q31  &  0.276355 & -0.021581 &  0.311197 &      0.0 & -0.258279 &  0.108430 &  0.180075 &  0.136549 \\
q33  & -0.270587 & -0.074903 & -0.271110 &      0.0 &  0.163379 & -0.122102 & -0.182615 & -0.109055 \\
q34  &  0.013855 & -0.446806 &  0.003489 &      0.0 &  0.118189 & -0.073985 & -0.134511 &  0.087867 \\
q36  &  0.024034 & -0.365329 &  0.035556 &      0.0 &  0.005342 & -0.101658 &  0.086948 &  0.047533 \\
q37a & -0.002479 & -0.094320 &  0.014897 &      0.0 &  0.032020 &  0.027129 & -0.044884 &  0.000106 \\
d3a  &  0.013835 & -0.259995 & -0.020982 &      0.0 &  0.010438 &  0.113828 & -0.011600 &  0.075408 \\
d6   & -0.010729 & -0.079336 & -0.040659 &      0.0 & -0.056727 & -0.018921 &  0.027291 &  0.061168 \\
d7   & -0.039018 & -0.279497 &  0.137705 &      0.0 & -0.035465 &  0.047200 &  0.149902 &  0.052898 \\
d9a  &  0.043838 & -0.115847 &  0.043950 &      0.0 & -0.058634 & -0.100970 &  0.076157 &  0.040297 \\
d1a  &  0.032435 & -0.393590 &  0.021741 &      0.0 &  0.113705 &  0.035856 & -0.014708 &  0.009417 \\
\bottomrule
\end{tabular}

\end{table}

\clearpage

\section{Results for the MNIST Handwritten Digits Dataset}

\begin{figure*}[h]
    \captionsetup{farskip=0pt}
	\centering
    \includegraphics[width=1.0\linewidth]{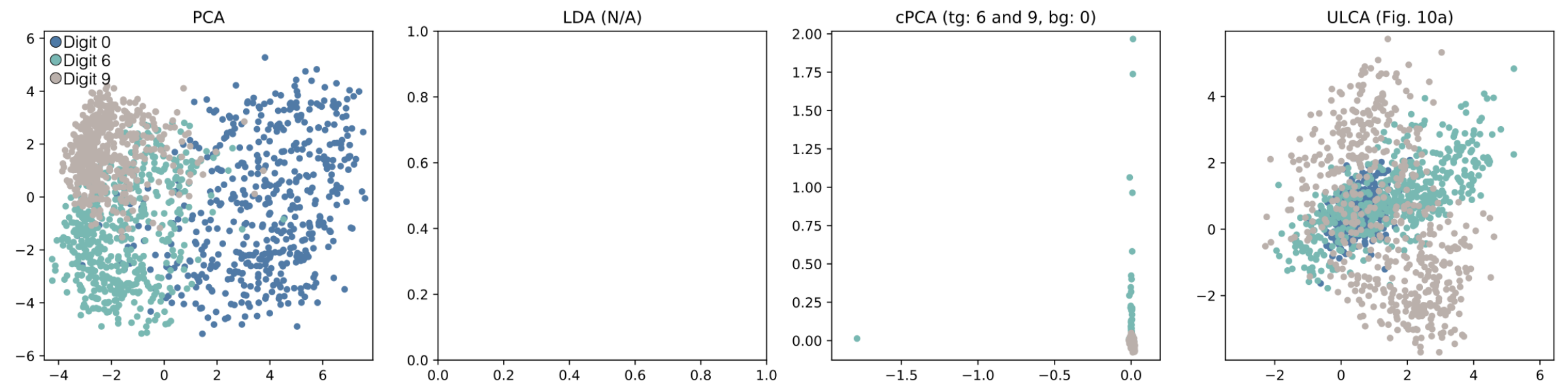}
  	\caption{The embedding results corresponding to Fig. 10a.}
	\label{fig:mnist1}
\end{figure*}

\begin{figure*}[h]
    \captionsetup{farskip=0pt}
	\centering
    \includegraphics[width=1.0\linewidth]{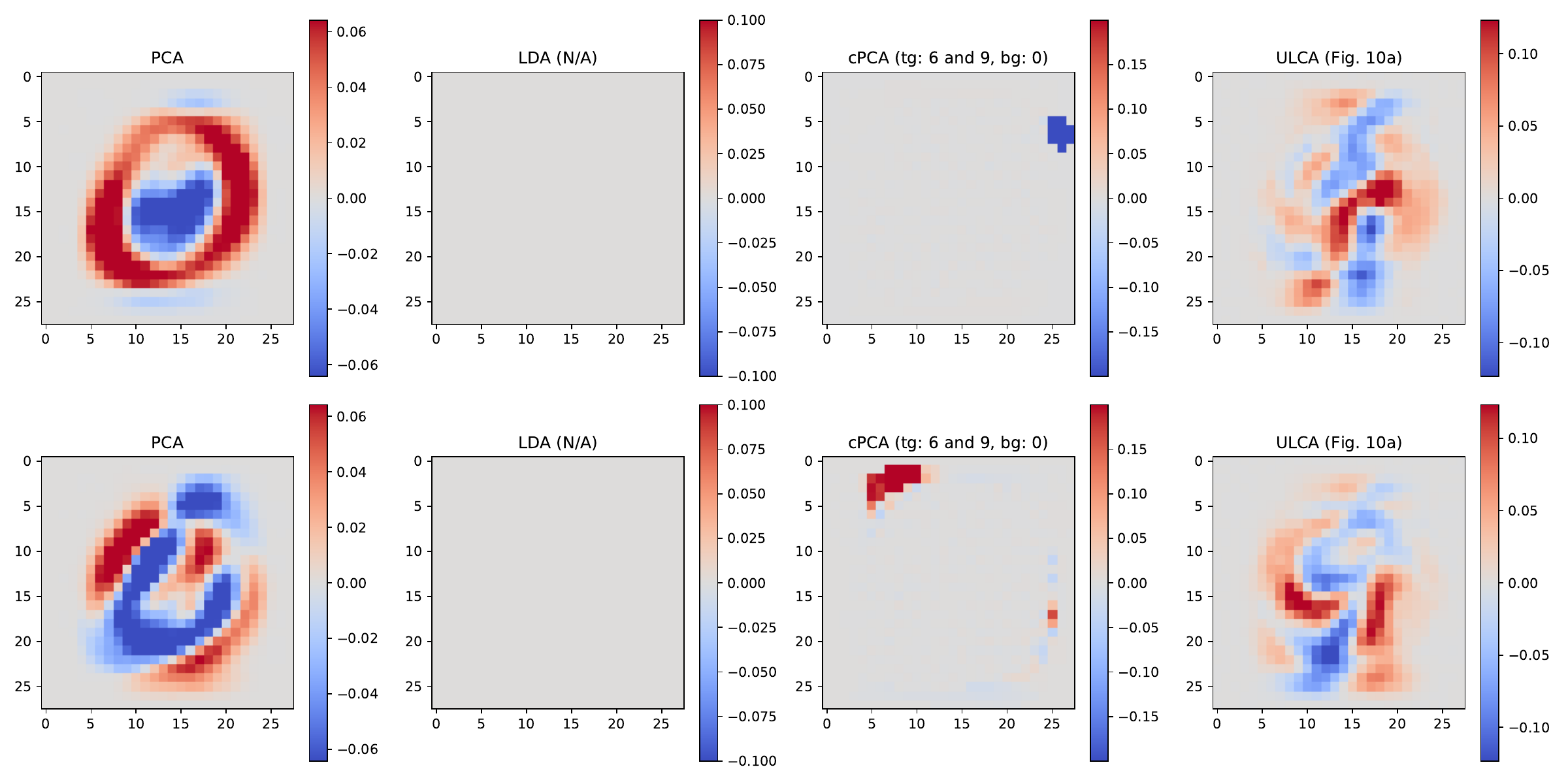}
  	\caption{The axis information of \autoref{fig:mnist1}. The heatmaps placed in the first and second rows correspond to $x$- and $y$-embedding axes, respectively.}
	\label{fig:mnist1_heatmap}
\end{figure*}

The results corresponding to Fig. 10a are shown below. 
In the analysis related to Fig. 10a, we review highly diverse structures  in \texttt{6} and \texttt{9} but not in \texttt{0}. 
Thus, for this analysis, we do not use LDA. 
For cPCA, we set \texttt{6} and \texttt{9} as a target group and \texttt{0} as a background group, and $\alpha$ is selected automatically. 
From \autoref{fig:mnist1} and \autoref{fig:mnist1_heatmap}, we can see that only ULCA clearly captures \texttt{9}'s straight lines mentioned in the case study. 

\clearpage

The results corresponding to Fig. 10b are shown below. 
In Fig. 10b, we adjust the weight related to \texttt{9}'s variance. 
In \autoref{fig:mnist2_heatmap}, we can see that ULCA captures the strokes in \texttt{6} better than \autoref{fig:mnist1_heatmap}.

\begin{figure*}[h]
    \captionsetup{farskip=0pt}
	\centering
    \includegraphics[width=1.0\linewidth]{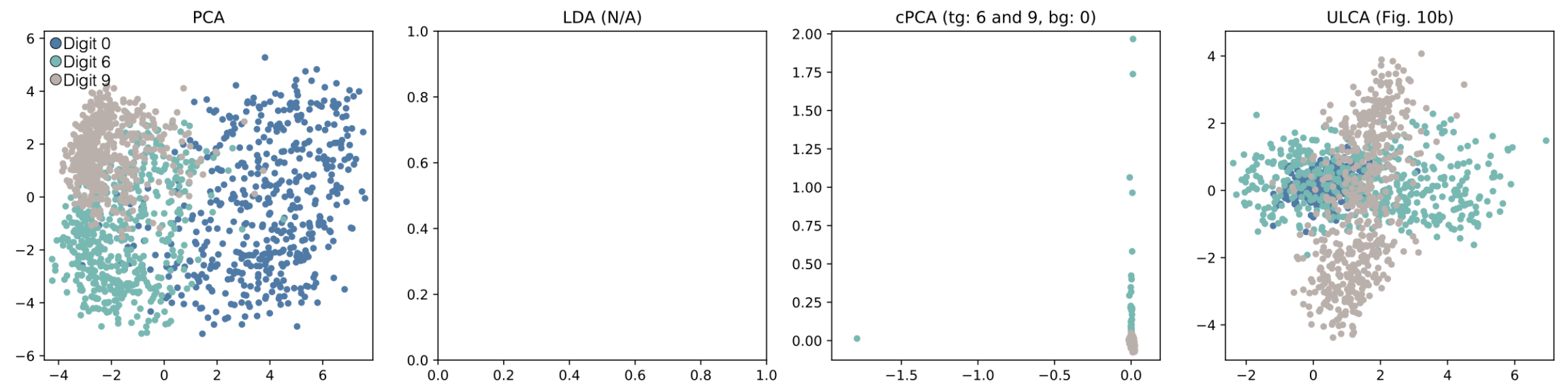}
  	\caption{The embedding results corresponding to Fig. 10b.}
	\label{fig:mnist2}
\end{figure*}

\begin{figure*}[h]
    \captionsetup{farskip=0pt}
	\centering
    \includegraphics[width=1.0\linewidth]{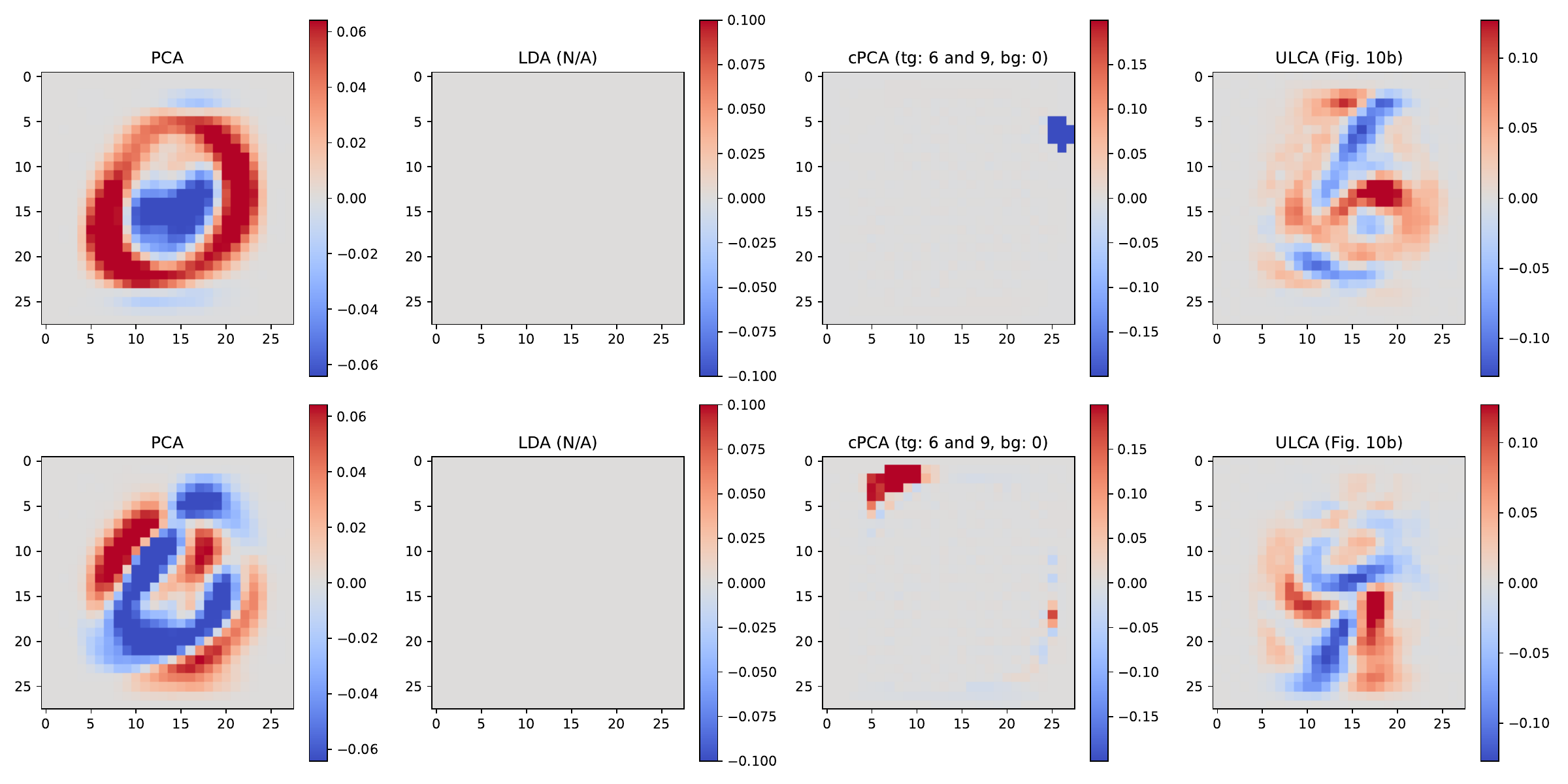}
  	\caption{The axis information of \autoref{fig:mnist2}.}
	\label{fig:mnist2_heatmap}
\end{figure*}

\clearpage

\begin{figure*}[h]
    \captionsetup{farskip=0pt}
	\centering
    \includegraphics[width=1.0\linewidth]{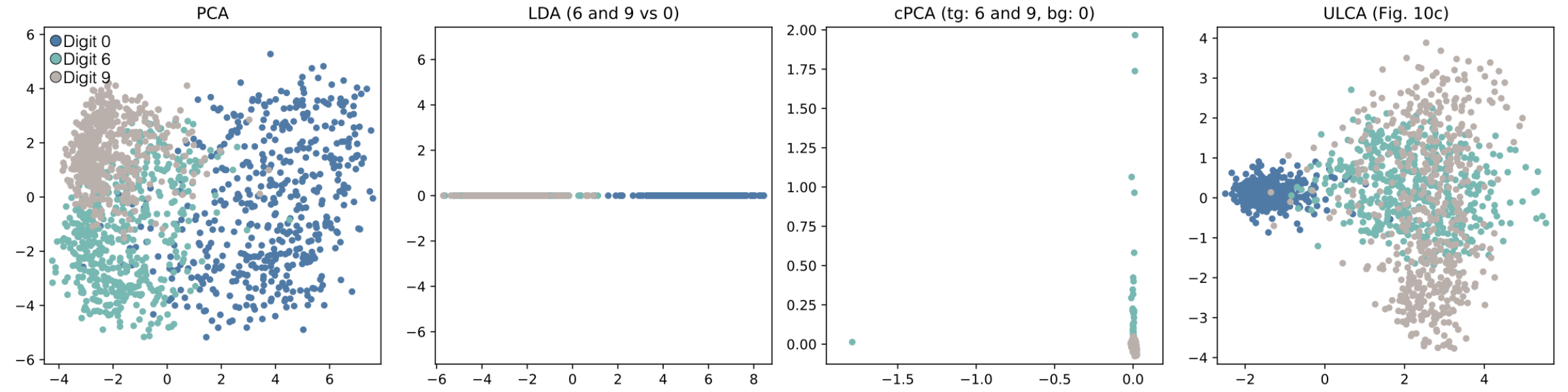}
  	\caption{The embedding results corresponding to Fig. 10c.}
	\label{fig:mnist3}
\end{figure*}

\begin{figure*}[h]
    \captionsetup{farskip=0pt}
	\centering
    \includegraphics[width=1.0\linewidth]{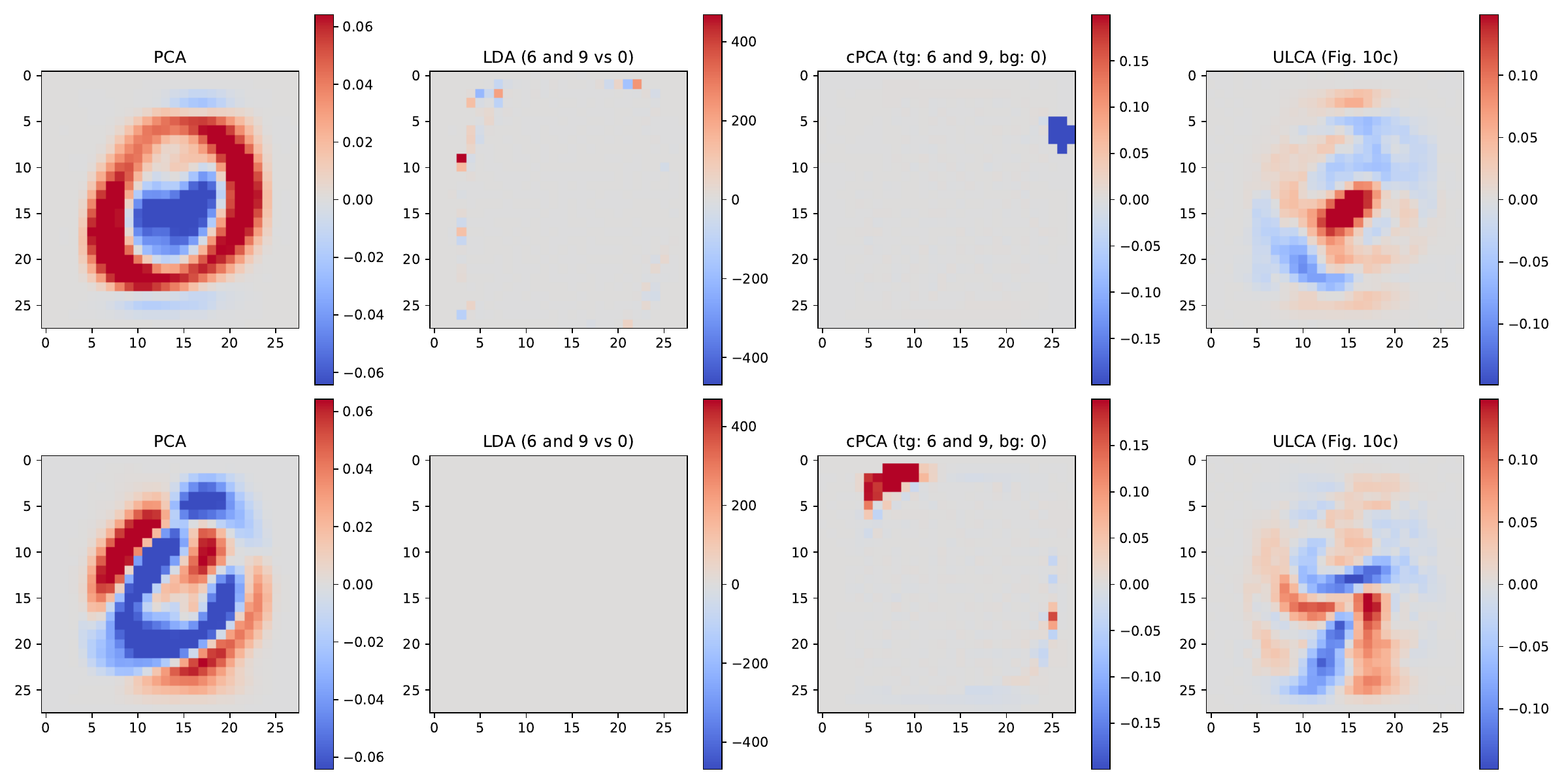}
  	\caption{The axis information of \autoref{fig:mnist3}.}
	\label{fig:mnist3_heatmap}
\end{figure*}

The results corresponding to Fig. 10c are shown below.
In Fig. 10c, we review the strokes that clearly differentiate \texttt{6} and \texttt{9} from \texttt{0} but are still written in various ways for \texttt{6} and \texttt{9}. 
Thus, for this analysis, we apply LDA to distinguish an aggregated group of \texttt{6} and \texttt{9} from a group of \texttt{0}. 
Again, as shown in \autoref{fig:mnist3}, only ULCA achieves to generate the embedding result that satisfies our analysis purpose.
Also, from \autoref{fig:mnist3_heatmap}, only ULCA reveals the useful information about the strokes.

\clearpage